\documentclass[journal]{IEEEtran}

\usepackage{cite}

\usepackage{times}
\usepackage{epsfig}
\usepackage{graphicx}
\usepackage{caption}
\graphicspath{{fig/}}
\usepackage{amsmath}
\usepackage{amssymb}

\usepackage[utf8]{inputenc} 
\usepackage[T1]{fontenc}    
\usepackage{hyperref}       
\usepackage{url}            
\usepackage{booktabs}       
\usepackage{amsfonts}       
\usepackage{nicefrac}       
\usepackage{microtype}      
\usepackage{algorithm}
\usepackage{algorithmic}
\usepackage{array}
\usepackage{bm}
\usepackage{multirow}
\usepackage{fancyhdr}
\usepackage{color}

\usepackage{colortbl}

\begin{document}
	
	\title{Kernelized Similarity Learning and Embedding for Dynamic Texture Synthesis}

	\author{
		Shiming~Chen,
		Peng~Zhang,
		Guo-Sen~Xie,
		Qinmu~Peng,
		Zehong~Cao,
		Wei~Yuan,
		and~Xinge~You,~\IEEEmembership{Senior Memember,~IEEE}
		
		\thanks{ This work was supported in part by the National Natural Science Foundation of China~( 62172177 and 61772220), Natural Science Foundation of Hubei Province under Grant (2021CFB332), the Key Science and Technology Innovation Program of Hubei Province~(2020BAB027).(Corresponding author: \textit{Xinge You}.)}
		
		\thanks{S. Chen, P. Zhang, Q. Peng and X. You are with the Department
			of Electronic Information and Communication, Huazhong University of Science and Technology, Wuhan 430074, China. (e-mail:gchenshiming@gmail.com; youxg@hust.edu.cn).}%
		\thanks{G. Xie is with the Shool of Computer Science and Engineering, Nanjing University of Science and Technology, Nanjing, China.}
		\thanks{Z. Cao is with STEM, the University of South Australia, Adelaide, Australia.}
	}

	\markboth{IEEE Transactions on Systems, Man and Cybernetics: Systems}%
	{Shell \MakeLowercase{\textit{et al.}}: Bare Demo of IEEEtran.cls for IEEE Journals}

	\maketitle

	%
	%
	
	\begin{abstract}
		Dynamic texture (DT) exhibits statistical stationarity in the spatial domain and stochastic repetitiveness in the temporal dimension, indicating that different frames of DT possess a high similarity correlation that is critical prior knowledge. However, existing methods cannot effectively learn a synthesis model for high-dimensional DT from a small number of training samples. In this paper, we propose a novel DT synthesis method, which makes full use of similarity as prior knowledge to address this issue. Our method is based on the proposed kernel similarity embedding, which can not only mitigate the high-dimensionality and small sample issues, but also has the advantage of modeling nonlinear feature relationships. Specifically, we first put forward two hypotheses that are essential for the DT model to generate new frames using similarity correlations. Then, we integrate kernel learning and the extreme learning machine into a unified synthesis model to learn kernel similarity embeddings for representing DTs. Extensive experiments on DT videos collected from the internet and two benchmark datasets, i.e., Gatech Graphcut Textures and Dyntex, demonstrate that the learned kernel similarity embeddings can provide discriminative representations for DTs. Further, our method can preserve the long-term temporal continuity of the synthesized DT sequences with excellent sustainability and generalization. Meanwhile, it effectively generates realistic DT videos with higher speed and lower computation than the current state-of-the-art methods. The code and more synthesis videos are available at our project page \url{https://shiming-chen.github.io/Similarity-page/Similarit.html}.
	\end{abstract}
	

	%
	\IEEEpeerreviewmaketitle

	\section{Introduction and Motivation}\label{sec1}
	
	\IEEEPARstart{D}{ynamic} texture (DT), which exhibits statistical stationarity in the spatial domain and stochastic repetitiveness in the temporal dimension, is one type of dynamic pattern in computer vision \cite{doretto2003dynamic}, e.g., moving vehicles, falling water, flaming fire, rotating windmill. Due to the demands of dynamic patterns synthesis in video technology applications (e.g., texture recognition \cite{feichtenhofer2017temporal}, video segmentation \cite{Mei2020SemanticSO} and super-resolution \cite{Hsu2015TemporallyCS}), synthesizing DTs has gradually become a topic of interest in computer graphics and computer vision \cite{you2016kernel,xie2017synthesizing,chen2017photographic,Xie2020CooperativeTO,tesfaldet2018two,Xie2019LearningDG}. DT synthesis aims to infer a generation process from a DT example, which then allows an infinitely varying stream of similar looking texture videos to be produced.
	
	In general, DT synthesis methods can be categorized into two groups: non-neural-network-based and neural-network-based methods. The first group methods are popular early approaches for DT synthesis, and can be further classified as physics-based methods \cite{Pegoraro2006PhysicallyBasedRF,Costantini2008HigherOS} and dynamic system (DS) modeling methods \cite{doretto2003dynamic,you2016kernel,Chan2007ClassifyingVW,Abraham2005DynamicTW,Siddiqi2007ACG}. The second group of methods automatically learn the texture distribution thanks to the effective representation of neural networks \cite{Gatys2015TextureSU,xie2017synthesizing,tesfaldet2018two,Xie2021LearningES,tulyakov2018mocogan,Xie2019LearningDG}. However, DT is high-dimensional data and with limited samples (DT samples are made up of a single training video with a short sequences length). Non-neural-network-based methods usually seek to reduce the dimensionality of DT for modeling, which may lead to information loss. Further, it is difficult to design a proper dimensionality-reduction algorithm. Meanwhile, neural-network-based methods also fail to effectively fit their large number of parameters when learning from a small number of training samples. To overcome these challenges, we propose a novel model for DT synthesis, which mines and exploits similarity correlations as prior knowledge.
	
	\begin{figure}[t]
		\begin{center}
			\includegraphics[width=1\linewidth]{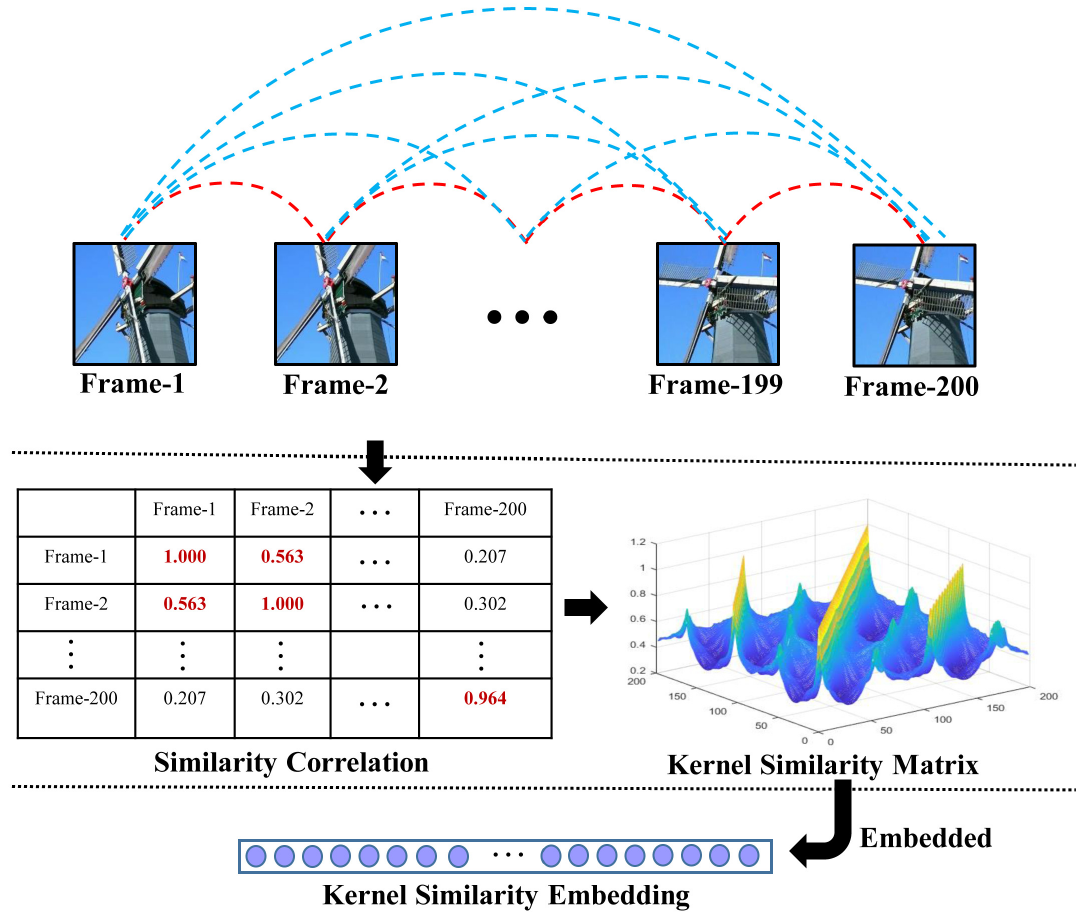}
			\\
			
			\caption{The core idea of the proposed method. DT
				exhibits statistical stationarity in the spatial domain and stochastic repetitiveness in the temporal dimension, indicating that different frames of DT possess high similarity correlation. Meanwhile, this correlation can be expressed by kernel similarity matrix and embedded into kernel similarity embedding.}
			\label{fig:similarity-embedding}
		\end{center}
	\end{figure}
	
	In fact, the similarity correlation between frames is an explicit expression of the statistical stationarity and stochastic repetitiveness of DT. It is critical for distinguishing DTs from other videos. Similarity representations serve as the learning objective of metric learning for discriminative models \cite{Chen2019SemisupervisedFL,feichtenhofer2017temporal,Zheng2019JointDA}, which indicates the importance of similarity correlations for representation. Some researchers have also attempted to mine the potential similarity knowledge of samples to improve the performance of discriminative models \cite{Cheng2016PersonRB,Fu2019SelfsimilarityGA,Hermans2017InDO,Lin2019TowardsOD,Liu2017CrossModalityBC}. This suggests that similarity correlations are as important as class labels and other annotation information when it comes to prior knowledge. Moreover, similarity correlations can explicitly capture the homogeneous and heterogeneous correlation between different frames of DT. Nevertheless, to the best of our knowledge, there are no studies on DT synthesis that consider similarity as prior knowledge to address the high-dimensionality and small sample issues. This is therefore the focus of the present paper. 
	
	To make full use of similarity as prior knowledge, we embed it into the representation of the generative model for DT synthesis. Thus, we put forward two hypotheses: 1) The content of texture video frames varies over time, and closer frames should be more similar.\footnote{This hypothesis is not limited in that the frames at several time points in the same time interval are more similar.}  2) The transition between frames can be modeled as a linear or nonlinear function to capture the similarity correlation. These hypotheses are essential if the DT model is to generate new frames based on current ones, using the similarity correlation of different frames. Fortunately, the kernel function possesses the exciting property of being able to elegantly represent the similarity of two inputs \cite{ShaweTaylor2004KernelMF}. Thus, our core idea is that the statistical stationarity in the spatial domain and the stochastic repetitiveness in the temporal dimension of DTs can be partially captured by the similarity correlation between frames. This correlation can be further elegantly exhibited by incorporating the kernel similarity matrix into the kernel similarity embedding for representation, as demonstrated in Figure \ref{fig:similarity-embedding}. Furthermore, the extreme learning machine (ELM) is an emergent technology that overcomes some challenges faced by other computational intelligence techniques. It has thus recently attracted significant research attention \cite{Huang2012ExtremeLM,Gautam2021AdaptiveOL,Lu2020RestrictedBoltzmannBasedEL,Chen2019DomainST,Li2020OnlineAE,Yang2021GraphED,Deng2019ContentInsensitiveBI}. Therefore, we attempt to make full advantage of ELM and jointly utilize kernel learning to learn a kernel similarity embedding for improving DT synthesis.
	
	In this work, we propose a novel DT synthesis method to generate high-quality, long-term DT sequences with high speed and low computation. We integrate kernel learning and ELM into a powerfully unified synthesis model to learn kernel similarity embeddings for representing the statistical stationarity in the spatial domain and stochastic repetitiveness in the temporal dimension of DT sequences. Specifically, we preprocess each input DT sequence $S_N$ ($N$ is the length of the sequence), which are divided into two parts: explanatory frames and response frames. Then, our method uses a kernel function to replace the feature mapping function of the hidden layer of the ELM, and thus the kernel similarity embedding is easily learned after training. Finally, the DT sequence is iteratively generated by our trained model.
	
	To summarize, this study makes the following salient contributions:
	
	\begin{itemize}
		
		\item  We introduce two important hypotheses that benefit the DT system, allowing it to generate new frames based on current ones, using the similarity correlation of different frames.
		
		\item  We propose a novel DT synthesis method, which learns kernel similarity embeddings to synthesize realistic video sequences with good sustainability and generalization.
		
		\item  We introduce kernel similarity embedding to mine and exploit similarity as prior knowledge of DT and analyze its availability with intuitive and theoretical insights.
		
		\item  We carry out extensive experiments on benchmark datasets to demonstrate that our method provides consistent improvement over the state-of-the-art methods. 
	\end{itemize}
	
	The remainder of this paper is organized as follows. Section \ref{sec2} provides an overview of the background and related works of DT synthesis. The proposed method based on kernel similarity embedding is elaborated in Section \ref{sec3}. The performance and evaluation are given in Section \ref{sec4}. Section \ref{sec5} presents the discussion. Section \ref{sec6} provides a summary and the directions for future research.

	\section{Background and Related Work}\label{sec2}
	The goal of DT synthesis is to generate an infinitely varying stream of similar-appearance texture videos. It aims to accurately learn a transition function $f$ from the input texture sequence $\boldsymbol{Y}=\left[\mathbf{y}_{1}, \ldots, \mathbf{y}_{N}\right]^{\top} \in \mathbb{R}^{N\times m}$ ($N$ is the length of the sequence, $m$ is the dimensionality of the frame) of training set, which can be formulated as:
	\begin{gather}
	\label{Eq1}
	\mathbf{y}_{i}=f\left(\mathbf{y}_{i-1}\right).
	\end{gather}
	A straightforward way to learn $f$ is to solve the following objective function:
	\begin{gather}
	\label{Eq2}
	f^{\prime}=\underset{f}{\operatorname{argmin}}\frac{1}{2}\sum_{t=2}^{N}\left(\mathbf{y}_{i}-f\left(\mathbf{y}_{i-1}\right)\right)^{2}.
	\end{gather}
	After training, $f^{\prime}$ can be learnt. Subsequently, given an initial frame $\mathbf{y}_{l-1}^{\prime}$ from an input texture sequence of test set, the endless sequences $\left\{\mathbf{y}_{i}^{\prime}\right\}_{l=2,3, \dots}$ can be generated iteratively by:
	\begin{gather}
	\label{Eq3}
	\mathbf{y}_{i}^{\prime}={f^{\prime}}\left(\mathbf{y}_{i-1}^{\prime}\right).
	\end{gather}

	In the following, we provide a comprehensive review of related work on DT synthesis based on various methods.
	
	\subsection{Non-Neural-Network-Based Methods}\label{sec2.1}
	\subsubsection{Physics-Based Methods}\label{sec2.1.1}
	Physics-based methods describe DT by simulating its physical mechanism using complicated models. In \cite{Pegoraro2006PhysicallyBasedRF}, Pegoraro and Parker presented a new method for the physically-based rendering of frames from detailed simulations of frame dynamics, which accounts for their unique characteristics. This method can synthesize highly realistic renderings of various types of frames. Higher-order singular value decomposition (HOSVD) analysis for DT synthesis was proposed in \cite{Costantini2008HigherOS}. It decomposes the DT into a multi-dimensional signal without unfolding the video sequences as column vectors, thus allowing dimensionality-reduction to be performed in the spatial, temporal, and chromatic domain. In summary, although physics-based methods can generate impressive DTs, they are highly application-specific with weak generalization.
	
	\subsubsection{Dynamic System Modeling Methods}\label{sec2.1.2}
	Dynamic system (DS) modeling methods for DT synthesis are the most popular non-neural-network-based methods. DS modeling methods typically learn transition functions to represent the correlation between different frames of DTs using linear or nonlinear dimensionality-reduction algorithms. However, a proper dimensionality-reduction algorithm is difficult to design, which may lead to the information loss of DT. This is the main limitation of DS modeling methods. In \cite{doretto2003dynamic}, Doretto et al. proposed a pioneering DS method for DT synthesis using a simple linear dynamic system (LDS) to project the input video frames into lower-dimensional space by SVD. Siddiqi et al. \cite{Siddiqi2007ACG} proposed a stable-LDS (SLDS) based method to incrementally add constraints to a relaxed system solution and to improve stability. To better adapt the standard LDS-based method to devices with limited memory and computational power, Abraham proposed a new DT synthesis with Fourier descriptors (FFT-LDS) \cite{Abraham2005DynamicTW}, which requires far fewer parameters compared to standard LDS approaches. In \cite{Chan2007ClassifyingVW}, Chain and Vasconcelos introduced a new method (Kernel-DT) for DT synthesis using kernel principal component analysis (KPCA) to learn a nonlinear observation function. However, the aforementioned DS modeling method faces the fundamental problem in that the column vector dimension of the unfolded frame is often too large compared to the number of given texture frames. To address this problem, the kernel principal component regression (KPCR) method was proposed for DT synthesis by You \cite{you2016kernel}.

	\subsection{Neural-Network-Based Methods}\label{sec2.2}
	Neural networks have proven to be immensely successful discriminative and generative learning machines \cite{Chen2019SemisupervisedFL,Saito2017TemporalGA,Xing2019UnsupervisedDO,Zhou2016LearningTT}. In terms of DT synthesis, various approaches based on ConvNet have been proposed \cite{Gatys2015TextureSU,tesfaldet2018two,xie2017synthesizing,Xie2021LearningES,Xie2019LearningDG}. In \cite{Gatys2015TextureSU}, Gatys et al. introduced a new model for DT synthesis based on the feature spaces of convolutional neural networks that represent texture using the correlations between feature maps in several layers. Motivated by the works on style transfer and enabled by the two-stream model, Tesfaldet et al. proposed a two-stream model for DT synthesis \cite{tesfaldet2018two}. This method represents the texture's appearance and dynamic nature using a set of Gram matrices. \cite{tulyakov2018mocogan} presented the motion and content decomposed generative adversarial network (MoCoGAN) for video generation. MoCoGAN is good at generating DTs when learning from a large amount of training data. In \cite{xie2017synthesizing,Xie2021LearningES}, Xie et al. proposed an energy-based spatial-temporal generative ConvNet to model and synthesize dynamic patterns. This model is beneficial for generating realistic dynamic patterns when the input sequences are incomplete with either occluded pixels or missing frames. \cite{Xie2019LearningDG} presented a dynamic generator model using an alternating back-propagation through time algorithm for DT synthesis. This model is efficient in terms of computational cost and model parameter size because it does not need to train a discriminative network or an inference network. Nevertheless, neural-network-based methods fail to exhibit powerful representations because DTs are high-dimensional data and with limited samples. Meanwhile, these methods are time-consuming and computationally expensive due to their large number of parameters. Therefore, an extreme learning machine may be the desired successor for DT synthesis, with good generalization performance and a high learning speed expected.

	\section{The Proposed DT Synthesis Method}\label{sec3}
	Similarity, which exists in different samples or self-samples, is as important prior knowledge as the class labels and other annotation information. Thus, some researchers have recently considered making use of the similarity knowledge of samples to improve the performance of the discriminative model for various tasks, i.e., person re-identification (re-ID) \cite{Cheng2016PersonRB,Fu2019SelfsimilarityGA,Hermans2017InDO}, and content-based image retrieval \cite{Lin2019TowardsOD,Liu2017CrossModalityBC}. DT exhibits statistical stationarity in the spatial domain and stochastic repetitiveness in the temporal dimension, indicating that different frames of DT possess high similarity correlations. These similarity correlations can also be viewed as critical prior knowledge, which may mitigate the high-dimensionality and small sample issues for DT synthesis. Therefore, we propose a novel DT synthesis method. Our model synthesizes desirable DTs at a high speed using kernel similarity embedding, based on ELM and kernel learning.

	In this section, we first revisit ELM to better understand the proposed method. We then illustrate our method, which uses ELM-based kernel similarity embedding. Meanwhile, we introduce the additive regularization factor for smoothing the kernel similarity embedding, which means our method is stabler with better generalization. Finally, we intuitively and theoretically analyze the mechanism of how and why our method can generate realistic, long-term DT videos.
	
	\subsection{Revisiting the Extreme Learning Machine}\label{sec3.1}
	ELM was initially proposed by Huang et al. \cite{Huang2004ExtremeLM}, serving as an emergent technology that has recently attracted much attention \cite{Huang2004ExtremeLM,Huang2012ExtremeLM,Chen2019DomainST,Gautam2021AdaptiveOL,Li2020OnlineAE,Lu2020RestrictedBoltzmannBasedEL,Yang2021GraphED,Deng2019ContentInsensitiveBI}. ELM works for generalized single-hidden layer feedforward networks (SLFNs). Its essence is that the hidden layer of SLFNs need not be tuned, which means that the feature mapping between the input layer and hidden layer is randomly assigned. With better generalization performance, ELM overcomes several challenges (e.g., slow learning speed, trivial human intervention, and poor computational scalability) faced by other computational intelligence techniques. Moreover, the parameters of the hidden layer are randomly initialized during training, and then the weights of the output layer are learned. Therefore, in this paper, we take full advantage of ELM to expedite the generation speed for DT synthesis, with good generalization.
	
	Before introducing ELM formally, we define some notations. Consider a dataset $D=\left\{\left(\mathbf{x}_{1}, \mathbf{y}_{1}\right), \cdots,\left(\mathbf{x}_{N}, \mathbf{y}_{N}\right)\right\}$, where $\mathbf{x}_{i} \in R^{n}$, $\mathbf{y}_{i} \in R^{m}$, $i=1,...,N$. The ELM with $T$ hidden nodes can be formulated as:
	\begin{gather}
	\label{eq1}
	f(x)=\sum_{t=1}^{T} \beta_{t} h_{t}(\mathbf{x})=\mathbf{h}(\mathbf{x}) \boldsymbol{\beta},
	\end{gather}
	where $\boldsymbol{\beta}=\left[\boldsymbol{\beta}^{1}, \ldots, \boldsymbol{\beta}^{T}\right]^{\mathrm{\top}}\in \mathbb{R}^{T\times m}$ ($T$ and $m$ are the number of nodes of the hidden layer and output layer, respectively) is the output weights between the hidden layer and the output layer, $\mathbf{h}(\mathbf{x})=\left[h_{1}(\mathbf{x}), \ldots, h_{T}(\mathbf{x})\right]$ is the output vector of the hidden layer with respect to the input $\mathbf{x}$. Intuitively, $\mathbf{h}(\mathbf{x})$ is a feature mapping, which maps the input $\mathbf{x}$ from $n$-dimensional input space to the $T$-dimensional hidden layer feature space (ELM feature space) $\boldsymbol{H}$.

	\begin{figure*}[t]
		\centering    
		\includegraphics[width=16cm,height=6cm]{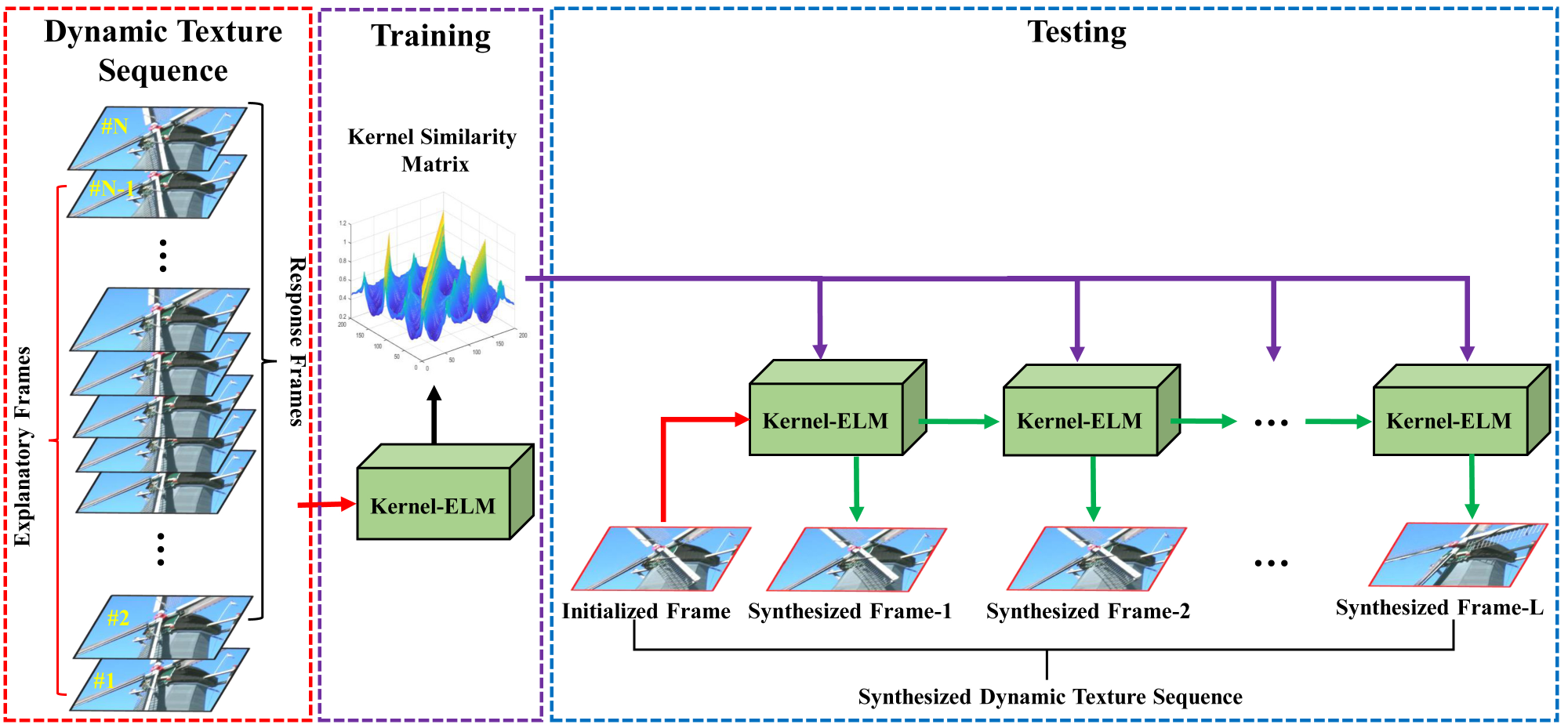}
		\caption{The architectural overview of our proposed DT synthesis method. First, we preprocess every input DT sequence, which is divided into two parts: explanatory frames and response frames. Then, we learn the kernel similarity matrix $\boldsymbol{\Omega}_{KSM}$ during training. Finally, the DT sequence is iteratively generated via the trained Kernel-ELM that embeds the learned $\boldsymbol{\Omega}_{KSM}$ into kernel similarity embedding for each new frame.}\label{fig:architecture}
	\end{figure*}

	According to Bartlett's theory \cite{Bartlett1996TheSC}, the smaller the norm of weights, the better the generalization performance of networks, and the smaller the training errors. That is, minimizing the norm of the output weights actually involves maximizing the distance between the separating margins of different domains in the feature space. Therefore, different from traditional intelligent learning algorithms, ELM aims to minimize the training errors and the norm of the output weights simultaneously. This is shown in Eq. (\ref{eq2}):
	\begin{gather}
	\label{eq2}
	Minimize : \mathcal{L}=\|\boldsymbol{H} \boldsymbol{\beta}-\boldsymbol{Y}\|^{2} + \|\boldsymbol{\beta}\|,
	\end{gather}
	where $\boldsymbol{H}$ is the output matrix of the hidden layer, shown in Eq. (\ref{eq3}).
	\begin{gather}
	\label{eq3}
	\boldsymbol{H}=\left[\begin{array}{c}{\mathbf{h}\left(\mathbf{x}_{1}\right)} \\ {\vdots} \\ {\mathbf{h}\left(\mathbf{x}_{N}\right)}\end{array}\right]=\left[\begin{array}{ccc}{h_{1}\left(\mathbf{x}_{1}\right)} & {\cdots} & {h_{T}\left(\mathbf{x}_{1}\right)} \\ {\vdots} & {\vdots} & {\vdots} \\ {h_{1}\left(\mathbf{x}_{N}\right)} & {\vdots} & {h_{T}\left(\mathbf{x}_{N}\right)}\end{array}\right].
	\end{gather}
	
	To solve Eq. (\ref{eq2}), the minimal norm least square method is typically used, and the solution is written as Eq. (\ref{eq4}):
	\begin{gather}
	\label{eq4}
	\boldsymbol{\beta}=\boldsymbol{H}^{\dagger} \boldsymbol{Y},
	\end{gather}
	where $\boldsymbol{H}^{\dagger}$ is the Moore-Penrose generalized inverse of $\boldsymbol{H}$, $\boldsymbol{Y}=\left[\mathbf{y}_{1}, \ldots, \mathbf{y}_{N}\right]^{\top}\in \mathbb{R}^{N\times m}$. Different methods can be used to calculate the Moore-Penrose generalized inverse of $\boldsymbol{H}$, e.g., orthogonalization, orthogonal projection, or singular value decomposition. Here we use the orthogonal projectional method, which has two forms: 1) if $\boldsymbol{H}^{\mathrm{\top}} \boldsymbol{H}$ is nonsingular, $\boldsymbol{H}^{\dagger}=\left(\boldsymbol{H}^{\mathrm{\top}} \boldsymbol{H}\right)^{-1} \boldsymbol{H}^{\mathrm{\top}}$, or 2) if $\boldsymbol{H} \boldsymbol{H}^{\mathrm{\top}}$ is nonsingular, $\boldsymbol{H}^{\dagger}=\boldsymbol{H}^{\mathrm{\dagger}}\left(\boldsymbol{H} \boldsymbol{H}^{\mathrm{\top}}\right)^{-1}$.

	Therefore, Eq. (\ref{eq4}) can be rewritten as Eq. (\ref{eq5}) or Eq. (\ref{eq6}):
	\begin{gather}
	\label{eq5}
	\boldsymbol{\beta}=\boldsymbol{H}^{\top}\left(\boldsymbol{H} \boldsymbol{H}^{\top}\right)^{-1} \boldsymbol{Y},
	\end{gather}
	\begin{gather}
	\label{eq6}
	\boldsymbol{\beta}=\left(\boldsymbol{H}^{\top} \boldsymbol{H}\right)^{-1} \boldsymbol{H}^{\top} \boldsymbol{Y}.
	\end{gather}
	Finally, the ELM can be written as Eq. (\ref{eq7})  or Eq. (\ref{eq8}):
	\begin{gather}
	\label{eq7}
	f(x)=\mathbf{h}(\mathbf{x}) \boldsymbol{\beta}=\mathbf{h}(\mathbf{x}) \boldsymbol{H}^{\top}\left(\boldsymbol{H} \boldsymbol{H}^{\top}\right)^{-1} \boldsymbol{Y},
	\end{gather}
	\begin{gather}
	\label{eq8}
	f(x)=\mathbf{h}(\mathbf{x}) \boldsymbol{\beta}=\mathbf{h}(\mathbf{x})\left(\boldsymbol{H}^{\top} \boldsymbol{H}\right)^{-1} \boldsymbol{H}^{\top} \boldsymbol{Y}.
	\end{gather}

	Note that the size of $\boldsymbol{H} \boldsymbol{H}^{\mathrm{\top}}$ is $N \times N$, while the size of $\boldsymbol{H}^{\mathrm{\top}} \boldsymbol{H}$ is $T \times T$. $N< T$ in the field of DT synthesis. From a practical point of view, we can obtain the solution for ELM based on Eq. (\ref{eq7}) as follows.

	\subsection{Kernel Similarity Embedding for DT Synthesis}\label{sec3.2}
	
	As discussed om Section \ref{sec3.1}, feature mapping $\mathbf{h}(\mathbf{x})$ is crucial for ELM. However, $\mathbf{h}(\mathbf{x})$ is known to the user and selected artificially, which is similar to the selection of the dimensionality-reduction function of physics-based methods and DS modeling methods for DT synthesis. Moreover, the number of nodes $T$ in the hidden layer of ELM is typically higher than the dimensions of the input data. This means that the feature mapping function $\mathbf{h}(\mathbf{x})$ explicitly maps samples to a high-dimensional space, which is equivalent to the original idea of the kernel function. Furthermore, the kernel function possesses the exciting property of being able to measure the similarity between different samples, which can elegantly exhibit the similarity correlation between different frames for DT. Following \cite{Huang2012ExtremeLM}, we extend ELM to kernel-ELM by kernel learning (kernel function: $K(\boldsymbol{u}, \boldsymbol{v})$) for DT synthesis. 
	
	The architectural overview of our method is shown in Figure \ref{fig:architecture}. Before training, all input DT sequences subtract their temporal mean, $\boldsymbol{S}_l \leftarrow \boldsymbol{S}_l-\overline{S}$, and the input video sequence is divided into two sub-sequences: explanatory frames and response frames. During training, we use kernel-ELM to learn the kernel similarity matrix for representing the statistical stationarity in the spatial domain and the stochastic repetitiveness in the temporal dimension of DT. The kernel similarity matrix is then further incorporated into the kernel similarity embedding for representation. During testing, high-fidelity long-term DT sequences are synthesized iteratively using the kernel similarity embedding.

	At first, we define a kernel similarity matrix $\Omega_{KSM}$:
	\begin{gather}
	\label{eq9}
	\boldsymbol{\Omega}_{KSM}=\boldsymbol{H} \boldsymbol{H}^{\top},
	\end{gather} 
	and
	\begin{gather}
	\label{eq10}
	\Omega_{KSM_{i, j}}=\mathbf{h}\left(\mathbf{x}_{i}\right) \cdot \mathbf{h}\left(\mathbf{x}_{j}\right)=K\left(\mathbf{x}_{i}, \mathbf{x}_{j}\right).
	\end{gather}
	In fact, kernel-ELM shares a similar network structure to ELM, optimizing the output weights $\boldsymbol{\beta}$ using a kernel function $K(u; v)$ to learn the kernel similarity embedding. Therefore, kernel-ELM is easier to learn than ELM, while maintaining the same merits. According to ridge regression theory \cite{Hoerl2000RidgeRB}, we can add a regularization factor $\lambda$ (positive small value) to control the regularization performance of $\|\boldsymbol{\beta}\|$ during optimization, which is dissimilar to \cite{Huang2012ExtremeLM}, which regularizes the term $\|\boldsymbol{H} \boldsymbol{\beta}-\boldsymbol{Y}\|^{2}$. If a proper $\lambda$ is used, the kernel similarity matrix will be smoother, and thus our method will be stabler and tend to have better generalization performance. Then, the optimization objective of our method can be formulated as:
	\begin{gather}
	\label{eq:appendix1}
	\begin{aligned}
	&Minimize : \mathcal{L}=\frac{1}{2}\lambda\|\boldsymbol{\beta}\|^{2}+\frac{1}{2} \sum_{i=1}^{N}\left\|\boldsymbol{\xi}_{i}\right\|^{2},
	\\&s.t. \quad \mathbf{h}\left(\mathbf{x}_{i}\right) \boldsymbol{\beta}-\mathbf{y}_{i}^{\mathrm{\top}}=\boldsymbol{\xi}_{i}^{\mathrm{\top}},
	\end{aligned}
	\end{gather}
	where $i=1, \ldots, N$ ($N$ is the number of training frames), and $\boldsymbol{\xi}_{i}=\left[\xi_{i, 1}, \ldots, \xi_{i, m}\right]^{\mathrm{\top}}$ is the training error vector of the training sample $x_{i}$. According to the Lagrange theorem, training our method is equivalent to solving the following optimization object:
	\begin{gather}
	\label{eq:appendix2}
	\begin{aligned} \mathcal{L}=\frac{1}{2}\lambda\|\boldsymbol{\beta}\|^{2} &+\frac{1}{2} \sum_{i=1}^{N}\left\|\boldsymbol{\xi}_{i}\right\|^{2} \\ &-\sum_{i=1}^{N} \sum_{j=1}^{m} \alpha_{i, j}\left(\mathbf{h}\left(\mathbf{x}_{i}\right) \boldsymbol{\beta}_{j}-y_{i, j}+\xi_{i, j}\right), 
	\end{aligned}
	\end{gather}
	where $\boldsymbol\beta_{j}$ is the vector of the weights that links the hidden layer to the $j$th output node and $\boldsymbol{\beta}=\left[\boldsymbol{\beta}_{1}, \ldots, \boldsymbol{\beta}_{m}\right]$, and $\alpha_{i, j}$ is the Lagrange multiplier corresponding to the $j$th output of the $i$th training sample. Then, we have the following KKT corresponding optimality conditions: 
	\begin{align}
	\label{eq:appendix3}
	&\frac{\partial L}{\partial \boldsymbol{\beta}_{j}} =0 \rightarrow \lambda \boldsymbol{\beta}_{j}=\sum_{i=1}^{N} \alpha_{i, j} \mathbf{h}\left(\mathbf{x}_{i}\right)^{\mathrm{\top}} \rightarrow \boldsymbol{\beta}=\frac{1}{\lambda}\boldsymbol{H}^{\mathrm{\top}} \boldsymbol{\alpha},\\
	\label{eq:appendix4}
	&\frac{\partial L}{\partial \boldsymbol{\xi}_{i}} =0 \rightarrow \boldsymbol{\alpha}_{i}= \boldsymbol{\xi}_{i},\\
	\label{eq:appendix5}
	&\frac{\partial L}{\partial \boldsymbol{\alpha}_{i}} =0 \rightarrow \mathbf{h}\left(\mathbf{x}_{i}\right) \boldsymbol{\beta}-\boldsymbol{y}_{i}^{\mathrm{\top}}+\boldsymbol{\xi}_{i}^{\mathrm{\top}}=0 ,
	\end{align}
	where $\boldsymbol{\alpha}_{i}=\left[\alpha_{i, 1}, \ldots, \alpha_{i, m}\right]^{\mathrm{\top}}$ and $\boldsymbol{\alpha}=\left[\boldsymbol{\alpha}_{1}, \ldots, \boldsymbol{\alpha}_{N}\right]^{\mathrm{\top}}$.

	We then substitute Eq. (\ref{eq:appendix3})  and Eq. (\ref{eq:appendix4}) into Eq. (\ref{eq:appendix5}), which can be written as:
	\begin{gather}
	\label{eq:appendix6}
	\left(\mathbf{I}+\frac{1}{\lambda}\boldsymbol{H} \boldsymbol{H}^{\mathrm{\top}}\right) \boldsymbol{\alpha}=\boldsymbol{Y} \rightarrow \boldsymbol{\alpha}=\left(\mathbf{I}+\frac{1}{\lambda}\boldsymbol{H} \boldsymbol{H}^{\mathrm{\top}}\right)^{-1}\boldsymbol{Y},
	\end{gather}
	where $\mathbf{I}$ is an identity matrix. Then, by combining Eq. (\ref{eq:appendix3}) and Eq. (\ref{eq:appendix6}), the output weights $\boldsymbol{\beta}$ of the hidden layer can be formulated as:
	\begin{gather}
	\label{eq:appendix7}
	\begin{aligned}
	\boldsymbol{\beta}&=\frac{1}{\lambda}\boldsymbol{H}^{\mathrm{\top}}\left(\mathbf{I}+\frac{1}{\lambda}\boldsymbol{H} \boldsymbol{H}^{\mathrm{\top}}\right)^{-1} \boldsymbol{Y},\\
	&=\boldsymbol{H}^{\mathrm{\top}}\left(\lambda\mathbf{I}+\boldsymbol{H} \boldsymbol{H}^{\mathrm{\top}}\right)^{-1} \boldsymbol{Y}.
	\end{aligned}
	\end{gather}
	
	Thus, the transition function of our method (output function of kernel-ELM) can be formulated as Eq. (\ref{eq:appendix8})) according to Eq. (\ref{eq1}) and Eq. (\ref{eq:appendix7}):
	\begin{gather}
	\label{eq:appendix8}
	f(x)=\mathbf{h}(\mathbf{x}) \boldsymbol{H}^{\top}\left(\lambda\mathbf{I}+\boldsymbol{H} \boldsymbol{H}^{\top}\right)^{-1} \boldsymbol{Y}.
	\end{gather}
	By Combining Eq. (\ref{Eq1}) and Eq. (\ref{eq:appendix8}), the transition function of our method can be rewritten as Eq. (\ref{eq11}):
	\begin{gather}
	\label{eq11}
	f(x) =\left[\begin{array}{c}{K\left(\mathbf{x}, \mathbf{x}_{1}\right)} \\ {\vdots} \\ {K\left(\mathbf{x}, \mathbf{x}_{N}\right)}\end{array}\right]^{\top}\left(\lambda\mathbf{I}+\boldsymbol{\Omega}_{KSM}\right)^{-1} \boldsymbol{Y}.
	\end{gather}
	Here, $\mathbf{x}$ is the test frame, $[\mathbf{x}_{1},...,\mathbf{x}_{N}]$ is with respect to the elements of the explanatory frames, and $\boldsymbol{Y}$ is with respect to the response frames. From Eq. (\ref{eq11}), we find that the proposed model is only related to the kernel function, input data $\mathbf{x}_{i}$ and number of training samples. The kernel similarity embedding is not related to the number of outputs nodes. Thus, our method does not need to artificially select $\mathbf{h}(\mathbf{x})$ and implicitly maps the input data to a high-dimensional space. Furthermore, the feature mapping $\mathbf{h}(\mathbf{x})$ and dimensionality of the feature space (nodes of the hidden layer) are unknown to the user; instead, the corresponding kernel $K(\boldsymbol{u}, \boldsymbol{v})$ is given. Moreover, Eq. (\ref{eq11}) intuitively shows that the kernel similarity matrix is incorporated into the kernel similarity embedding, which will effectively use the similarity as prior information for representing DTs.

	\begin{algorithm}
		\caption{Kernel Similarity Embedding for DT Synthesis}
		\label{Similarity-DT}
		\begin{algorithmic}[1]
			
			\REQUIRE ~~\\
			(1) Training video sequences $\{\boldsymbol{S}_{l}, l=1,\cdots, N\}$ \\
			(2) Number of synthesized image sequences $L$\\
			(3) Kernel function $K(\boldsymbol{u}, \boldsymbol{v})$\\
			
			\ENSURE~~\\
			(1) Synthesized image sequences $\{ \boldsymbol{\tilde{S}}_{l}, l=1, \cdots,L\}$
			
			\item[]
			\STATE Calculate the temporal mean $\overline{S}$ of $\boldsymbol{S}_{t}$.
			\STATE Let $\boldsymbol{S}_l \leftarrow \boldsymbol{S}_l-\overline{S}$, $l=1, \cdots, N$.
			\STATE Initialize $\{ \boldsymbol{\tilde{S}}_{l}\}$, for $l=1, \cdots,L$.
			\STATE Define explanatory frames $\{\boldsymbol{S}_{j}, j=1,\cdots, N-1\}$ and response frames $\{\boldsymbol{S}_{k}, k=2,\cdots, N\}$ using training video sequences.
			
			\STATE Calculate $\boldsymbol{\Omega}_{KSM}$ and $\left(\lambda \mathbf{I}+\boldsymbol{\Omega}_{KSM}\right)^{-1} \boldsymbol{Y}$ according to Eq. (\ref{eq9}) and Eq. (\ref{eq10}).
			
			\REPEAT 
			
			\STATE Calculate  $ \boldsymbol{\tilde{S}}_{l}=f(\boldsymbol{\tilde{S}}_{l-1})$ by Eq. (\ref{eq11}), $l>1$.
			\STATE Let $\boldsymbol{\tilde{S}_{l}} \leftarrow \boldsymbol{\tilde{S}}_l+\overline{S}$. 
			\STATE Let $l \leftarrow l+1$
			\UNTIL $l = L$
		\end{algorithmic}
	\end{algorithm}

	See Algorithm \ref{Similarity-DT} for a description of the proposed DT synthesis method. Specifically, the algorithm first divides the input video sequence $\boldsymbol{S}_{1 : N}$ (after  subtracting the temporal mean $\overline{S}$) into two sub-sequences: explanatory frames $\{\boldsymbol{S}_{j}, j=1,\cdots, N-1\}$ and response frames $\{\boldsymbol{S}_{k}, k=2,\cdots, N\}$. Then, the kernel similarity matrix $\Omega_{KSM}$ is learned with respect to Eq. (\ref{eq10}) and Eq. (\ref{eq9}), where it is incorporated into kernel similarity embedding for representing DT. Finally,  the endless sequences $\{ \boldsymbol{\tilde{S}}_{l}, l=1, \cdots,L\}$ (after adding the temporal mean $\overline{S}$) can be generated iteratively with a pre-trained model according to Eq. (\ref{eq11}). 
	
	In fact, the dimensionality $D$ of the explanatory frames and response frames is equal, and thus $n=m=D$. During training, the computational complexity of our method is $\boldsymbol{O}\left(D^{2} N^{2}\right)$, including an $N \times N$ kernel operation, an inverse operation, and matrix multiplication. During testing, the computational complexity of our method is $\boldsymbol{O}\left(DN\right)$, including $N$ kernel operations and matrix multiplication.

	\begin{figure}[t]
		\begin{center}
			\includegraphics[width=.16\linewidth]{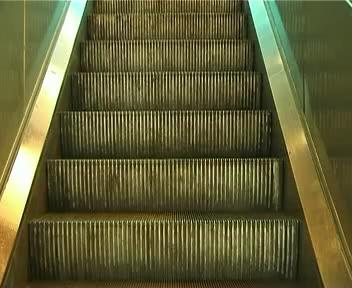}
			\includegraphics[width=.23\linewidth]{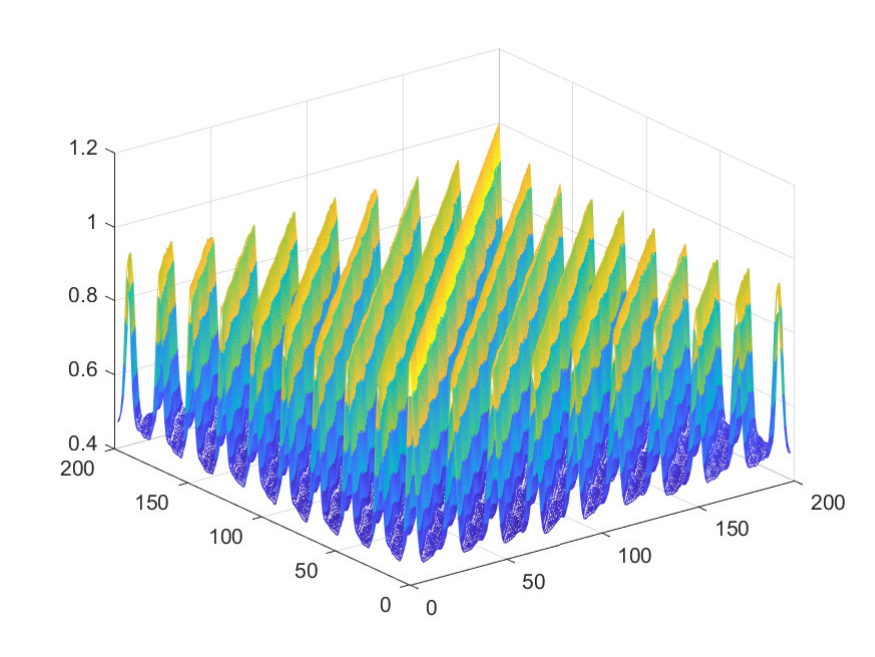}
			\includegraphics[width=.23\linewidth]{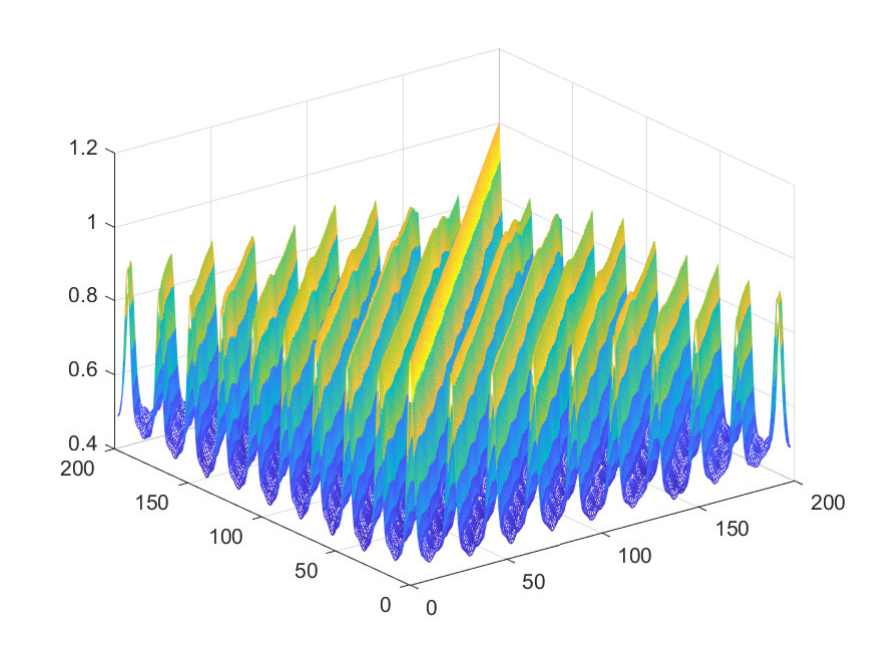}
			\includegraphics[width=.23\linewidth]{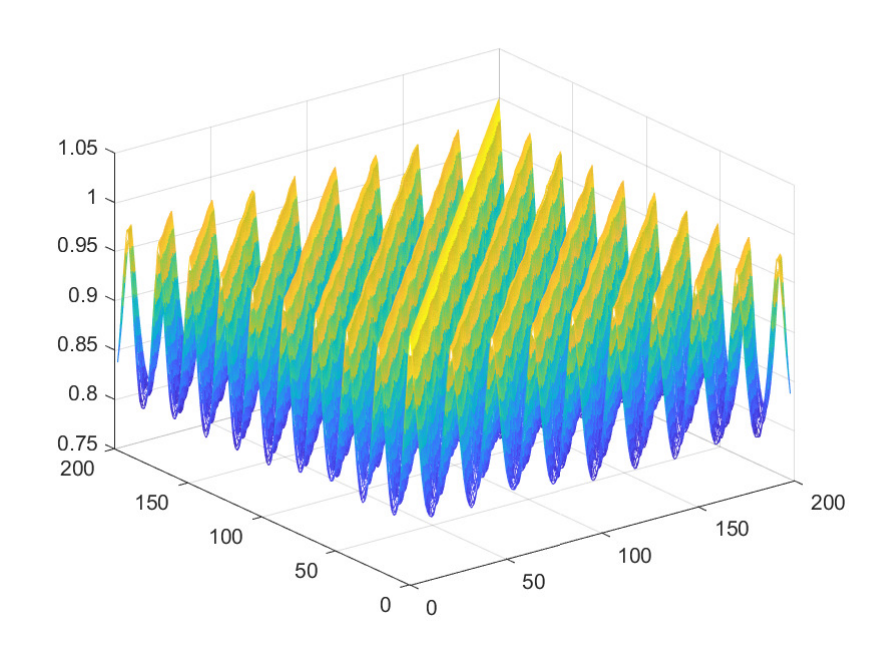}
			\\ (a) Elevator\\
			
			\includegraphics[width=.16\linewidth]{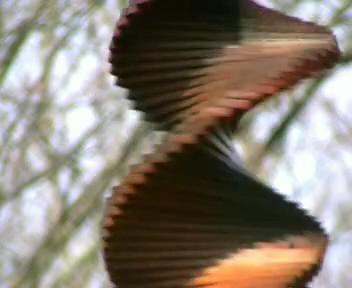}
			\includegraphics[width=.23\linewidth]{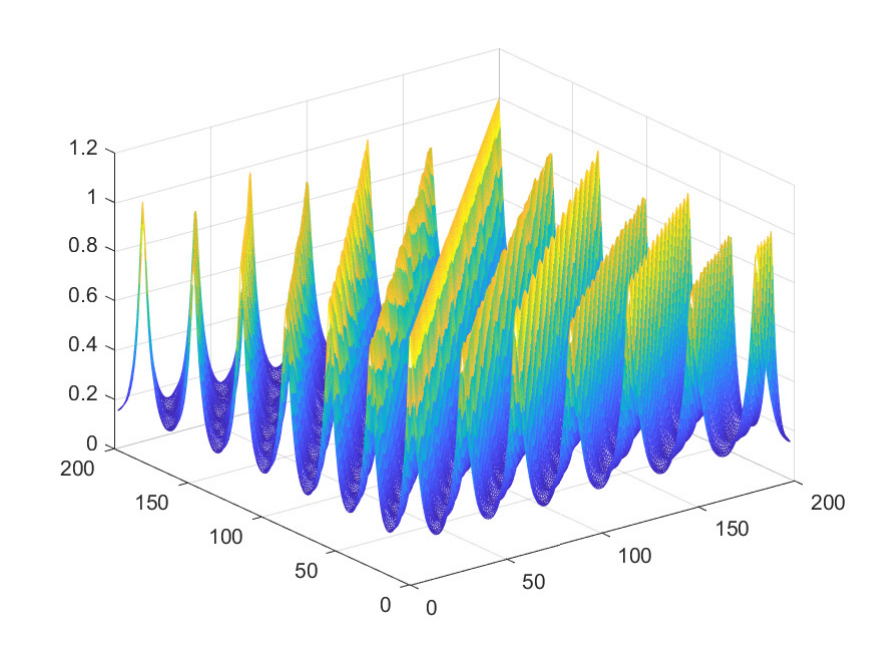}
			\includegraphics[width=.23\linewidth]{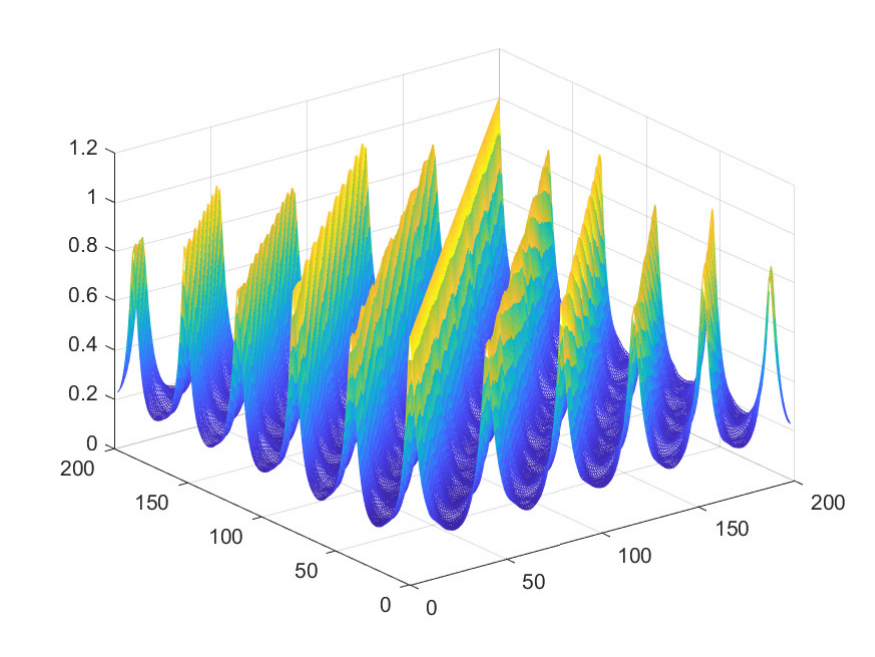}
			\includegraphics[width=.23\linewidth]{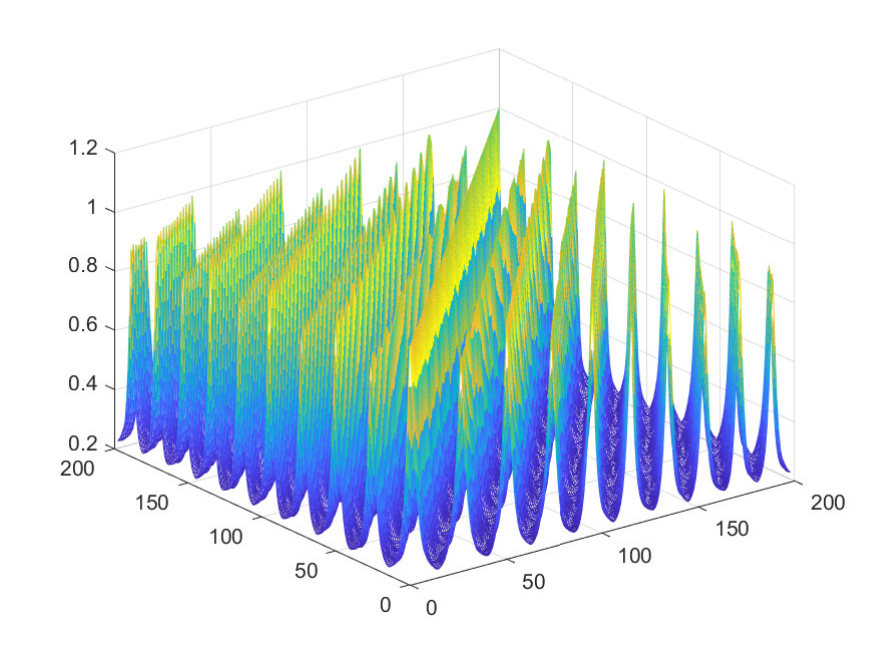}
			\\ (b) Rotating wind ornament\\
			
			\includegraphics[width=.16\linewidth]{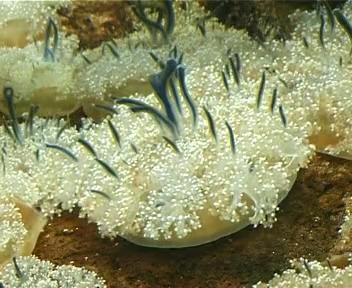}
			\includegraphics[width=.23\linewidth]{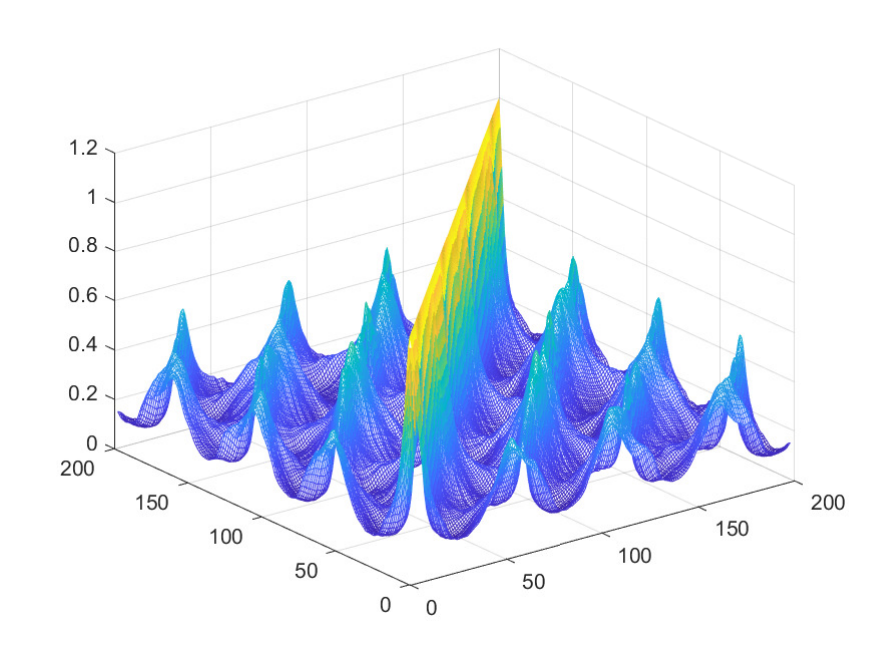}
			\includegraphics[width=.23\linewidth]{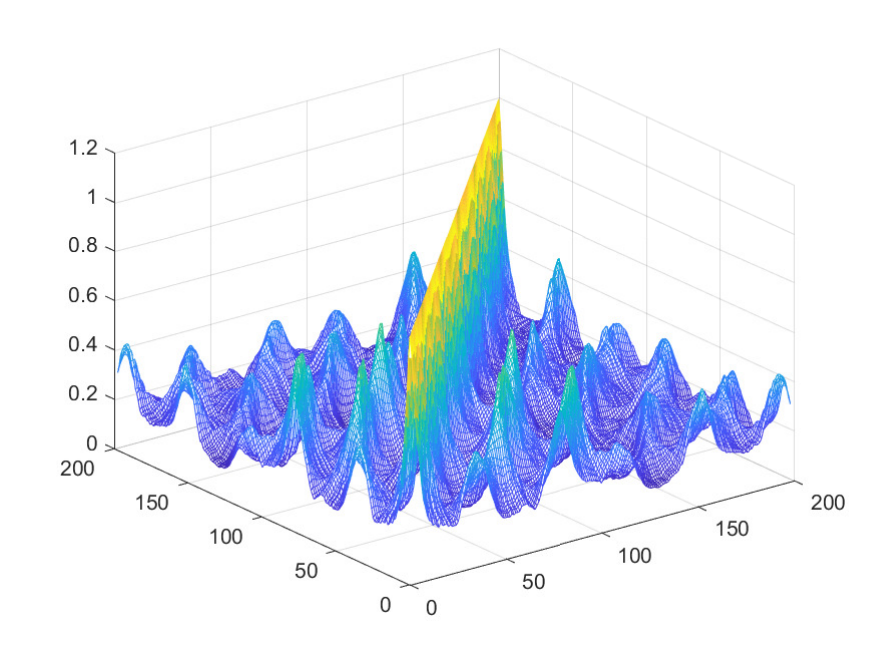}
			\includegraphics[width=.23\linewidth]{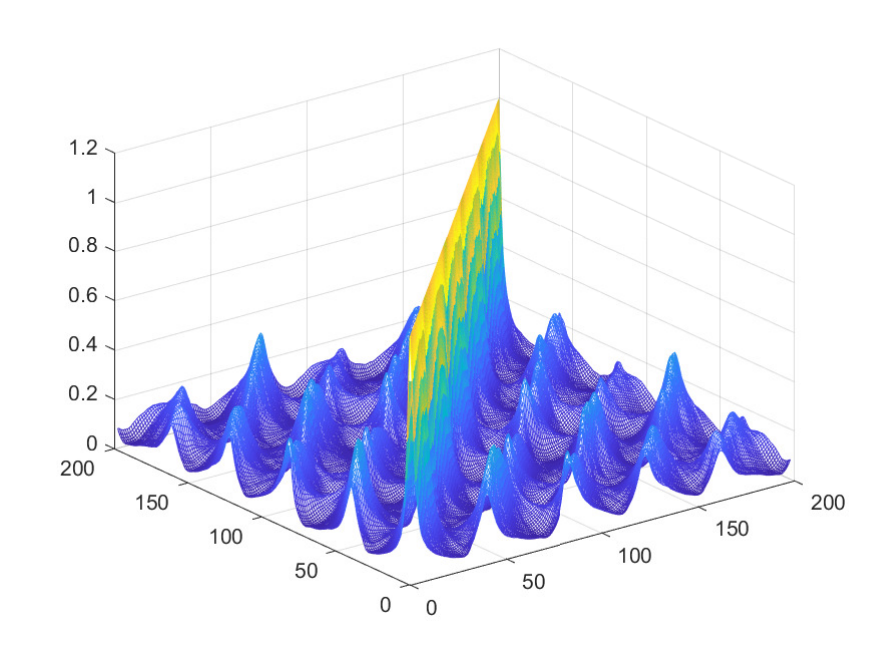}
			\\ (c) Flowers swaying with current\\
			
			\includegraphics[width=.16\linewidth]{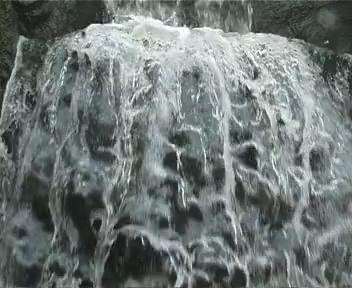}
			\includegraphics[width=.23\linewidth]{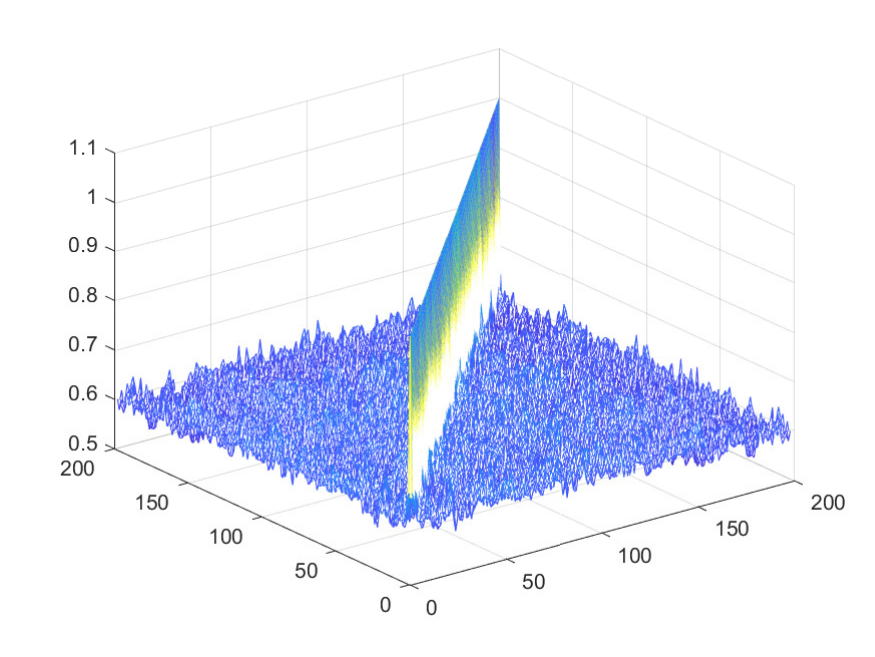}
			\includegraphics[width=.23\linewidth]{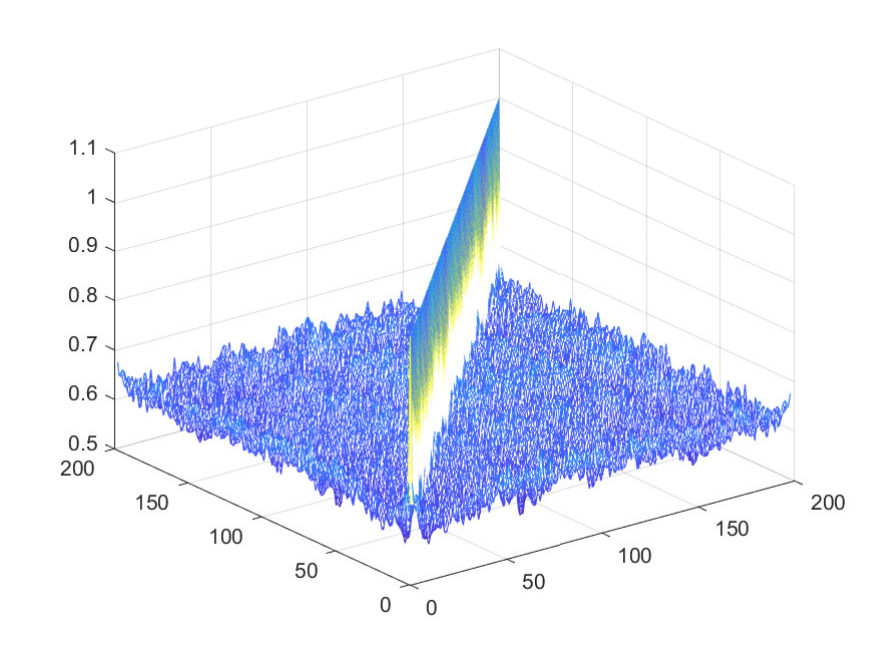}
			\includegraphics[width=.23\linewidth]{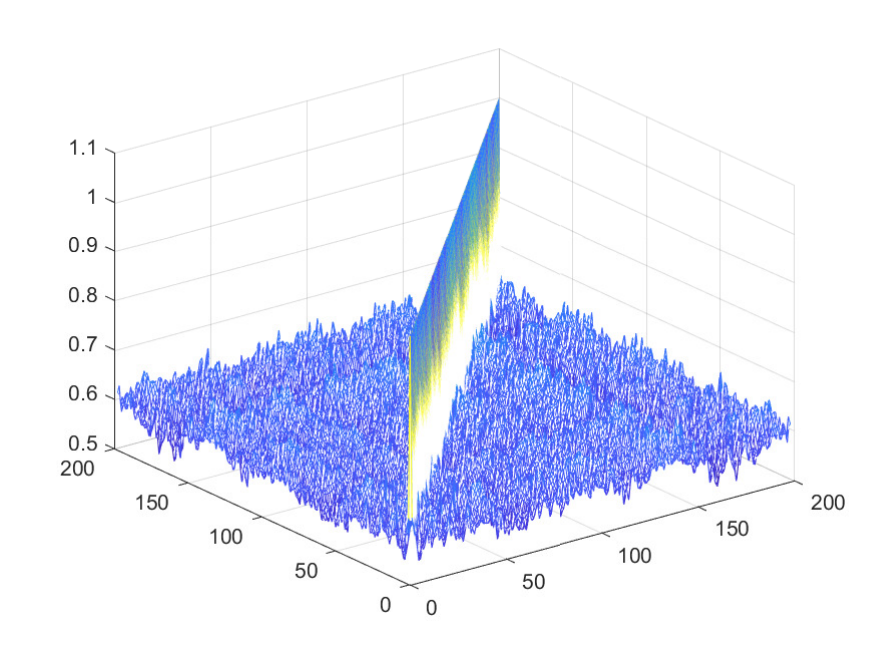}
			\\ (d) Waterfall\\
			
			\includegraphics[width=.16\linewidth]{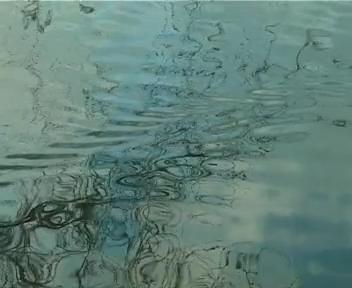}
			\includegraphics[width=.23\linewidth]{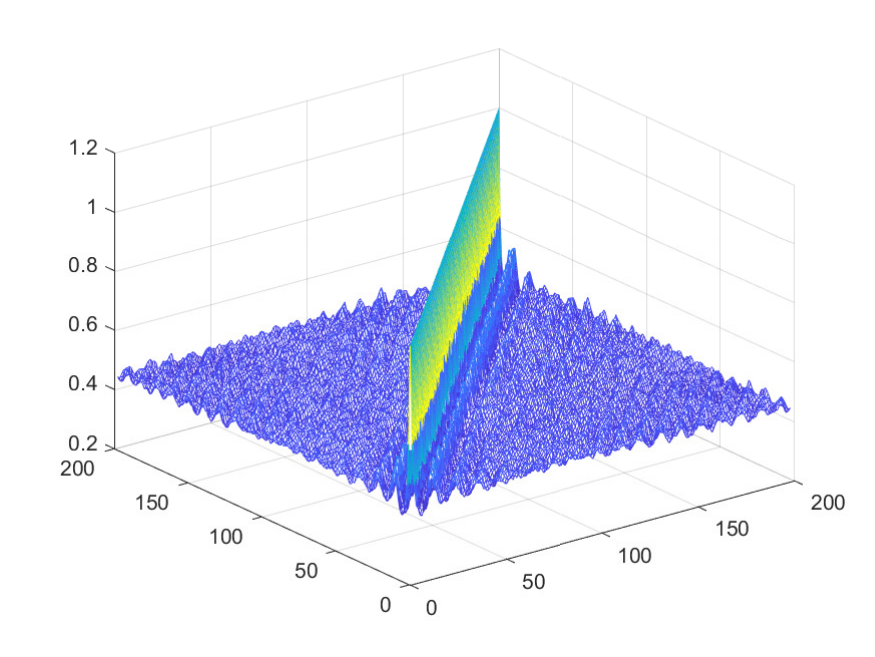}
			\includegraphics[width=.23\linewidth]{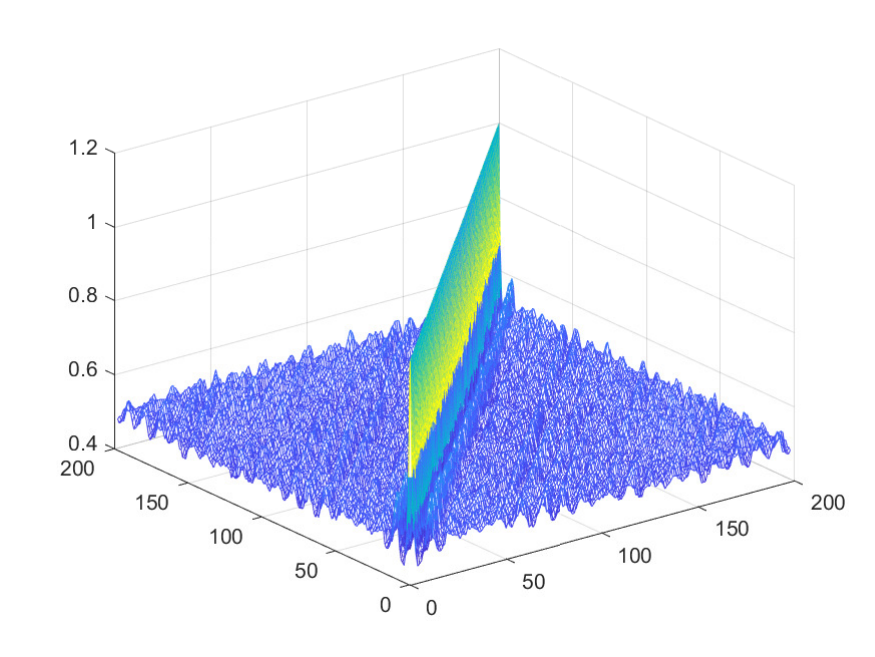}
			\includegraphics[width=.23\linewidth]{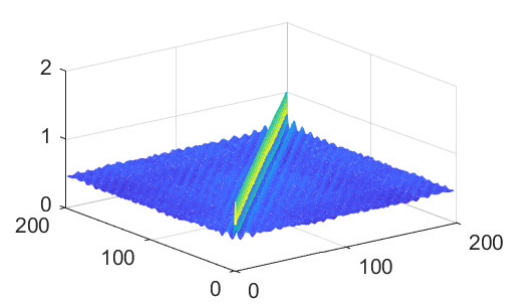}
			\\ (e) Water wave \\
			
			\includegraphics[width=.16\linewidth]{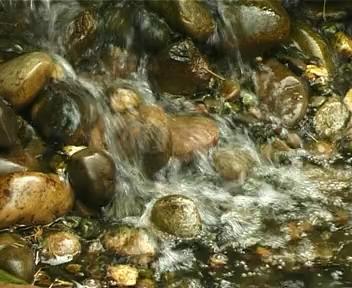}
			\includegraphics[width=.23\linewidth]{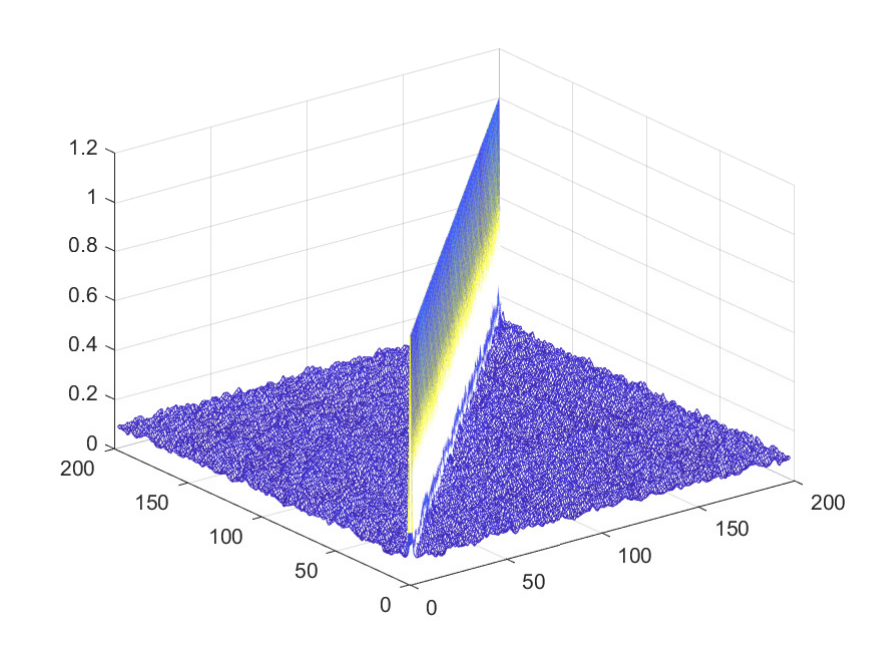}
			\includegraphics[width=.23\linewidth]{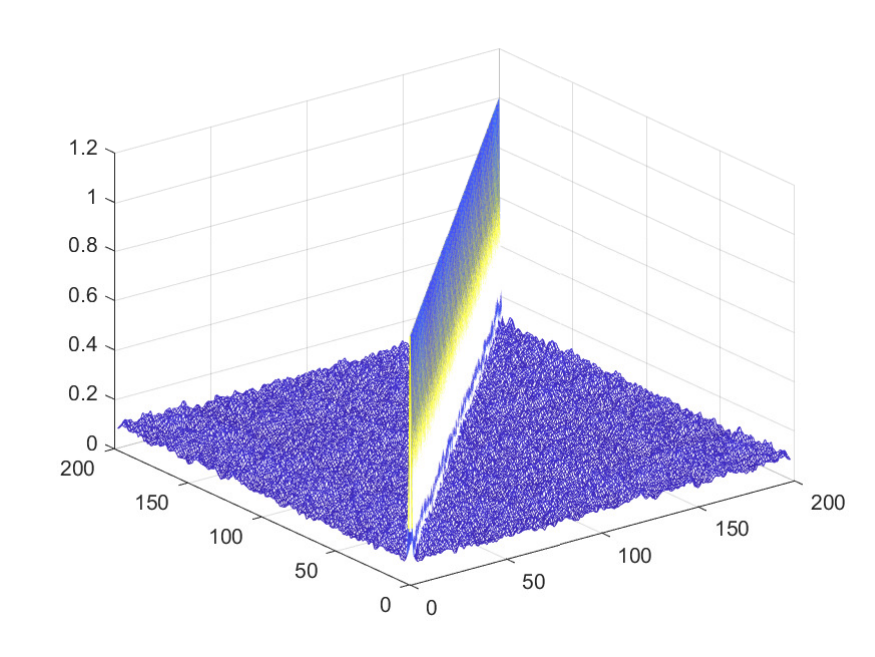}
			\includegraphics[width=.23\linewidth]{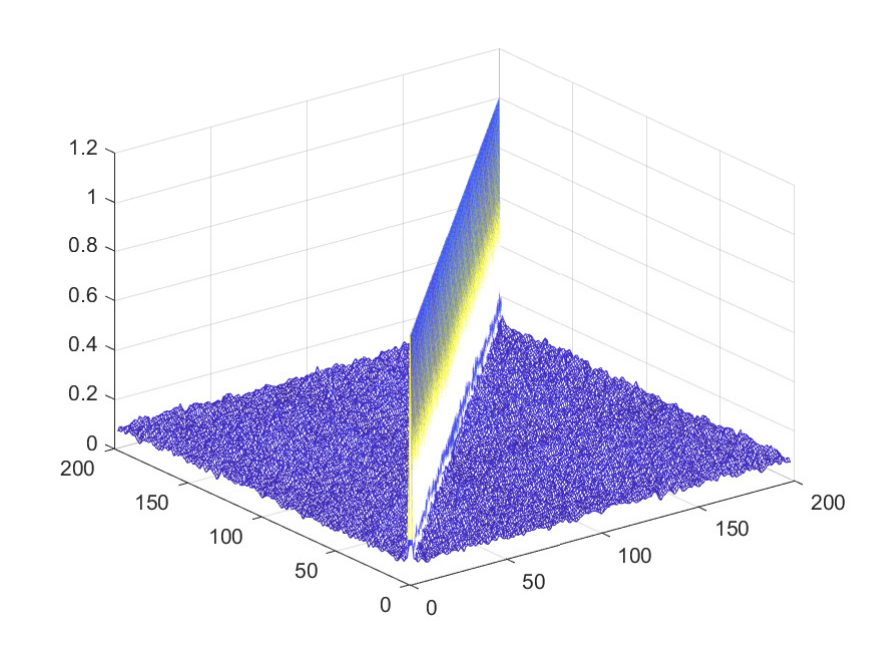}
			\\(f) Spring water

			\caption{Visualization of kernel similarity matrices of DTs are displayed. For each row, the first image is the frame of a DT video, and the other three are the kernel similarity matrices learned from different DT sequences of the same class.}
			\label{fig:kernel_embedding}
		\end{center}
	\end{figure}

	\subsection{Analysis of Kernel Similarity Embedding}\label{sec3.3}
	\subsubsection{Intuitive Insight}\label{sec3.3.1}
	DT videos exhibit statistical stationarity in the spatial domain and stochastic repetitiveness in the temporal dimension, which is the key cue for distinguishing DT videos from other videos and static images \cite{doretto2003dynamic,Gatys2015TextureSU,tesfaldet2018two}. Moreover, this cue can be further elaborated as the similarity correlation between frames, as shown in Figure \ref{fig:similarity-embedding}. Therefore, what we need to do for DT synthesis is to build a DT model for representing the features of the dynamic and texture elements, which are statistically similar and temporally stationary. Here, we integrate kernel learning and ELM into a powerfully unified DT synthesis model to learn kernel similarity embeddings for achieving this goal.

	Our method represents such features with a kernel similarity matrix, which is incorporated into the kernel similarity embedding. To intuitively analyze this mechanism, we visualize the learned kernel similarity matrices of some DT sequences (200 frames for each sequence) in the Dyntex dataset after training, as shown in Figure \ref{fig:kernel_embedding}. The kernel similarity matrices of each row in Figure \ref{fig:kernel_embedding} are learned from different DT sequences of the same class, e.g., elevator, waterfall, rotating wind ornament, flowers swaying with current, water wave, and spring water, which exist in everyday surroundings. The different DT sequences of the same class are acquired with different views or different periods. As can be seen from Figure \ref{fig:kernel_embedding}, the kernel similarity matrices produced by our method elegantly represent the similarity correlations of DT videos. Specifically, the repetitiveness and stationarity of the DT of the elevator, rotating wind ornament, and flowers swaying with current are clearly exhibited by the learned matrices. As for the waterfall, water wave, and spring water, although these objects do not originally have obvious repetitiveness influenced by natural factors, the kernel similarity matrices consistently exhibit the statistical stationarity and similarity for different DT videos of the same class. To this end, kernel similarity embedding effectively mines and exploits the similarity as prior knowledge for representing DT using kernel similarity matrices, overcoming the challenge of high-dimensionality and small sample issues. That is, the feature representations produced by our method are discriminative, enabling the proposed method to generate high-fidelity, long-term DT videos. Additionally, these results also show that our hypotheses are not limited to frames at several time points in the same time interval more similar.
	
	\subsubsection{Theoretical Insight}\label{sec3.3.2}
	Kernel similarity embedding for DT synthesis
	can be viewed as the kernel embedding of the conditional distribution for
	regression problems, where the feature vector $\Phi=\left[\phi\left(\mathbf{y}_{1}\right), \ldots, \phi\left(\mathbf{y}_{N}\right)\right]^{\top}$ ($\Phi$ is a mapping function) in the reproducing kernel Hilbert space (RKHS) is substituted by $Y=[\mathbf{y}_{1}, \dots, \mathbf{y}_{N}]^{\top}$ in the original data domain. The key idea is to map conditional distributions into infinite-dimensional feature spaces using kernels, such that all the statistical features of arbitrary distributions and high-dimensional data can ultimately be captured \cite{Song2009HilbertSE,Song2013KernelEO}. Its formulation is as follows:
	\begin{gather}
	\label{eq23}
	\begin{aligned} \widehat{\mu}_{\boldsymbol{Y} | \mathbf{x}}=\sum_{i=1}^{N} \phi\left(\mathbf{y}_{i}\right) W_{i}(\mathbf{x}) &=\boldsymbol{W}(\mathbf{x}) \Phi \\ &=K_{: \mathbf{x}}(\boldsymbol{G}+\lambda \boldsymbol{I})^{-1} \Phi,
	\end{aligned}
	\end{gather}
	where $K_{: \mathbf{x}}=\left[k\left(\mathbf{x}, \mathbf{x}_{1}\right), \ldots, k\left(\mathbf{x}, \mathbf{x}_{N}\right)\right]$, $\boldsymbol{G}$ is the Gram matrix for samples from variable $\boldsymbol{X}$, and $\boldsymbol{W}(x)=\left[W_{1}, \ldots, W_{N}\right]^{\top}$ is a non-uniform weight vector determined by the value $\mathbf{x}$ of the conditioning variable. Indeed, this non-uniform weight vector reflects the effects of conditioning on the embedding. As for the kernel similarity embedding, it can be rewritten as Eq. (\ref{eq24}) according to Eq. (\ref{eq11}):
	\begin{gather}
	\label{eq24}
	\begin{aligned}
	f(x) &=\left[\begin{array}{c}{K\left(\mathbf{x}, \mathbf{x}_{1}\right)} \\ {\vdots} \\ {K\left(\mathbf{x}, \mathbf{x}_{N}\right)}\end{array}\right]^{\top}\left(\lambda\mathbf{I}+\boldsymbol{\Omega}_{KSM}\right)^{-1} \boldsymbol{Y}\\
	&=\boldsymbol{W}(\mathbf{x})\boldsymbol{Y}.
	\end{aligned}
	\end{gather}
	
	It is obvious that the synthesized frames are conditioned by training sample $\mathbf{x}$. When comparing Eq. (\ref{eq23}) and Eq. (\ref{eq24}), we observe that the kernel similarity embedding for DT is similar to the kernel embedding for the conditional distribution. The difference is that the kernel similarity embedding for DT synthesis predicts the future frames $\mathbf{y}_i$ in the original data domain, while the kernel embedding for the conditional distribution predicts the feature $\phi(\mathbf{y}_i)$ in RKHS. To this end, kernel similarity embedding possesses the properties of kernel embedding for conditional distributions. Thus, it will effectively represent the statistical features (e.g., similarity correlation representation) by modeling nonlinear feature relationships for DT.
	
	\section{Experiments AND Evaluation}\label{sec4}
	In the following sections, we provide the implementation details and parameter settings. Furthermore, we intuitively analyze the sustainability and generalization of our method. Finally, we demonstrate, by visual evaluation, time consumption, and quantitative evaluation, that our method is superior to nine baseline methods, including both non-neural-network-based and neural-network-based DT synthesis methods.

	\subsection{Implementation Details}\label{sec4.1}
	
	For the following experiments, the DT videos were collected from the internet and two benchmark datasets, i.e., Gatech Graphcut Textures\footnote{http://www.cc.gatech.edu/cpl/projects/graphcuttextures} \cite{Kwatra2003GraphcutTI} and Dyntex\footnote{http://projects.cwi.nl/dyntex/database.html} \cite{Pteri2010DynTexAC}. These two benchmark datasets are publicly available and have been widely used in recent publications \cite{doretto2003dynamic,Gatys2015TextureSU,you2016kernel,xie2017synthesizing,tesfaldet2018two,Xie2021LearningES,Costantini2008HigherOS,Chan2007ClassifyingVW,Abraham2005DynamicTW,Siddiqi2007ACG}. We resize the frame size of all DT videos to $150\times100$ pixels, following to \cite{you2016kernel} for direct comparison. Moreover, we train our model using the first 59 to 200 frames because the length of the shortest observed DT sequence is 59. We then synthesize a new frame with the long-term frames for quantitative evaluation. All the experiments presented in this paper are conducted in MATLAB 2018b under Windows 10 with a 64-bit operating system.
	
	\begin{figure}[t]
		\begin{center}
			\includegraphics[width=.07\linewidth]{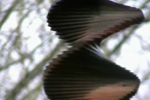}
			\includegraphics[width=.07\linewidth]{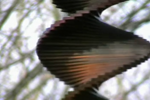}
			\includegraphics[width=.07\linewidth]{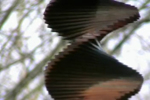}
			\includegraphics[width=.07\linewidth]{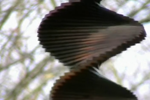}
			\includegraphics[width=.07\linewidth]{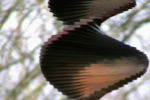}
			\includegraphics[width=.07\linewidth]{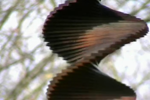}
			\hspace{1mm}
			\includegraphics[width=.07\linewidth]{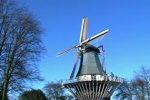}
			\includegraphics[width=.07\linewidth]{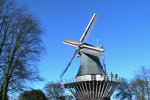}
			\includegraphics[width=.07\linewidth]{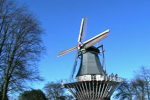}
			\includegraphics[width=.07\linewidth]{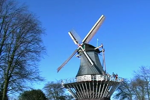}
			\includegraphics[width=.07\linewidth]{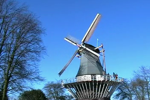}
			\includegraphics[width=.07\linewidth]{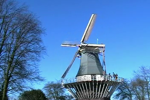} \\
			\vspace{1mm}
			
			\includegraphics[width=.07\linewidth]{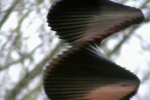}
			\includegraphics[width=.07\linewidth]{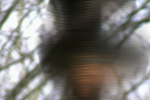}
			\includegraphics[width=.07\linewidth]{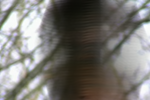}
			\includegraphics[width=.07\linewidth]{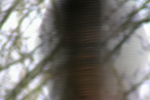}
			\includegraphics[width=.07\linewidth]{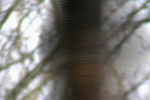}
			\includegraphics[width=.07\linewidth]{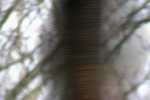}
			\hspace{1mm}
			\includegraphics[width=.07\linewidth]{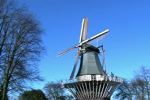}
			\includegraphics[width=.07\linewidth]{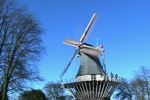}
			\includegraphics[width=.07\linewidth]{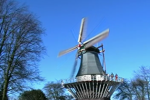}
			\includegraphics[width=.07\linewidth]{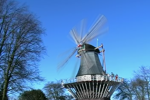}
			\includegraphics[width=.07\linewidth]{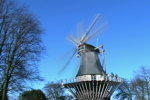}
			\includegraphics[width=.07\linewidth]{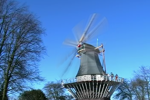} \\
			\vspace{1mm}
			
			\includegraphics[width=.07\linewidth]{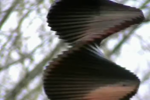}
			\includegraphics[width=.07\linewidth]{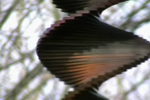}
			\includegraphics[width=.07\linewidth]{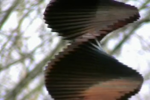}
			\includegraphics[width=.07\linewidth]{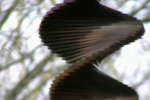}
			\includegraphics[width=.07\linewidth]{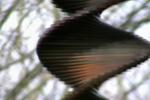}
			\includegraphics[width=.07\linewidth]{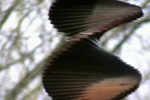}
			\hspace{1mm}
			\includegraphics[width=.07\linewidth]{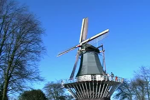}
			\includegraphics[width=.07\linewidth]{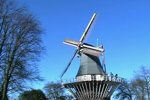}
			\includegraphics[width=.07\linewidth]{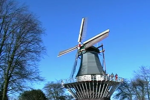}
			\includegraphics[width=.07\linewidth]{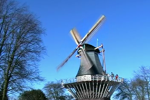}
			\includegraphics[width=.07\linewidth]{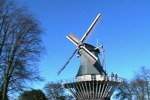}
			\includegraphics[width=.07\linewidth]{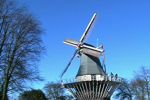} \\
			\vspace{1mm}
			
			\includegraphics[width=.07\linewidth]{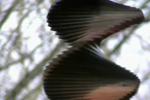}
			\includegraphics[width=.07\linewidth]{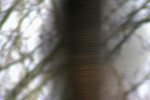}
			\includegraphics[width=.07\linewidth]{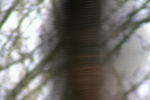}
			\includegraphics[width=.07\linewidth]{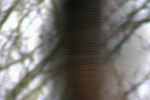}
			\includegraphics[width=.07\linewidth]{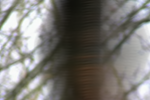}
			\includegraphics[width=.07\linewidth]{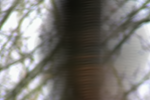}
			\hspace{1mm}
			\includegraphics[width=.07\linewidth]{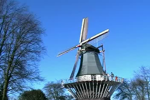}
			\includegraphics[width=.07\linewidth]{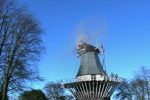}
			\includegraphics[width=.07\linewidth]{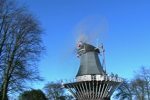}
			\includegraphics[width=.07\linewidth]{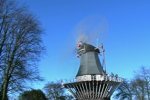}
			\includegraphics[width=.07\linewidth]{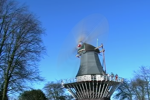}
			\includegraphics[width=.07\linewidth]{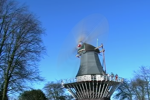} \\
			\vspace{1mm}
			
			\includegraphics[width=.07\linewidth]{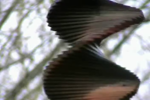}
			\includegraphics[width=.07\linewidth]{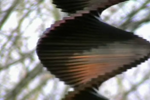}
			\includegraphics[width=.07\linewidth]{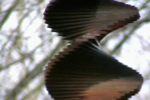}
			\includegraphics[width=.07\linewidth]{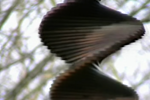}
			\includegraphics[width=.07\linewidth]{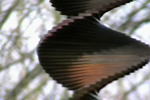}
			\includegraphics[width=.07\linewidth]{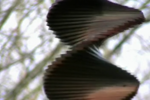}
			\hspace{1mm}
			\includegraphics[width=.07\linewidth]{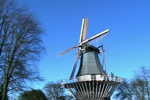}
			\includegraphics[width=.07\linewidth]{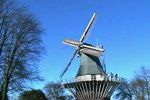}
			\includegraphics[width=.07\linewidth]{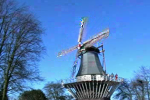}
			\includegraphics[width=.07\linewidth]{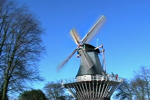}
			\includegraphics[width=.07\linewidth]{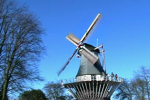}
			\includegraphics[width=.07\linewidth]{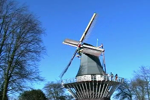} \\
			\vspace{1mm}
			
			\includegraphics[width=.07\linewidth]{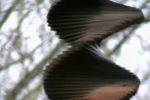}
			\includegraphics[width=.07\linewidth]{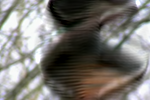}
			\includegraphics[width=.07\linewidth]{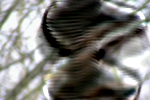}
			\includegraphics[width=.07\linewidth]{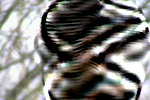}
			\includegraphics[width=.07\linewidth]{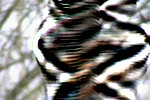}
			\includegraphics[width=.07\linewidth]{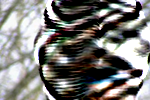}
			\hspace{1mm}
			\includegraphics[width=.07\linewidth]{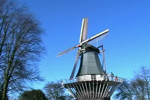}
			\includegraphics[width=.07\linewidth]{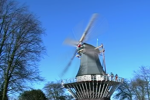}
			\includegraphics[width=.07\linewidth]{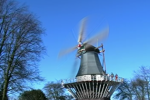}
			\includegraphics[width=.07\linewidth]{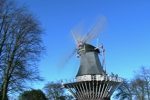}
			\includegraphics[width=.07\linewidth]{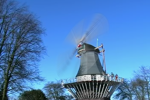}
			\includegraphics[width=.07\linewidth]{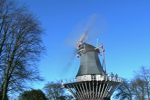} \\
			\vspace{1mm}

			\includegraphics[width=.07\linewidth]{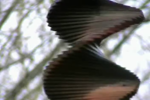}
			\includegraphics[width=.07\linewidth]{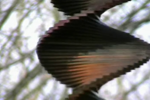}
			\includegraphics[width=.07\linewidth]{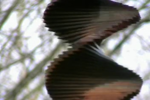}
			\includegraphics[width=.07\linewidth]{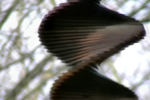}
			\includegraphics[width=.07\linewidth]{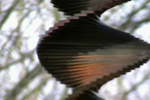}
			\includegraphics[width=.07\linewidth]{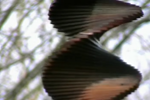}
			\hspace{1mm}
			\includegraphics[width=.07\linewidth]{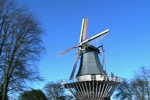}
			\includegraphics[width=.07\linewidth]{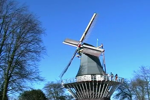}
			\includegraphics[width=.07\linewidth]{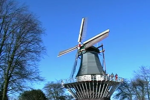}
			\includegraphics[width=.07\linewidth]{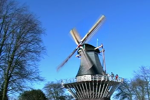}
			\includegraphics[width=.07\linewidth]{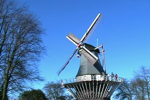}
			\includegraphics[width=.07\linewidth]{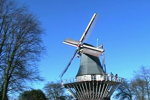} 
			\\ 
			
			\caption{Performance comparisons between different kernel functions used in our method for the videos ``rotating wind ornament" (left) and ``windmill" (right). For each category, the first row displays 6 frames of the observed sequence, and the other rows display the corresponding frames (left-to-right: 1-th, 100-th, 150-th, 210-th, 230-th, 250-th) of synthesized sequences generated by our method using different kernel functions (top-to-bottom: Linear kernel, Rational Quadratic kernel, Polynomial kernel, Multiquadric kernel, Sigmoid kernel, Gaussian kernel). }
			\label{fig:different-kernel-images}
		\end{center}
	\end{figure}
	
	In addition, we use two metrics to quantitatively evaluate the
	performance of the proposed method, including peak signal-noise ratio (PSNR) \cite{Wang2009MeanSE} and structural similarity (SSIM) \cite{wang2004image}. These are common quantitative evaluation metrics in static image generation \cite{chen2017photographic,Song2020MultimodalIS,Marivani2020MultimodalDU,Liu2019DepthSV}, DT synthesis \cite{Xie2020MotionBasedGM,xie2017synthesizing,Xie2021LearningES} and other future frame prediction problems \cite{tulyakov2018mocogan,Castrejn2019ImprovedCV}. Formally, PSNR can be written as Eq. (\ref{eq12}):
	
	\begin{gather}
	\label{eq12}
	PSNR=\frac{1}{L-1} \sum_{t=2}^{L} 10 \log _{10} \frac{255^{2}}{M S E\left(\widehat{S}_{l}-S_{l}\right)},
	\end{gather}
	where $L$ is the length of the observed sequence, and $S_{l}$ ($l=2, 3,\cdots, L$) and $\widehat{S}_{l}$ ($l=2, 3,\cdots, L$) are the observed video frames and generated video frames, respectively. Intuitively, PNSR is presented with the prediction error between the observed sequence and generated sequence. The higher the PSNR, the better the high-fidelity DT video generated.
	
	SSIM was originally designed for image quality assessment, and later used to provide a perceptual judgment on similarity between videos. It can be formulated as Eq. (\ref{eq13}):
	\begin{gather}
	\label{eq13}
	\operatorname{SSIM}(\mathbf{x}, \mathbf{y})=\frac{\left(2 \mu_{x} \mu_{y}+C_{1}\right)\left(2 \sigma_{x y}+C_{2}\right)}{\left(\mu_{x}^{2}+\mu_{y}^{2}+C_{1}\right)\left(\sigma_{x}^{2}+\sigma_{y}^{2}+C_{2}\right)},
	\end{gather}
	where $x$ and $y$ are the frames of the observed sequence ($\boldsymbol{S}$) and generated sequence ($\boldsymbol{\widehat{S}}$), respectively; $\mu_{x}$ and $\mu_{y}$ are the local means; $\sigma_{x}$ and $\sigma_{y}$  are the standard deviations; $\sigma_{x y}$ is the cross-covariance for frame $x$ and $y$; and  $C_{1}$ and $C_{2}$ are smoothing factors. However, the SSIM in Eq. (\ref{eq13}) is used for evaluating the similarity between two frames. To evaluate the whole video sequence, the mean of SSIM is used, as shown in Eq. (\ref{eq14}).
	\begin{gather}
	\label{eq14}
	\operatorname{SSIM}(\boldsymbol{S}, \boldsymbol{\widehat{S}})=\frac{1}{L-1} \sum_{t=2}^{L} \operatorname{SSIM}\left(S_{l}, \widehat{S}_{l}\right),
	\end{gather}
	SSIM ranges from -1 to 1, with a larger score indicating greater similarity. A larger SSIM indicates a better synthesis quality due to the higher perceptual similarity between the synthesized and observed sequences. Therefore, the SSIM in Eq. (\ref{eq14}) is used in this paper.

	\begin{figure}[t]
		\begin{center}
			\includegraphics[width=4cm,height=3.2cm]{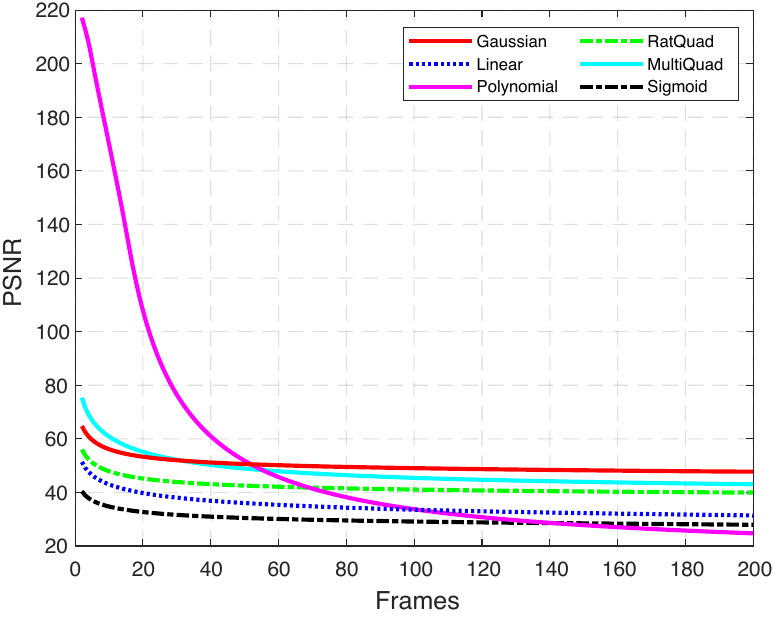}
			\hspace{4mm}
			\includegraphics[width=4cm,height=3.2cm]{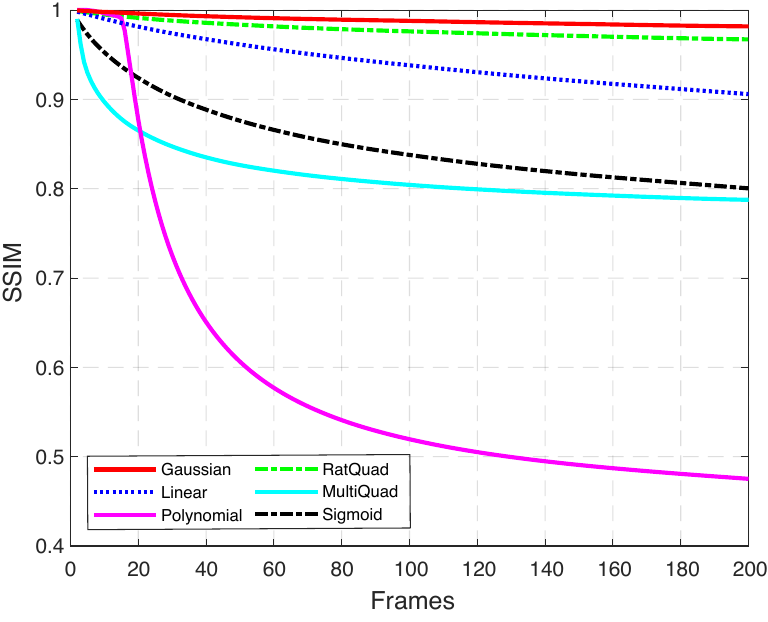}
			\\ \hspace{3mm} (a) PSNR  \hspace{33mm} (b) SSIM\\ 
			\caption{Quantitative comparison of different kernel functions (Gaussian: Gaussian kernel; Linear: Linear kernel; Polynomial: Polynomial kernel; RatQuad: Rational Quadratic kernel; MultiQuad: Multiquadric kernel; Sigmoid: Sigmoid kernel).}
			\label{fig:different-kernel}
		\end{center}
	\end{figure}

	\subsection{Experiment 1: Hyper-Parameters Selection}\label{sec4.2}
	In principle, there are two hyper-parameters influencing the performance of our proposed method: the kernel function $K(u,v)$ (in Eq. (\ref{eq11})) and the regularization factor $\lambda$ of the kernel similarity embedding (in Eq. (\ref{eq11})). As the kernel function selection is important for kernel learning \cite{you2016kernel,Liu2018LateFI,Liu2016MultipleKK,Liu2021HierarchicalMK,AlioschaPrez2020SVRGMKLAF} and it can directly affect the stability of our method, we comprehensively test the effects on the overall performance in order to obtain the optimal solution. Moreover, since the regularization factor $\lambda$ and kernel size $\gamma$ interfere with the learning stability for the kernel similarity embedding and impair the generalization performance of our model, we also investigate the optimal selections of $\lambda$ and $\gamma$.

	\subsubsection{Kernel Function $K(u,v)$}
	Most DT sequences lie in nonlinear manifolds containing different data modalities in their appearance distributions, structure dimensions, and stochastic repetitiveness, which are difficult to describe using low-dimensional latent variables with linear observation functions. Furthermore, the kernel function effectively represents the similarity correlation between different frames with Euclidean distance. Therefore, in this work, a kernel function is critical for kernel similarity embedding to make use of similarity prior knowledge. Here, we take several generic kernel functions (e.g., a linear kernel, polynomial kernel, Gaussian kernel, rational quadratic kernel, multiquadric kernel, and sigmoid kernel) for testing and select the optimal one for our model.

	For evaluation, we test different kernel functions for our method on the Dyntex dataset. As shown in Figure \ref{fig:different-kernel}, the different kernel functions exhibit various performances. Gaussian and rational quadratic kernel outperforms other kernel functions, with better PSNR and SSIM scores. This shows that our method can synthesize higher-quality DT sequences using these two kernel functions. Furthermore, the Gaussian kernel still achieves better performance after 200 frames of synthesized DT videos, which do not exist in the training set. This suggests that the Gaussian kernel function may provide better generalization performance. Meanwhile, we display different frames of two synthesized DT sequences (rotating wind ornament and windmill) with different kernel functions in Figure \ref{fig:different-kernel-images}. Intuitively, the frames of DTs synthesized using the Gaussian kernel, multiquadric kernel, and rational quadratic kernel are realistic, while other kernels fail (especially after 200 frames). In summary, the Gaussian kernel achieves better DT quality and sustainability. Therefore, we integrate the Gaussian kernel function ($K(u,v)=exp(-\gamma \lVert u-v \rVert^{2})$) with ELM into our powerfully unified DT synthesis system to learn the kernel similarity embedding for representing the spatial-temporal transition of DT videos in the later experiments.
	
	\begin{figure}[t]
		\begin{center}
			\includegraphics[width=4cm,height=3.2cm]{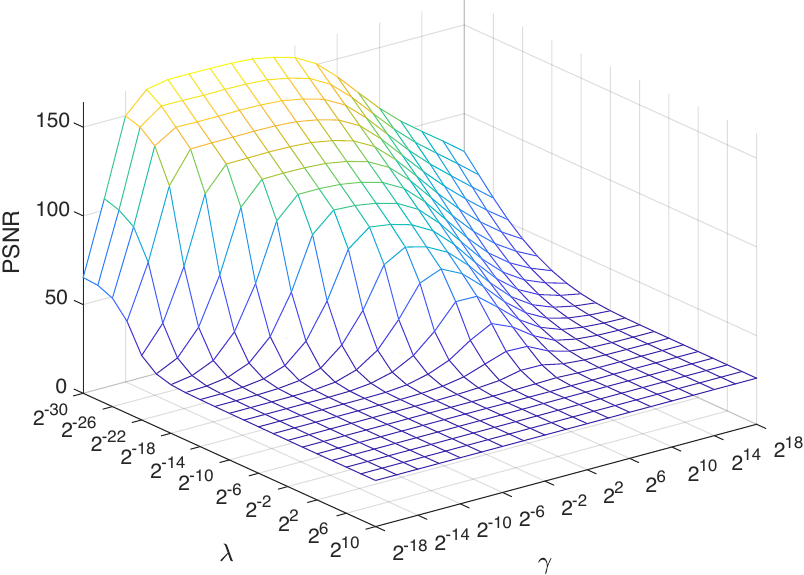}
			\hspace{4mm}
			\includegraphics[width=4cm,height=3.2cm]{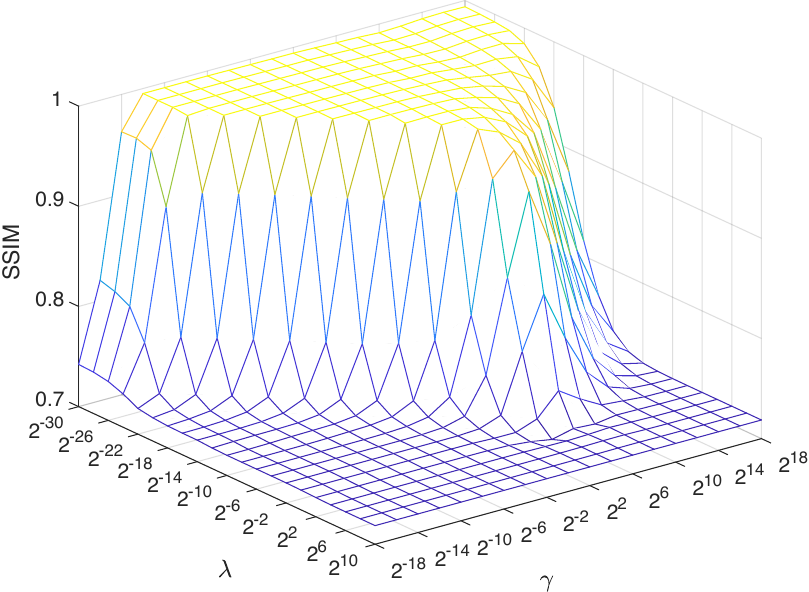}
			\\ \hspace{3mm} (a) PSNR  \hspace{33mm} (b) SSIM\\ 
			
			\caption{Quantitative comparison of various regularization factors $\lambda$ and kernel size $\gamma$ used in our method on the whole Dyntex.}
			\label{fig:different-parameters}
		\end{center}
	\end{figure}

	\begin{figure}[t]
		\begin{center}
			\includegraphics[width=.16\linewidth]{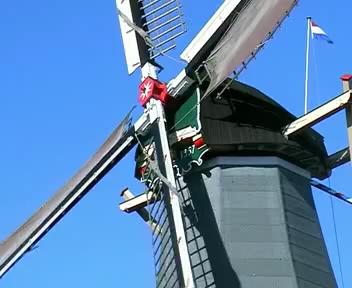}
			\includegraphics[width=.23\linewidth]{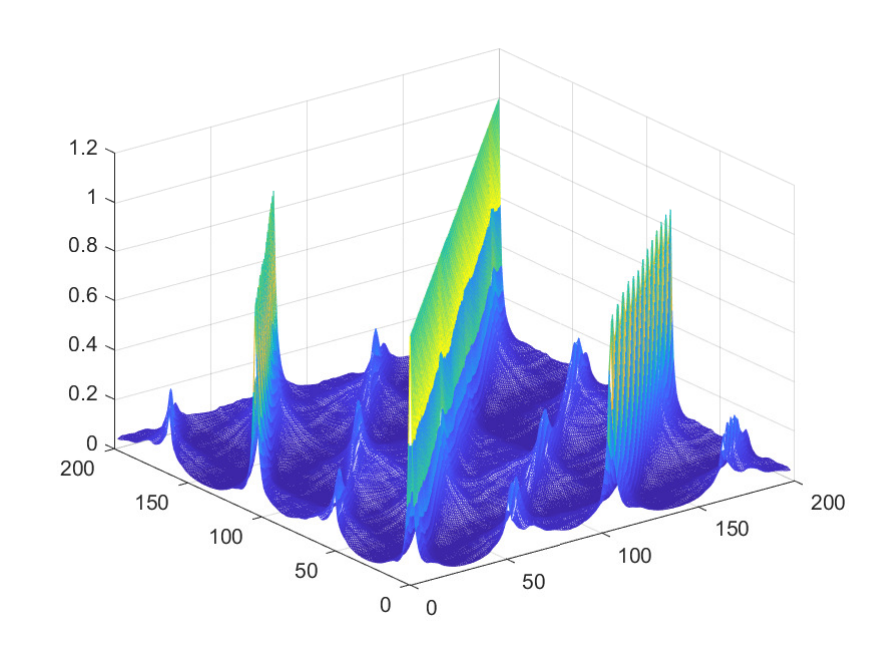}
			\includegraphics[width=.23\linewidth]{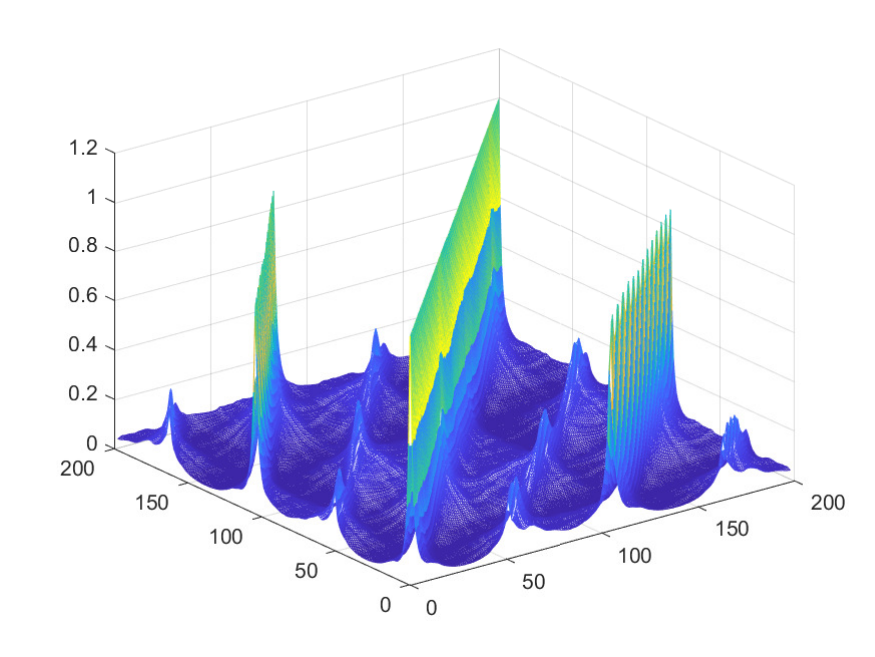}
			\includegraphics[width=.23\linewidth]{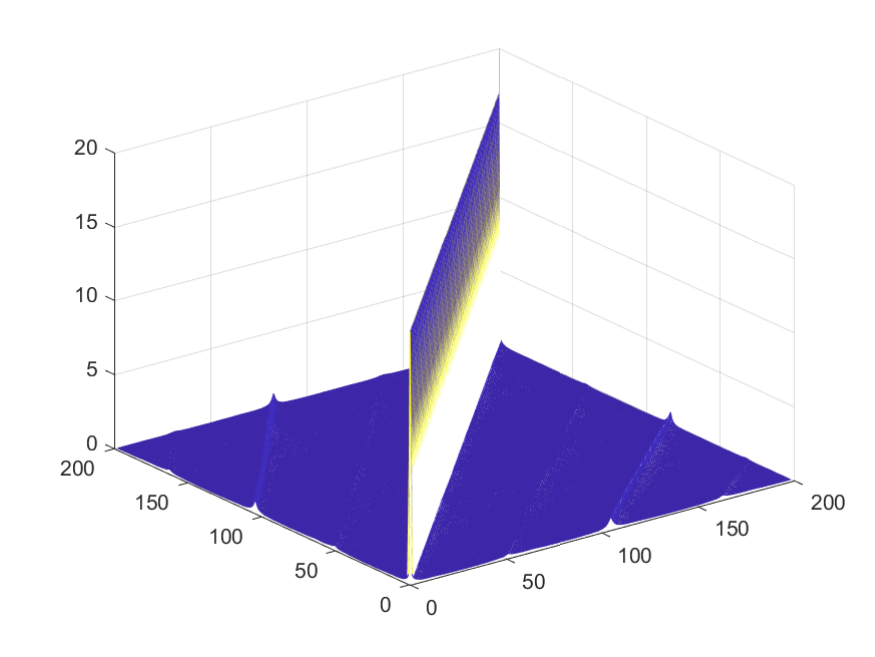}
			\\ {\footnotesize frame of sample \hspace{3mm} original \hspace{3mm} $\lambda=2^{-20}$, $\gamma=10^{8}$ \hspace{3mm} $\lambda=2^{4}$, $\gamma=2^{8}$}
			\\ (a) Windmill\\
			\includegraphics[width=.16\linewidth]{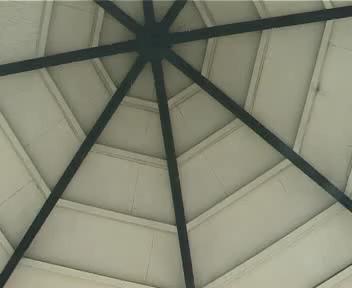}
			\includegraphics[width=.23\linewidth]{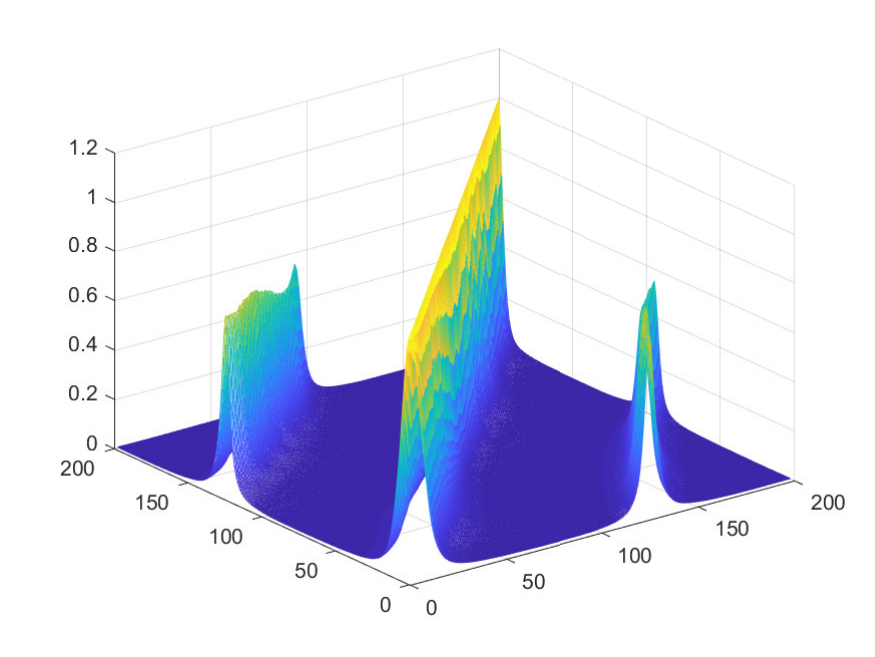}
			\includegraphics[width=.23\linewidth]{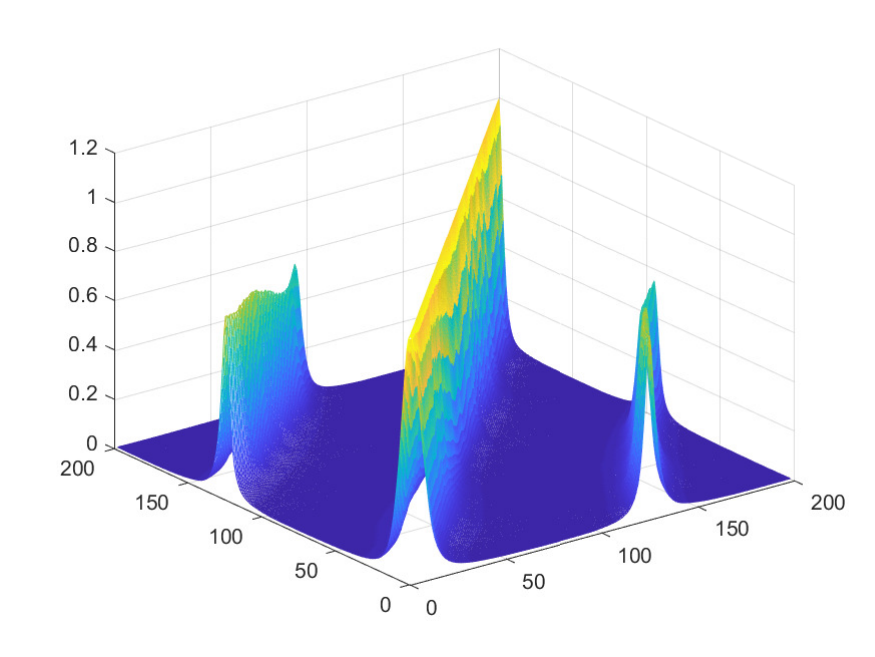}
			\includegraphics[width=.23\linewidth]{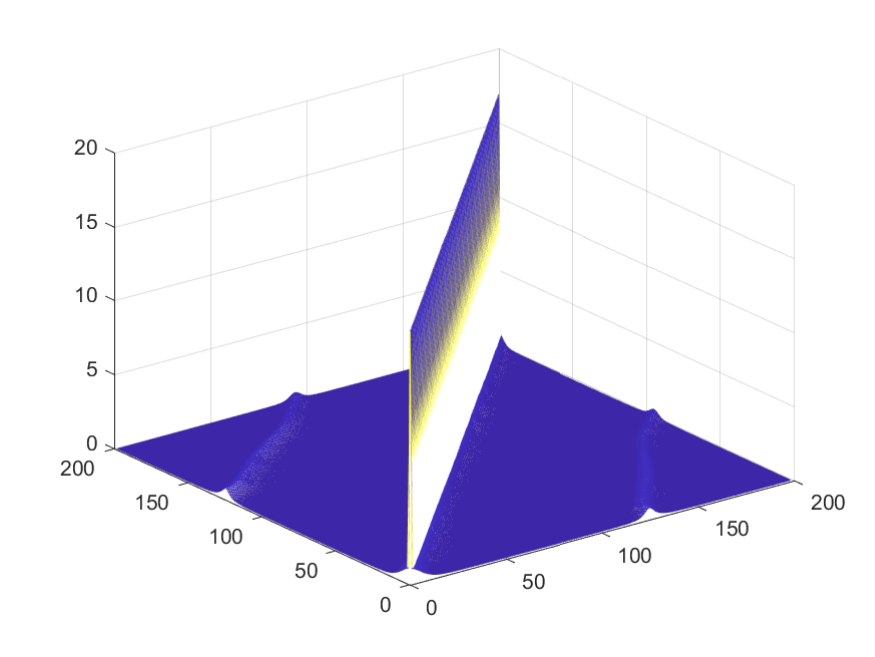}
			\\  {\footnotesize frame of sample \hspace{3mm} original \hspace{3mm} $\lambda=2^{-20}$, $\gamma=2^{8}$ \hspace{3mm} $\lambda=2^{4}$, $\gamma=2^{8}$}
			\\(b) Rotating\\	
			\caption{Demonstrating the regularization ability of regularization factor $\lambda$. We display the corresponding kernel similarity matrices on two samples with different $\lambda$ values: ($\lambda=2^{-20}$ shows under-regularization; $\lambda=2^{4}$ shows over-regularization).}
			\label{fig:different-parameters-Omega}
		\end{center}
	\end{figure}

	\subsubsection{Regularization Factor $\lambda$, Kernel Size $\gamma$}
	Following ridge regression theory [60], we add a positive value $\lambda\mathbf{I}$ ($\mathbf{I}$ is the identity matrix) to the diagonal axis of the kernel similarity matrix (Eq. \ref{eq11}) to achieve a more stable and better generalization performance for the DT synthesis model. Furthermore, the performance of SVM is known to be sensitive to the combination of the regularization factor and kernel size ($\lambda, \gamma$). Therefore, we also simultaneously analyze the influence of these two parameters ($\lambda, \gamma$) on our method.

	\begin{figure}[t]
		\begin{center}
			
			\includegraphics[width=.076\linewidth]{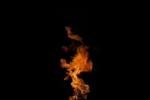}
			\includegraphics[width=.076\linewidth]{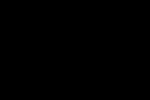}
			\includegraphics[width=.076\linewidth]{./sustainability/blank-frame.jpg}
			\includegraphics[width=.076\linewidth]{./sustainability/blank-frame.jpg}
			\includegraphics[width=.076\linewidth]{./sustainability/blank-frame.jpg}
			\includegraphics[width=.076\linewidth]{./sustainability/blank-frame.jpg}
			\includegraphics[width=.076\linewidth]{./sustainability/blank-frame.jpg}
			\includegraphics[width=.076\linewidth]{./sustainability/blank-frame.jpg}
			\includegraphics[width=.076\linewidth]{./sustainability/blank-frame.jpg}
			\includegraphics[width=.076\linewidth]{./sustainability/blank-frame.jpg}
			\includegraphics[width=.076\linewidth]{./sustainability/blank-frame.jpg}
			\\
			\vspace{1mm}

			\includegraphics[width=.076\linewidth]{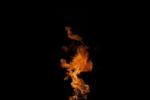}
			\includegraphics[width=.076\linewidth]{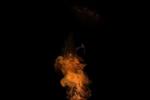}
			\includegraphics[width=.076\linewidth]{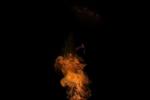}
			\includegraphics[width=.076\linewidth]{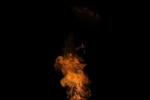}
			\includegraphics[width=.076\linewidth]{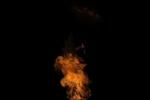}
			\includegraphics[width=.076\linewidth]{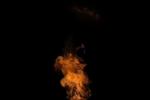}
			\includegraphics[width=.076\linewidth]{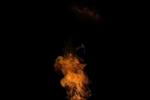}
			\includegraphics[width=.076\linewidth]{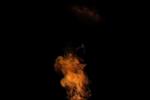}
			\includegraphics[width=.076\linewidth]{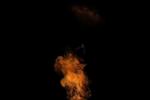}
			\includegraphics[width=.076\linewidth]{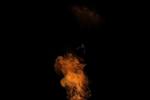}
			\includegraphics[width=.076\linewidth]{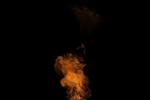}
			\\ (a) Flame

			\includegraphics[width=.076\linewidth]{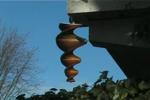}
			\includegraphics[width=.076\linewidth]{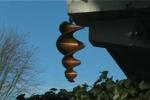}
			\includegraphics[width=.076\linewidth]{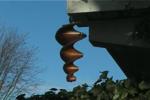}
			\includegraphics[width=.076\linewidth]{./sustainability/blank-frame.jpg}
			\includegraphics[width=.076\linewidth]{./sustainability/blank-frame.jpg}
			\includegraphics[width=.076\linewidth]{./sustainability/blank-frame.jpg}
			\includegraphics[width=.076\linewidth]{./sustainability/blank-frame.jpg}
			\includegraphics[width=.076\linewidth]{./sustainability/blank-frame.jpg}
			\includegraphics[width=.076\linewidth]{./sustainability/blank-frame.jpg}
			\includegraphics[width=.076\linewidth]{./sustainability/blank-frame.jpg}
			\includegraphics[width=.076\linewidth]{./sustainability/blank-frame.jpg}
			\\
			\vspace{1mm}
			
			\includegraphics[width=.076\linewidth]{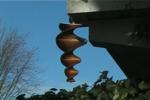}
			\includegraphics[width=.076\linewidth]{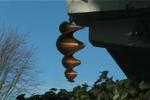}
			\includegraphics[width=.076\linewidth]{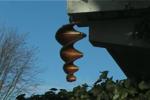}
			\includegraphics[width=.076\linewidth]{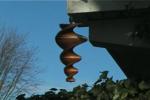}
			\includegraphics[width=.076\linewidth]{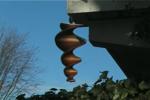}
			\includegraphics[width=.076\linewidth]{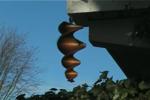}
			\includegraphics[width=.076\linewidth]{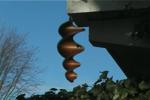}
			\includegraphics[width=.076\linewidth]{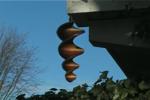}
			\includegraphics[width=.076\linewidth]{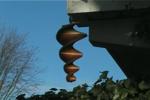}
			\includegraphics[width=.076\linewidth]{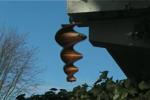}
			\includegraphics[width=.076\linewidth]{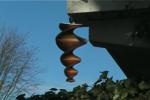}
			\\ (b) Rotating wind ornament

			\includegraphics[width=.076\linewidth]{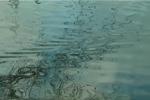}
			\includegraphics[width=.076\linewidth]{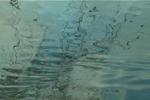}
			\includegraphics[width=.076\linewidth]{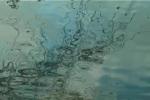}
			\includegraphics[width=.076\linewidth]{./sustainability/blank-frame.jpg}
			\includegraphics[width=.076\linewidth]{./sustainability/blank-frame.jpg}
			\includegraphics[width=.076\linewidth]{./sustainability/blank-frame.jpg}
			\includegraphics[width=.076\linewidth]{./sustainability/blank-frame.jpg}
			\includegraphics[width=.076\linewidth]{./sustainability/blank-frame.jpg}
			\includegraphics[width=.076\linewidth]{./sustainability/blank-frame.jpg}
			\includegraphics[width=.076\linewidth]{./sustainability/blank-frame.jpg}
			\includegraphics[width=.076\linewidth]{./sustainability/blank-frame.jpg}
			\\
			\vspace{1mm}
			
			\includegraphics[width=.076\linewidth]{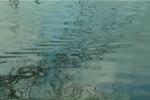}
			\includegraphics[width=.076\linewidth]{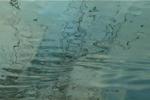}
			\includegraphics[width=.076\linewidth]{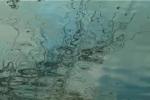}
			\includegraphics[width=.076\linewidth]{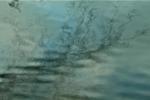}
			\includegraphics[width=.076\linewidth]{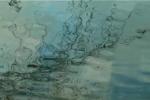}
			\includegraphics[width=.076\linewidth]{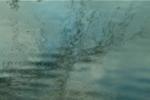}
			\includegraphics[width=.076\linewidth]{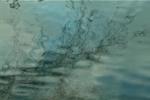}
			\includegraphics[width=.076\linewidth]{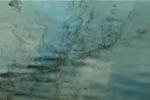}
			\includegraphics[width=.076\linewidth]{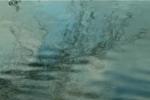}
			\includegraphics[width=.076\linewidth]{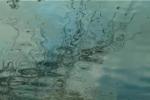}
			\includegraphics[width=.076\linewidth]{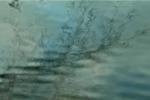}
			\\ (c) Water wave

			\includegraphics[width=.076\linewidth]{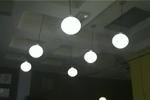}
			\includegraphics[width=.076\linewidth]{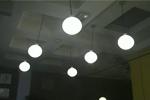}
			\includegraphics[width=.076\linewidth]{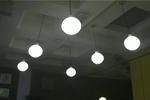}
			\includegraphics[width=.076\linewidth]{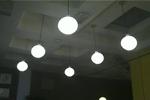}
			\includegraphics[width=.076\linewidth]{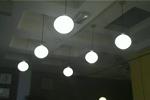}
			\includegraphics[width=.076\linewidth]{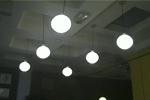}
			\includegraphics[width=.076\linewidth]{./sustainability/blank-frame.jpg}
			\includegraphics[width=.076\linewidth]{./sustainability/blank-frame.jpg}
			\includegraphics[width=.076\linewidth]{./sustainability/blank-frame.jpg}
			\includegraphics[width=.076\linewidth]{./sustainability/blank-frame.jpg}
			\includegraphics[width=.076\linewidth]{./sustainability/blank-frame.jpg}
			\\
			\vspace{1mm}
			
			\includegraphics[width=.076\linewidth]{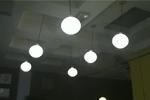}
			\includegraphics[width=.076\linewidth]{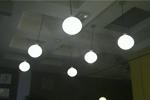}
			\includegraphics[width=.076\linewidth]{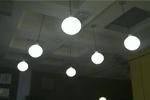}
			\includegraphics[width=.076\linewidth]{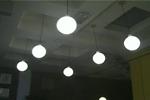}
			\includegraphics[width=.076\linewidth]{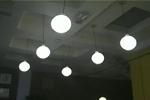}
			\includegraphics[width=.076\linewidth]{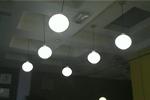}
			\includegraphics[width=.076\linewidth]{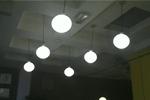}
			\includegraphics[width=.076\linewidth]{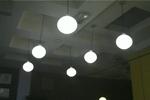}
			\includegraphics[width=.076\linewidth]{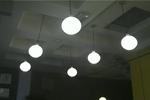}
			\includegraphics[width=.076\linewidth]{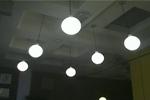}
			\includegraphics[width=.076\linewidth]{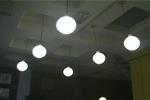}
			\\ (d) Bulb

			\includegraphics[width=.076\linewidth]{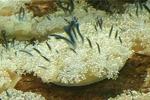}
			\includegraphics[width=.076\linewidth]{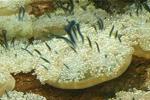}
			\includegraphics[width=.076\linewidth]{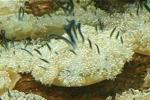}
			\includegraphics[width=.076\linewidth]{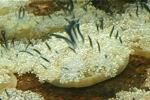}
			\includegraphics[width=.076\linewidth]{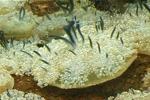}
			\includegraphics[width=.076\linewidth]{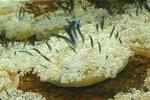}
			\includegraphics[width=.076\linewidth]{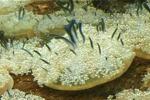}
			\includegraphics[width=.076\linewidth]{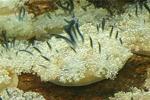}
			\includegraphics[width=.076\linewidth]{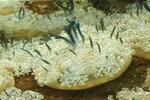}
			\includegraphics[width=.076\linewidth]{./sustainability/blank-frame.jpg}
			\includegraphics[width=.076\linewidth]{./sustainability/blank-frame.jpg}
			\\
			\vspace{1mm}
			
			\includegraphics[width=.076\linewidth]{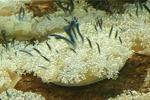}
			\includegraphics[width=.076\linewidth]{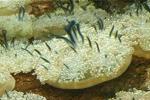}
			\includegraphics[width=.076\linewidth]{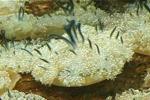}
			\includegraphics[width=.076\linewidth]{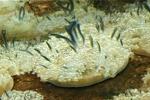}
			\includegraphics[width=.076\linewidth]{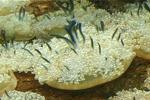}
			\includegraphics[width=.076\linewidth]{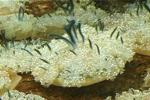}
			\includegraphics[width=.076\linewidth]{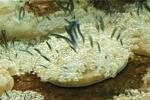}
			\includegraphics[width=.076\linewidth]{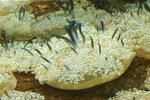}
			\includegraphics[width=.076\linewidth]{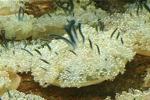}
			\includegraphics[width=.076\linewidth]{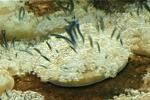}
			\includegraphics[width=.076\linewidth]{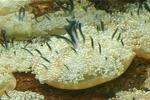}
			\\ (e) Flowers swaying with current

			\includegraphics[width=.076\linewidth]{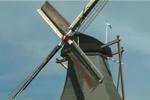}
			\includegraphics[width=.076\linewidth]{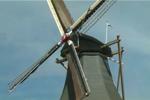}
			\includegraphics[width=.076\linewidth]{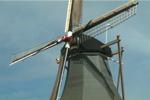}
			\includegraphics[width=.076\linewidth]{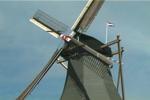}
			\includegraphics[width=.076\linewidth]{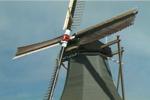}
			\includegraphics[width=.076\linewidth]{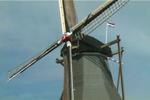}
			\includegraphics[width=.076\linewidth]{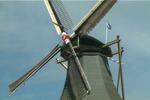}
			\includegraphics[width=.076\linewidth]{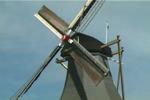}
			\includegraphics[width=.076\linewidth]{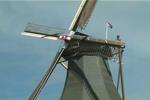}
			\includegraphics[width=.076\linewidth]{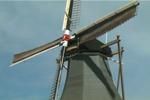}
			\includegraphics[width=.076\linewidth]{./sustainability/blank-frame.jpg}
			\\
			\vspace{1mm}
			
			\includegraphics[width=.076\linewidth]{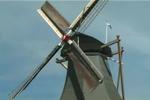}
			\includegraphics[width=.076\linewidth]{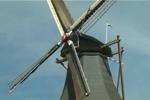}
			\includegraphics[width=.076\linewidth]{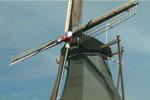}
			\includegraphics[width=.076\linewidth]{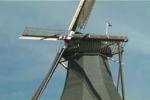}
			\includegraphics[width=.076\linewidth]{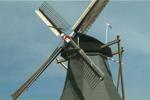}
			\includegraphics[width=.076\linewidth]{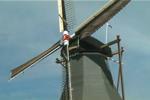}
			\includegraphics[width=.076\linewidth]{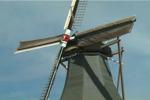}
			\includegraphics[width=.076\linewidth]{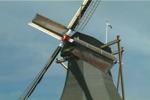}
			\includegraphics[width=.076\linewidth]{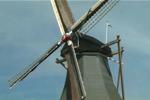}
			\includegraphics[width=.076\linewidth]{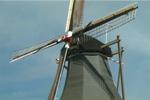}
			\includegraphics[width=.076\linewidth]{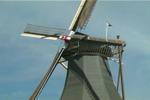}
			\\ (f) Windmill
			
			\caption{ Synthesizing long-term DT sequences using our method. For each category, the first row displays the 11 frames of the observed sequence (black frames indicate that the corresponding frame of the observed sequence is lacking), and the second row displays the corresponding frames of synthesized videos. From left to right, the columns are the 2-nd, 100-th, 200-th, 300-th, 400-th, 500-th, 600-th, 700-th, 800-th, 900-th, 1000-th frames of observed and synthesized sequences.}
			\label{fig:Sustainability-images}
		\end{center}
	\end{figure}

	To achieve good generalization performance, the regularization factor $\lambda$ and the kernel size $\gamma$ of the model need to be chosen appropriately. We investigate a wide range of $\lambda$ and $\gamma$. Specifically, we use 21 different values of $\lambda=\{2^{-30}, 2^{-28},\cdots, 2^{8}, 2^{10}\}$) and 19 different values of $\gamma=\{2^{-18}, 2^{-16},\cdots, 2^{16}, 2^{18}\}$) for evaluation on the Dyntex dataset, resulting in a total of 399 pairs of ($\lambda, \gamma$), as shown in Figure \ref{fig:different-parameters}. We find that the performance of our method becomes stable after two time periods (period 1:{$\lambda<10^{-14}$}, period 2:{$\lambda>10^{2}$}), which indicates that our method is over-fitting and under-fitting, respectively. Note that when $\lambda$ is small, the PSNR cannot be stable (see Figure \ref{fig:different-parameters}(a)) because $PSNR\propto +\infty$ when some generated frames are extremely similar to observed frames. The regularization of the model is insufficient if the regularization factor $\lambda$ is too small, which results in the DT synthesis model being overly confident for the training frames (the first 200 frames) and failing to generate high-quality DT frames after 200 frames. However, if too large $\lambda$ is used, the model is over-regularized, which leads to the stationarity and repetitiveness of the DT being overly smooth. That is, the weak correlations between different frames decrease excessively  (see Figure \ref{fig:different-parameters-Omega}). Figure \ref{fig:different-parameters} further shows that the kernel size $\gamma$ also closely interferes with the regularization ability of $\lambda$. We can observe that the PSNR and SSIM are not sensitive to $\gamma$ when $\lambda$ is small, but a relatively large $\gamma$ seems to yield better results. Therefore, the optimal combination of $(\lambda, \gamma)$ of our method with Gaussian kernel is chosen for later experiments ($\lambda=10^{-10}, \gamma=10^{8}$).

	\begin{figure}[t]
		\begin{center}
			\includegraphics[width=4cm,height=3.2cm]{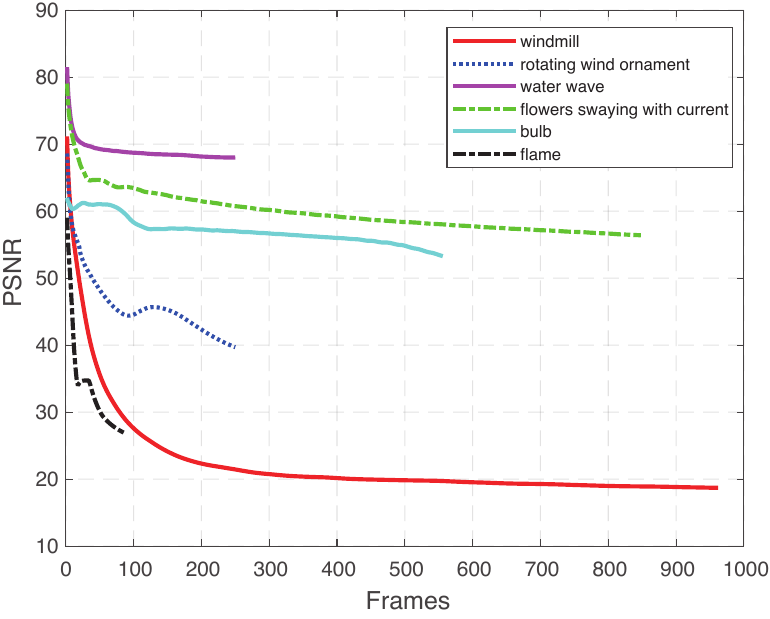}
			\hspace{4mm}
			\includegraphics[width=4cm,height=3.2cm]{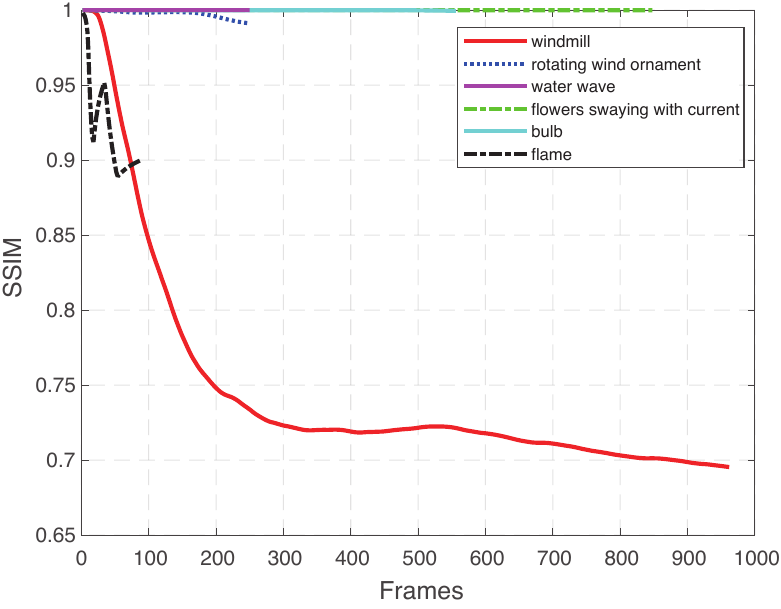}
			\\ \hspace{3mm} (a) PSNR  \hspace{33mm} (b) SSIM\\ 
			
			\caption{Demonstrating the sustainability of our method with quantitative evaluation on 6six DT videos.}
			\label{fig:Sustainability-evaluation}
		\end{center}
	\end{figure}

	\begin{figure*}[t]
		\begin{center}
			\includegraphics[width=.041\linewidth]{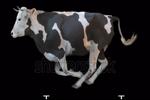}
			\includegraphics[width=.041\linewidth]{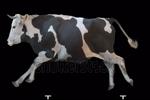}
			\includegraphics[width=.041\linewidth]{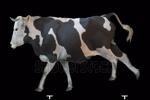}
			\includegraphics[width=.041\linewidth]{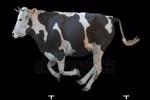}
			\hspace{1mm}
			\includegraphics[width=.041\linewidth]{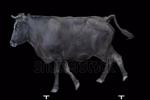}
			\includegraphics[width=.041\linewidth]{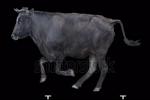}
			\includegraphics[width=.041\linewidth]{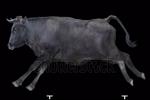}
			\includegraphics[width=.041\linewidth]{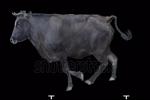}
			\hspace{1mm}
			\includegraphics[width=.041\linewidth]{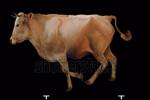}
			\includegraphics[width=.041\linewidth]{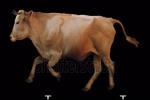}
			\includegraphics[width=.041\linewidth]{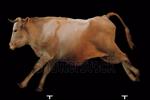}
			\includegraphics[width=.041\linewidth]{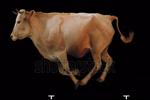}
			\hspace{1mm}
			\includegraphics[width=.041\linewidth]{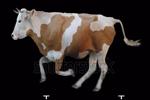}
			\includegraphics[width=.041\linewidth]{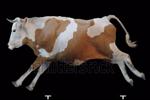}
			\includegraphics[width=.041\linewidth]{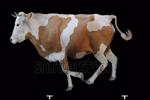}
			\includegraphics[width=.041\linewidth]{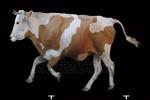}
			\hspace{1mm}
			\includegraphics[width=.041\linewidth]{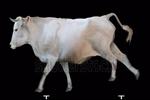}
			\includegraphics[width=.041\linewidth]{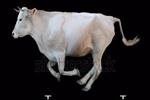}
			\includegraphics[width=.041\linewidth]{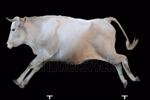}
			\includegraphics[width=.041\linewidth]{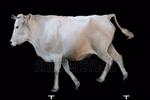}
			\vspace{1mm}\\

			\includegraphics[width=.041\linewidth]{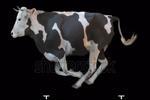}
			\includegraphics[width=.041\linewidth]{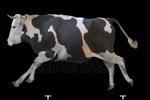}
			\includegraphics[width=.041\linewidth]{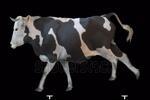}
			\includegraphics[width=.041\linewidth]{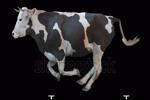}
			\hspace{1mm}
			\includegraphics[width=.041\linewidth]{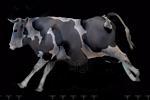}
			\includegraphics[width=.042\linewidth]{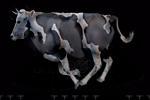}
			\includegraphics[width=.041\linewidth]{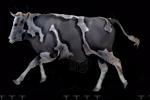}
			\includegraphics[width=.041\linewidth]{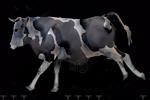}
			\hspace{1mm}
			\includegraphics[width=.041\linewidth]{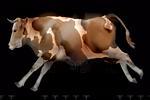}
			\includegraphics[width=.041\linewidth]{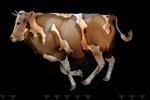}
			\includegraphics[width=.041\linewidth]{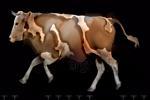}
			\includegraphics[width=.041\linewidth]{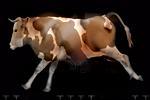}
			\hspace{1mm}
			\includegraphics[width=.041\linewidth]{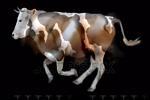}
			\includegraphics[width=.041\linewidth]{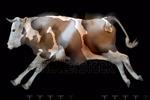}
			\includegraphics[width=.041\linewidth]{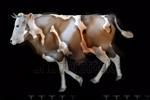}
			\includegraphics[width=.041\linewidth]{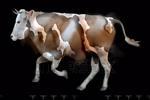}
			\hspace{1mm}
			\includegraphics[width=.041\linewidth]{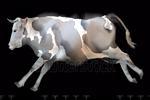}
			\includegraphics[width=.041\linewidth]{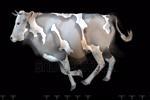}
			\includegraphics[width=.041\linewidth]{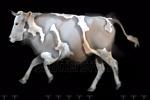}
			\includegraphics[width=.041\linewidth]{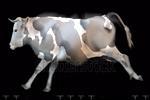}
			\vspace{1mm}\\
			
			\includegraphics[width=.041\linewidth]{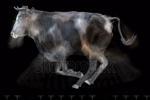}
			\includegraphics[width=.041\linewidth]{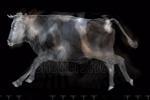}
			\includegraphics[width=.041\linewidth]{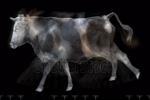}
			\includegraphics[width=.041\linewidth]{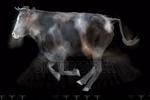}
			\hspace{1mm}
			\includegraphics[width=.041\linewidth]{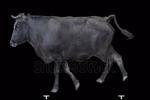}
			\includegraphics[width=.041\linewidth]{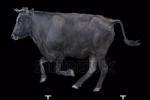}
			\includegraphics[width=.041\linewidth]{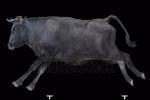}
			\includegraphics[width=.041\linewidth]{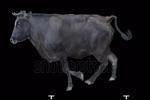}
			\hspace{1mm}
			\includegraphics[width=.041\linewidth]{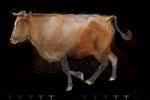}
			\includegraphics[width=.041\linewidth]{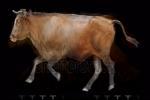}
			\includegraphics[width=.041\linewidth]{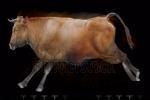}
			\includegraphics[width=.041\linewidth]{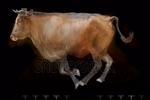}
			\hspace{1mm}
			\includegraphics[width=.041\linewidth]{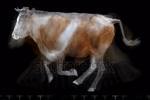}
			\includegraphics[width=.041\linewidth]{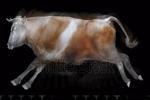}
			\includegraphics[width=.041\linewidth]{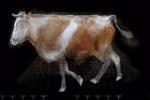}
			\includegraphics[width=.041\linewidth]{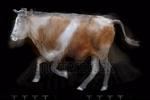}
			\hspace{1mm}
			\includegraphics[width=.041\linewidth]{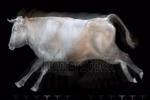}
			\includegraphics[width=.041\linewidth]{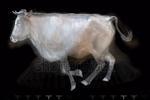}
			\includegraphics[width=.041\linewidth]{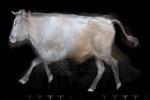}
			\includegraphics[width=.041\linewidth]{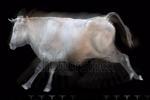}
			\vspace{1mm}\\
			
			\includegraphics[width=.041\linewidth]{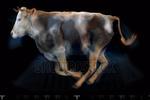}
			\includegraphics[width=.041\linewidth]{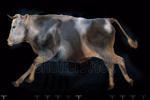}
			\includegraphics[width=.041\linewidth]{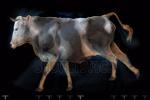}
			\includegraphics[width=.041\linewidth]{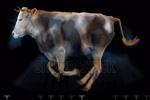}
			\hspace{1mm}
			\includegraphics[width=.041\linewidth]{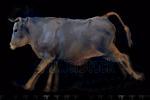}
			\includegraphics[width=.041\linewidth]{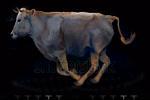}
			\includegraphics[width=.041\linewidth]{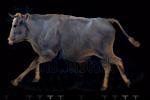}
			\includegraphics[width=.041\linewidth]{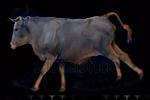}
			\hspace{1mm}
			\includegraphics[width=.041\linewidth]{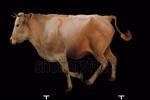}
			\includegraphics[width=.041\linewidth]{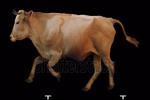}
			\includegraphics[width=.041\linewidth]{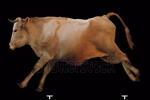}
			\includegraphics[width=.041\linewidth]{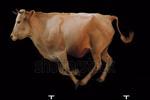}
			\hspace{1mm}
			\includegraphics[width=.041\linewidth]{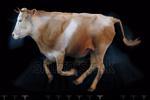}
			\includegraphics[width=.041\linewidth]{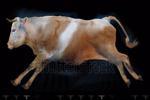}
			\includegraphics[width=.041\linewidth]{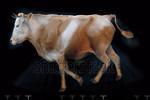}
			\includegraphics[width=.041\linewidth]{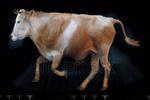}
			\hspace{1mm}
			\includegraphics[width=.041\linewidth]{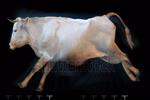}
			\includegraphics[width=.041\linewidth]{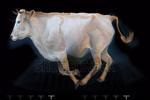}
			\includegraphics[width=.041\linewidth]{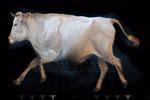}
			\includegraphics[width=.041\linewidth]{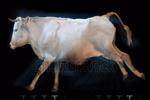}
			\vspace{1mm}\\
			
			\includegraphics[width=.041\linewidth]{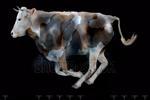}
			\includegraphics[width=.041\linewidth]{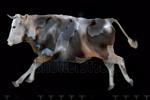}
			\includegraphics[width=.041\linewidth]{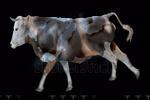}
			\includegraphics[width=.041\linewidth]{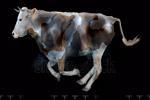}
			\hspace{1mm}
			\includegraphics[width=.041\linewidth]{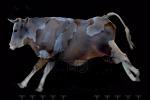}
			\includegraphics[width=.041\linewidth]{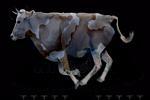}
			\includegraphics[width=.041\linewidth]{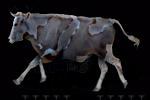}
			\includegraphics[width=.041\linewidth]{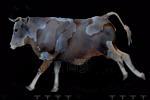}
			\hspace{1mm}
			\includegraphics[width=.041\linewidth]{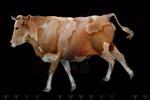}
			\includegraphics[width=.041\linewidth]{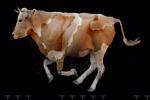}
			\includegraphics[width=.041\linewidth]{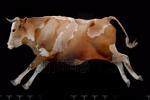}
			\includegraphics[width=.041\linewidth]{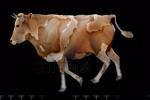}
			\hspace{1mm}
			\includegraphics[width=.041\linewidth]{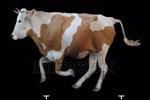}
			\includegraphics[width=.041\linewidth]{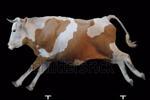}
			\includegraphics[width=.041\linewidth]{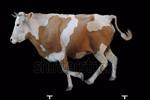}
			\includegraphics[width=.041\linewidth]{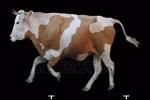}
			\hspace{1mm}
			\includegraphics[width=.041\linewidth]{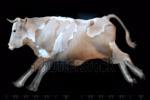}
			\includegraphics[width=.041\linewidth]{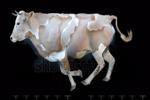}
			\includegraphics[width=.041\linewidth]{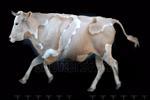}
			\includegraphics[width=.041\linewidth]{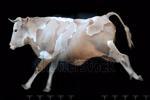}
			\vspace{1mm}\\
			
			\includegraphics[width=.041\linewidth]{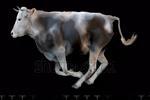}
			\includegraphics[width=.041\linewidth]{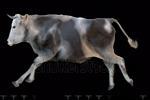}
			\includegraphics[width=.041\linewidth]{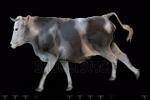}
			\includegraphics[width=.041\linewidth]{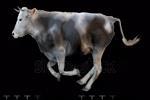}
			\hspace{1mm}
			\includegraphics[width=.041\linewidth]{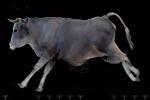}
			\includegraphics[width=.041\linewidth]{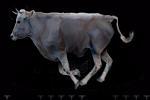}
			\includegraphics[width=.041\linewidth]{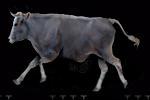}
			\includegraphics[width=.041\linewidth]{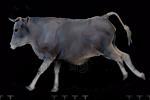}
			\hspace{1mm}
			\includegraphics[width=.041\linewidth]{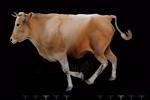}
			\includegraphics[width=.041\linewidth]{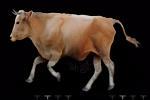}
			\includegraphics[width=.041\linewidth]{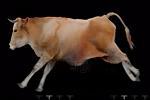}
			\includegraphics[width=.041\linewidth]{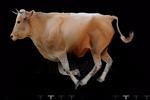}
			\hspace{1mm}
			\includegraphics[width=.041\linewidth]{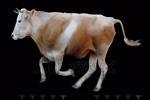}
			\includegraphics[width=.041\linewidth]{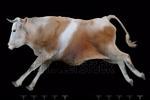}
			\includegraphics[width=.041\linewidth]{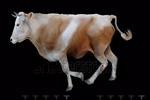}
			\includegraphics[width=.041\linewidth]{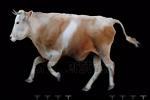}
			\hspace{1mm}
			\includegraphics[width=.041\linewidth]{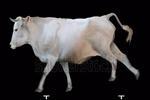}
			\includegraphics[width=.041\linewidth]{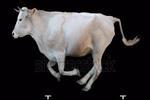}
			\includegraphics[width=.041\linewidth]{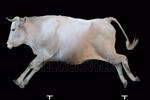}
			\includegraphics[width=.041\linewidth]{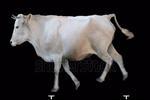} 
			\vspace{1mm}\\ (a) Running cows \vspace{2mm}\\

			\includegraphics[width=.041\linewidth]{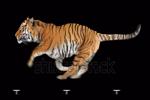}
			\includegraphics[width=.041\linewidth]{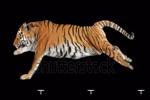}
			\includegraphics[width=.041\linewidth]{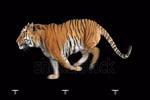}
			\includegraphics[width=.041\linewidth]{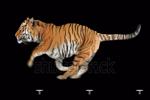}
			\hspace{1mm}
			\includegraphics[width=.041\linewidth]{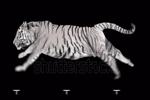}
			\includegraphics[width=.041\linewidth]{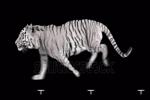}
			\includegraphics[width=.041\linewidth]{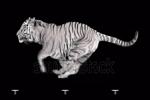}
			\includegraphics[width=.041\linewidth]{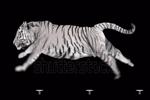} 
			\hspace{10mm}
			\includegraphics[width=.041\linewidth]{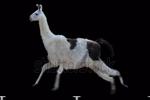}
			\includegraphics[width=.041\linewidth]{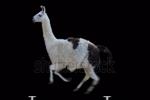}
			\includegraphics[width=.041\linewidth]{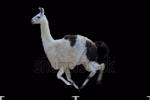}
			\includegraphics[width=.041\linewidth]{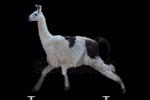}
			\hspace{1mm}
			\includegraphics[width=.041\linewidth]{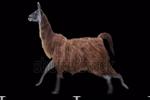}
			\includegraphics[width=.041\linewidth]{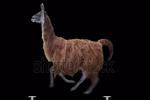}
			\includegraphics[width=.041\linewidth]{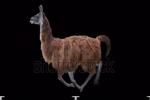}
			\includegraphics[width=.041\linewidth]{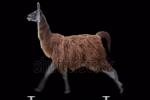} 
			\vspace{1mm}\\
			
			\includegraphics[width=.041\linewidth]{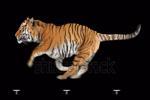}
			\includegraphics[width=.041\linewidth]{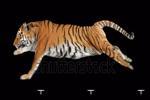}
			\includegraphics[width=.041\linewidth]{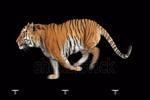}
			\includegraphics[width=.041\linewidth]{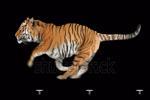}
			\hspace{1mm}
			\includegraphics[width=.041\linewidth]{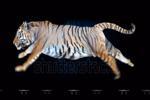}
			\includegraphics[width=.041\linewidth]{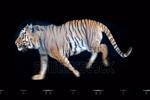}
			\includegraphics[width=.041\linewidth]{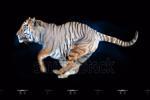}
			\includegraphics[width=.041\linewidth]{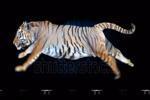}
			\hspace{10mm}
			\includegraphics[width=.041\linewidth]{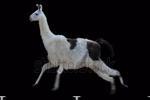}
			\includegraphics[width=.041\linewidth]{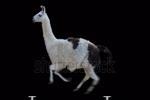}
			\includegraphics[width=.041\linewidth]{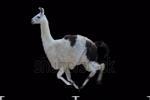}
			\includegraphics[width=.041\linewidth]{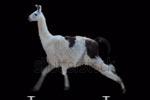}
			\hspace{1mm}
			\includegraphics[width=.041\linewidth]{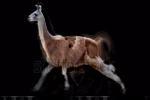}
			\includegraphics[width=.041\linewidth]{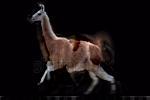}
			\includegraphics[width=.041\linewidth]{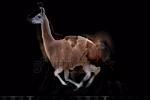}
			\includegraphics[width=.041\linewidth]{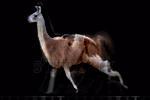}
			\vspace{1mm}\\
			
			\includegraphics[width=.041\linewidth]{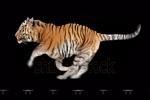}
			\includegraphics[width=.041\linewidth]{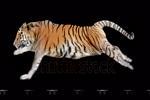}
			\includegraphics[width=.041\linewidth]{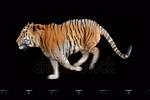}
			\includegraphics[width=.041\linewidth]{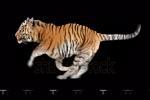}
			\hspace{1mm}
			\includegraphics[width=.041\linewidth]{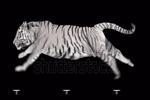}
			\includegraphics[width=.041\linewidth]{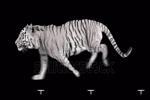}
			\includegraphics[width=.041\linewidth]{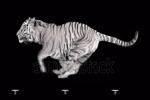}
			\includegraphics[width=.041\linewidth]{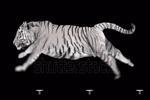}
			\hspace{10mm}
			\includegraphics[width=.041\linewidth]{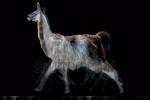}
			\includegraphics[width=.041\linewidth]{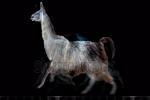}
			\includegraphics[width=.041\linewidth]{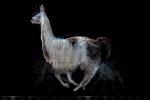}
			\includegraphics[width=.041\linewidth]{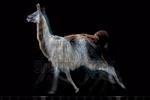}
			\hspace{1mm}
			\includegraphics[width=.041\linewidth]{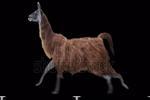}
			\includegraphics[width=.041\linewidth]{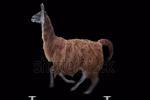}
			\includegraphics[width=.041\linewidth]{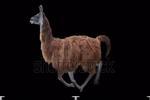}
			\includegraphics[width=.041\linewidth]{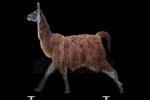}
			\vspace{1mm}\\ (b) Running tigers  \hspace{5cm} (c) Running llamas \\

			\caption{Sythesizing DTs by transferring the model trained using our method. For the running cows, the first row displays the 20 frames of 5 observed sequences, while the other rows display the frames of the synthesized sequences corresponding to the first row, with different trained models (from top to bottom: trained on cow1, trained on cow2, trained on cow3, trained on cow4, trained on cow5). The running tiger and running llamas are similar to the running cows, but we only use two observed sequences for training and testing.}
			\label{fig:generalization-images}
		\end{center}
	\end{figure*}

	\subsection{Experiment 2: Sustainability Analysis}\label{sec4.3}
	DT synthesis aims to generate high-quality, long-term DT sequences, which requires a synthesis method with good sustainability performance, i.e., no obvious visual decay, divergence, or abrupt jumps for generated long-term sequences. Therefore, we intuitively analyze the sustainability of our method using visual quality and quantitative evaluation metrics on two datasets. 
	
	For evaluation, the visual quality comparison over several synthesized DT sequences of different classes is presented in Figure \ref{fig:Sustainability-images}. To demonstrate the robustness of our method, we train the model using the first 200 frames if the length $\l$ of observed sequences is longer than 200, and otherwise, all frames of observed sequences are used. We observe that our method not only generates high-fidelity DTs in the short-term, but also generates high-quality DTs in the long-term even if the observed sequences are short, e.g. flame ($l=88$), rotating wind ornament ($l=250$), water wave ($l=250$), bulb ($l=556$), flowers swaying with current ($l=848$), windmill ($l=962$). Note that Figure \ref{fig:Sustainability-images}(a) seemingly shares with similar DT (from 100-th to 1000-th), because these generated frames are located in a similar/same cycle. Furthermore, our method still keeps synthesizing realistic DTs (including the details) in the long-term, e.g., the flag in the ``windmill'' sample exhibits its dynamics in the long-term generated sequences.

	We also show the quantitative evaluation results to demonstrate the sustainability of our method. Here we report the mean SSIM and PSNR in terms of frames (from 1 to 848) of the six observed sequences used in the former visual quality evaluation. As can be seen from Figure \ref{fig:Sustainability-evaluation}, our method achieves desirable mean PSNR and SSIM, which shows that it synthesizes high-fidelity DTs. Although the mean PSNR and SSIM decrease as the number of generated frames increases for some DT videos (e.g., windmill, frame), they are still large ($PSNR>18$ dB, $SSIM>0.69$). Notably, our method achieves a high SSIM index of 1 for whole long-term sequences just using 200 frames for training (e.g., flowers swaying with current, bulb, water wave), which suggests that it generates videos that are nearly the same as the observed sequences. These results prove that our method accurately exhibits the statistical stationarity in the spatial domain and the stochastic repetitiveness in the temporal dimension of DT sequences using kernel similarity embedding, and can thus synthesize realistic DTs in the long-term.

	\begin{table*}[t]
		\centering
		\captionsetup{justification=centering}
		\caption{\textsc{Comparison with Non-neural-network-based DT Synthesis Methods in Terms of PSNR} (dB) \textsc{and SSIM.}}\label{Table:SSIM-PSNR}
		\setlength{\tabcolsep}{0.8mm}{
			\begin{tabular}{l|c|c|c|c|c|c|c}
				\hline
				& \textbf{Ours} & FFT-LDS \cite{Abraham2005DynamicTW}& HOSVD \cite{Costantini2008HigherOS} & KPCR \cite{you2016kernel} & LDS \cite{doretto2003dynamic}  & SLDS \cite{Siddiqi2007ACG}& Kernel-DT \cite{Chan2007ClassifyingVW} \\
				& PSNR   \quad SSIM  & PSNR   \quad SSIM  & PSNR   \quad SSIM  & PSNR   \quad SSIM  & PSNR   \quad SSIM  & PSNR   \quad SSIM  & PSNR   \quad SSIM  \\
				\hline
				boiling water & \textbf{36.591} \quad \textbf{0.958} & 27.570 \quad 0.887 & 27.617 \quad 0.891 & 24.726 \quad 0.840 & 27.604 \quad 0.891 & 27.604 \quad 0.891 & 26.114 \quad 0.870 \\
				elevator & \textbf{46.149} \quad \textbf{0.996} & 34.307 \quad 0.949 & 34.289 \quad 0.946 & 31.029 \quad 0.913 & 34.420 \quad 0.952 & 34.384 \quad 0.951 & 30.109 \quad 0.893 \\
				rotating wind ornament & \textbf{15.882} \quad 0.564 & 13.387 \quad 0.500 & 12.644 \quad 0.473 & 15.038 \quad \textbf{0.569} & 13.387 \quad 0.500 & 13.387 \quad 0.500 & 12.131 \quad 0.459 \\
				flower in current & \textbf{47.609} \quad \textbf{0.999} & 30.372 \quad 0.922 & 31.778 \quad 0.949 & 37.069 \quad 0.988 & 31.392 \quad 0.946 & 30.922 \quad 0.937 & 27.297 \quad 0.891 \\
				bulb & \textbf{49.445} \quad \textbf{1.000} & 31.229 \quad 0.958 & 31.236 \quad 0.972 & 28.024 \quad 0.957 & 31.350 \quad 0.972 & 31.350 \quad 0.972 & 29.788 \quad 0.978 \\
				flashing lights & 17.874 \quad 0.748 & 26.714 \quad 0.875 & 26.712 \quad 0.876 & 22.242 \quad 0.796 & \textbf{26.724} \quad \textbf{0.878} & \textbf{26.724} \quad \textbf{0.878} & 25.668 \quad 0.788 \\
				spring water & \textbf{64.198} \quad \textbf{1.000} & 21.435 \quad 0.607 & 21.363 \quad 0.606 & 21.271 \quad 0.641 & 21.453 \quad 0.610 & 21.453 \quad 0.610 & 21.211 \quad 0.643 \\
				washing machine & \textbf{33.399} \quad \textbf{0.960} & 30.875 \quad 0.931 & 30.865 \quad 0.933 & 28.630 \quad 0.905 & 30.913 \quad 0.934 & 30.913 \quad 0.934 & 26.391 \quad 0.902 \\
				fountain & \textbf{68.641} \quad \textbf{1.000} & 19.567 \quad 0.401 & 19.554 \quad 0.401 & 18.394 \quad 0.357 & 19.569 \quad 0.402 & 19.569 \quad 0.402 & 18.745 \quad 0.382 \\
				water spray & \textbf{43.215} \quad \textbf{0.994} & 29.387 \quad 0.880 & 29.683 \quad 0.890 & 30.740 \quad 0.917 & 29.483 \quad 0.889 & 28.654 \quad 0.878 & 25.565 \quad 0.846 \\
				water spray in a pool & \textbf{67.475} \quad \textbf{1.000} & 21.076 \quad 0.426 & 21.036 \quad 0.423 & 20.603 \quad 0.433 & 21.079 \quad 0.427 & 21.079 \quad 0.427 & 19.483 \quad 0.394 \\
				water wave & \textbf{51.840} \quad \textbf{0.999} & 27.385 \quad 0.650 & 27.311 \quad 0.646 & 27.767 \quad 0.745 & 27.394 \quad 0.651 & 27.394 \quad 0.651 & 22.371 \quad 0.537 \\
				waterfall & \textbf{44.708} \quad \textbf{0.998} & 39.649 \quad 0.991 & 43.240 \quad 0.998 & 45.337 \quad 0.999 & 41.310 \quad 0.996 & 41.310 \quad 0.996 & 27.073 \quad 0.955 \\
				waterfall in a mountain & \textbf{71.851} \quad \textbf{1.000} & 18.509 \quad 0.539 & 18.507 \quad 0.540 & 18.408 \quad 0.534 & 18.513 \quad 0.540 & 18.513 \quad 0.540 & 18.303 \quad 0.535 \\
				flag & \textbf{53.605} \quad \textbf{1.000} & 23.839 \quad 0.858 & 23.797 \quad 0.859 & 20.537 \quad 0.802 & 23.840 \quad 0.859 & 23.840 \quad 0.859 & 23.068 \quad 0.854 \\
				\hline
				flame   & \textbf{46.185} \quad \textbf{0.910} & 27.567 \quad 0.874 & 27.463 \quad 0.867 & 26.974 \quad 0.904 & 27.558 \quad 0.877 & 26.435 \quad 0.857 & 33.495 \quad 0.887 \\
				beach   & 22.736 \quad 0.719 & 26.684 \quad 0.838 & 31.017 \quad 0.899 & \textbf{33.148} \quad \textbf{0.942} & 26.779 \quad 0.846 & 26.779 \quad 0.846 & 29.251 \quad 0.905 \\
				\hline
				mean    & \textbf{45.965} \quad \textbf{0.932} & 26.444 \quad 0.770 & 26.948 \quad 0.775 & 26.467 \quad 0.779 & 26.633 \quad 0.775 & 26.489 \quad 0.772 & 24.474 \quad 0.748 \\
				\hline
		\end{tabular}}
	\end{table*}

	\subsection{Experiment 3: Generalization Analysis}\label{sec4.4}
	Good generalization performance is a key goal for all learning tasks. Similar to \cite{Xie2021LearningES}, we also specialize in our method to learn roughly aligned DT videos, which are non-stationary in either the spatial or temporal domain. This study differs from \cite{Xie2021LearningES}, however, which trains a model using all roughly aligned video sequences for one example (e.g., five training sequences for the running cow). Our method instead trains a model using only one video sequence for each example, which confirms the generalization performance of our method. 
	
	Spatially aligned with the sense for each time step, the target objects in different videos possess the same locations, shapes, and poses, while it is the same as temporally aligned with the starting and ending times of the actions in different videos. We take the DT videos used in \cite{Xie2021LearningES} for evaluation. As can be seen from Figure \ref{fig:generalization-images}, the three results of modeling and synthesizing DTs from roughly aligned video sequences are displayed. Specifically, we first train a model on each sequence of the running cows/tigers/llamas, and then test the 5/2/2 trained models on the 5/2/2 observed sequences. Thus, we obtain 33 realistic, synthesized sequences. 
	
	The experimental results show that our method can transfer the trained model to generate new sequences for other spatial-temporally aligned DT sequences. In summary, our method is effective and efficient for synthesizing realistic appearances and motions for the test animals, which suggests that it has high generalization ability. However, the transferred model cannot synthesize consistent motions for some cows (e.g., cow1$\to$cow2), because these samples are not initially aligned well.

	\subsection{Experiment 4: Comparisons to State of the Art}\label{sec4.5}
	In this section, we compare our method with nine state-of-the-art methods for DT synthesis, including non-neural-network-based methods(LDS \cite{doretto2003dynamic}, FFT-LDS \cite{Abraham2005DynamicTW}, HOSVD \cite{Costantini2008HigherOS}, SLDS \cite{Siddiqi2007ACG}, Kernel-DT \cite{Chan2007ClassifyingVW} and KPCR \cite{you2016kernel}) and neural-network-based methods (TwoStream\footnote{https://ryersonvisionlab.github.io/two-stream-projpage/} \cite{tesfaldet2018two}, STGCN\footnote{http://www.stat.ucla.edu/ jxie/STGConvNet/STGConvNet.html} \cite{Xie2021LearningES} and DG\footnote{http://www.stat.ucla.edu/jxie/DynamicGenerator/DynamicGenerator.html} \cite{Xie2019LearningDG}). To better verify and validate the performance of our method, we simultaneously leverage quantitative evaluation metrics (SSIM, PSNR), time consumption and vision quality.
	
	Specifically, we first compare our method with six non-neural-network-based methods, including FFT-LDS \cite{Abraham2005DynamicTW}, HOSVD \cite{Costantini2008HigherOS}, KPCR \cite{you2016kernel}, LDS \cite{doretto2003dynamic}, SLDS \cite{Siddiqi2007ACG}, Kernel-DT \cite{Chan2007ClassifyingVW}. To facilitate direct comparison, we test all models with 150$\times$100 pixels using 17 gray DT videos, in terms of PSNR and SSIM. As can be seen from Table \ref{Table:SSIM-PSNR}, our method attains the best performance on most of DT sequences, except for the videos of flashing lights and beach. This is because these two videos lack the good spatial-temporal characteristics of DT. Notably, the proposed method beats the second-best results by a large margin (19.017 dB for average PSNR and 0.153 for average SSIM on 17 DT videos). Our method also achieves a significant SSIM index of 1 for some DT videos (e.g., bulb, fountain, spring water, etc.). Therefore, our method clearly makes full use of the similarity as prior knowledge for DT synthesis using kernel similarity embedding. However, all methods fail to synthesize high-quality DT videos for the rotating wind ornament, because this sample is originally blurry.

	\begin{figure}[ht]
		\begin{center}
			\includegraphics[width=.07\linewidth]{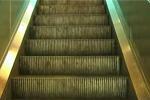}
			\includegraphics[width=.07\linewidth]{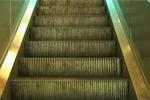}
			\includegraphics[width=.07\linewidth]{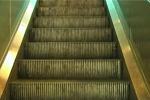}
			\includegraphics[width=.07\linewidth]{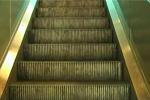}
			\includegraphics[width=.07\linewidth]{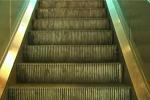}
			\includegraphics[width=.07\linewidth]{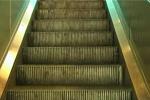}
			\hspace{1mm}
			\includegraphics[width=.07\linewidth]{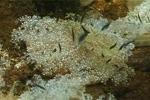}
			\includegraphics[width=.07\linewidth]{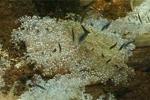}
			\includegraphics[width=.07\linewidth]{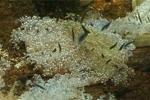}
			\includegraphics[width=.07\linewidth]{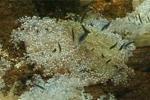}
			\includegraphics[width=.07\linewidth]{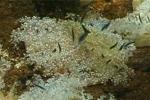}
			\includegraphics[width=.07\linewidth]{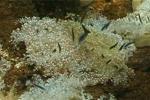} \\
			\vspace{1mm}
			
			\includegraphics[width=.07\linewidth]{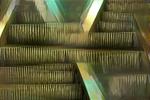}
			\includegraphics[width=.07\linewidth]{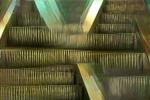}
			\includegraphics[width=.07\linewidth]{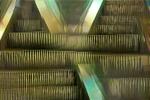}
			\includegraphics[width=.07\linewidth]{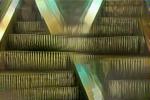}
			\includegraphics[width=.07\linewidth]{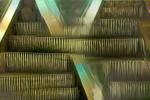}
			\includegraphics[width=.07\linewidth]{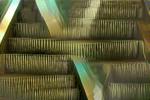}
			\hspace{1mm}
			\includegraphics[width=.07\linewidth]{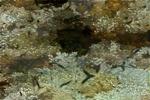}
			\includegraphics[width=.07\linewidth]{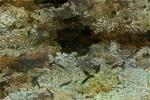}
			\includegraphics[width=.07\linewidth]{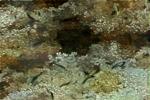}
			\includegraphics[width=.07\linewidth]{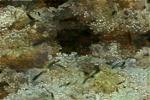}
			\includegraphics[width=.07\linewidth]{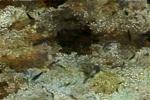}
			\includegraphics[width=.07\linewidth]{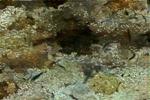} \\
			\vspace{1mm}
			
			\includegraphics[width=.07\linewidth]{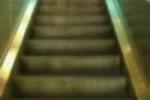}
			\includegraphics[width=.07\linewidth]{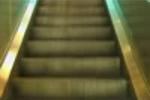}
			\includegraphics[width=.07\linewidth]{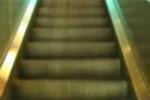}
			\includegraphics[width=.07\linewidth]{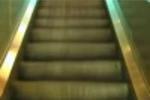}
			\includegraphics[width=.07\linewidth]{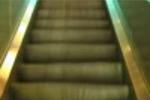}
			\includegraphics[width=.07\linewidth]{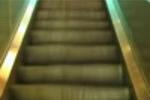}
			\hspace{1mm}
			\includegraphics[width=.07\linewidth]{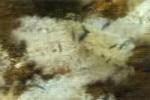}
			\includegraphics[width=.07\linewidth]{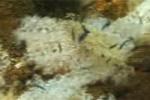}
			\includegraphics[width=.07\linewidth]{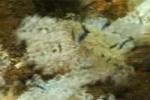}
			\includegraphics[width=.07\linewidth]{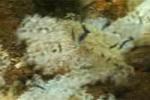}
			\includegraphics[width=.07\linewidth]{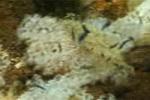}
			\includegraphics[width=.07\linewidth]{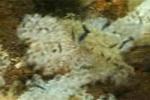} \\
			\vspace{1mm}

			\includegraphics[width=.07\linewidth]{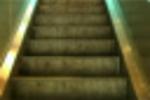}
			\includegraphics[width=.07\linewidth]{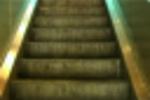}
			\includegraphics[width=.07\linewidth]{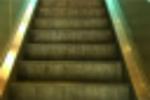}
			\includegraphics[width=.07\linewidth]{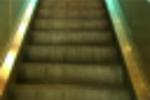}
			\includegraphics[width=.07\linewidth]{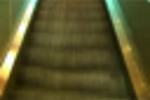}
			\includegraphics[width=.07\linewidth]{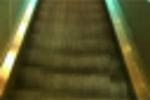}
			\hspace{1mm}
			\includegraphics[width=.07\linewidth]{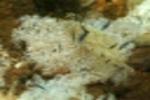}
			\includegraphics[width=.07\linewidth]{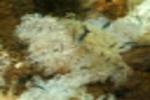}
			\includegraphics[width=.07\linewidth]{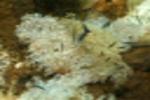}
			\includegraphics[width=.07\linewidth]{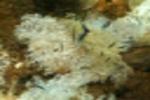}
			\includegraphics[width=.07\linewidth]{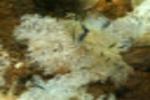}
			\includegraphics[width=.07\linewidth]{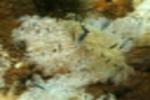} \\
			\vspace{1mm}
			
			\includegraphics[width=.07\linewidth]{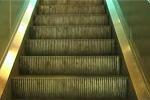}
			\includegraphics[width=.07\linewidth]{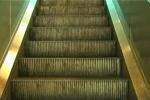}
			\includegraphics[width=.07\linewidth]{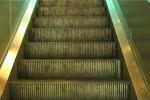}
			\includegraphics[width=.07\linewidth]{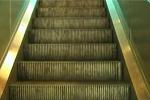}
			\includegraphics[width=.07\linewidth]{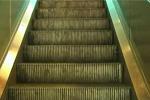}
			\includegraphics[width=.07\linewidth]{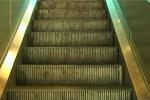}
			\hspace{1mm}
			\includegraphics[width=.07\linewidth]{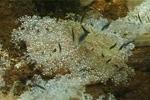}
			\includegraphics[width=.07\linewidth]{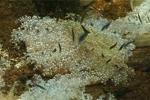}
			\includegraphics[width=.07\linewidth]{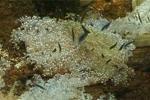}
			\includegraphics[width=.07\linewidth]{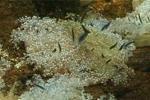}
			\includegraphics[width=.07\linewidth]{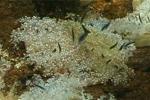}
			\includegraphics[width=.07\linewidth]{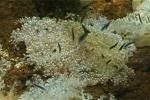} \\
			\hspace{5mm} (a) Elevator\hspace{20mm} (b) Flowers swaying
			
			\includegraphics[width=.07\linewidth]{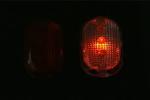}
			\includegraphics[width=.07\linewidth]{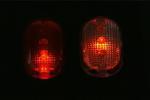}
			\includegraphics[width=.07\linewidth]{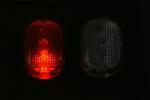}
			\includegraphics[width=.07\linewidth]{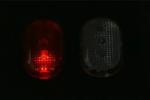}
			\includegraphics[width=.07\linewidth]{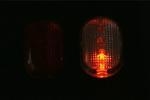}
			\includegraphics[width=.07\linewidth]{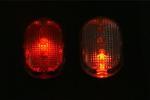}
			\hspace{1mm}
			\includegraphics[width=.07\linewidth]{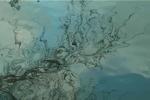}
			\includegraphics[width=.07\linewidth]{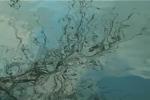}
			\includegraphics[width=.07\linewidth]{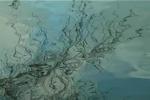}
			\includegraphics[width=.07\linewidth]{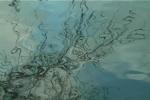}
			\includegraphics[width=.07\linewidth]{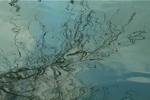}
			\includegraphics[width=.07\linewidth]{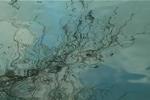} \\
			\vspace{1mm}
			
			\includegraphics[width=.07\linewidth]{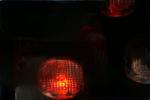}
			\includegraphics[width=.07\linewidth]{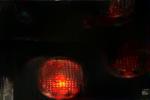}
			\includegraphics[width=.07\linewidth]{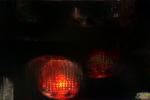}
			\includegraphics[width=.07\linewidth]{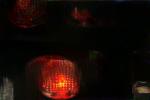}
			\includegraphics[width=.07\linewidth]{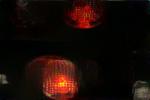}
			\includegraphics[width=.07\linewidth]{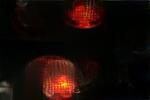}
			\hspace{1mm}
			\includegraphics[width=.07\linewidth]{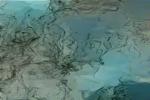}
			\includegraphics[width=.07\linewidth]{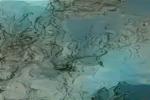}
			\includegraphics[width=.07\linewidth]{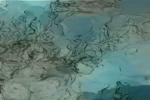}
			\includegraphics[width=.07\linewidth]{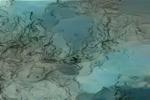}
			\includegraphics[width=.07\linewidth]{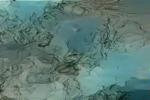}
			\includegraphics[width=.07\linewidth]{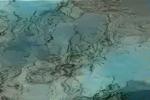} \\
			\vspace{1mm}
			
			\includegraphics[width=.07\linewidth]{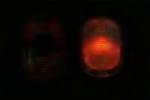}
			\includegraphics[width=.07\linewidth]{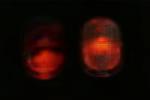}
			\includegraphics[width=.07\linewidth]{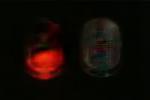}
			\includegraphics[width=.07\linewidth]{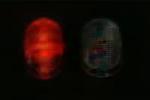}
			\includegraphics[width=.07\linewidth]{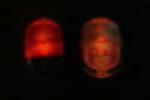}
			\includegraphics[width=.07\linewidth]{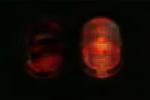}
			\hspace{1mm}
			\includegraphics[width=.07\linewidth]{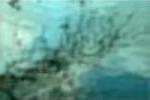}
			\includegraphics[width=.07\linewidth]{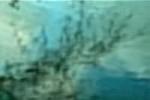}
			\includegraphics[width=.07\linewidth]{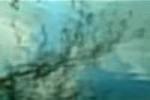}
			\includegraphics[width=.07\linewidth]{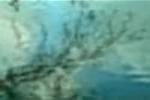}
			\includegraphics[width=.07\linewidth]{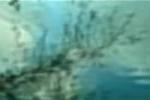}
			\includegraphics[width=.07\linewidth]{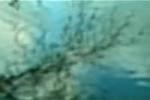} \\
			\vspace{1mm}
			
			\includegraphics[width=.07\linewidth]{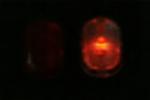}
			\includegraphics[width=.07\linewidth]{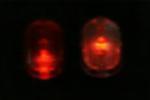}
			\includegraphics[width=.07\linewidth]{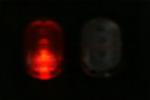}
			\includegraphics[width=.07\linewidth]{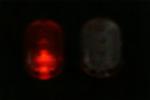}
			\includegraphics[width=.07\linewidth]{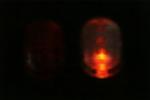}
			\includegraphics[width=.07\linewidth]{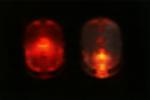}
			\hspace{1mm}
			\includegraphics[width=.07\linewidth]{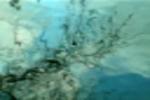}
			\includegraphics[width=.07\linewidth]{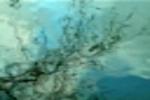}
			\includegraphics[width=.07\linewidth]{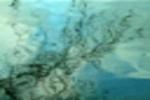}
			\includegraphics[width=.07\linewidth]{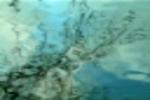}
			\includegraphics[width=.07\linewidth]{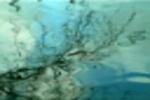}
			\includegraphics[width=.07\linewidth]{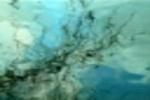} \\
			\vspace{1mm}
			
			\includegraphics[width=.07\linewidth]{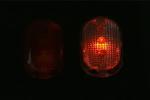}
			\includegraphics[width=.07\linewidth]{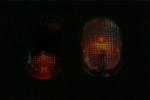}
			\includegraphics[width=.07\linewidth]{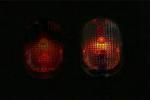}
			\includegraphics[width=.07\linewidth]{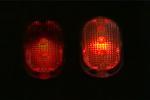}
			\includegraphics[width=.07\linewidth]{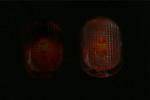}
			\includegraphics[width=.07\linewidth]{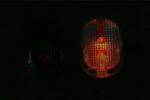}
			\hspace{1mm}
			\includegraphics[width=.07\linewidth]{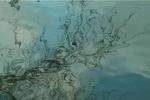}
			\includegraphics[width=.07\linewidth]{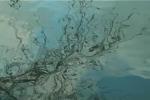}
			\includegraphics[width=.07\linewidth]{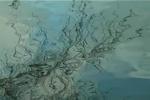}
			\includegraphics[width=.07\linewidth]{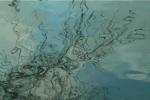}
			\includegraphics[width=.07\linewidth]{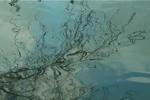}
			\includegraphics[width=.07\linewidth]{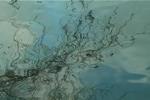} \\
			\hspace{3mm}(c) Flash lights  \hspace{20mm} (d) Water wave 
			
			\includegraphics[width=.07\linewidth]{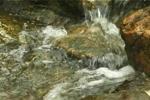}
			\includegraphics[width=.07\linewidth]{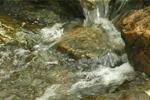}
			\includegraphics[width=.07\linewidth]{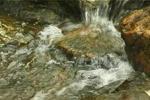}
			\includegraphics[width=.07\linewidth]{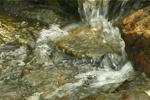}
			\includegraphics[width=.07\linewidth]{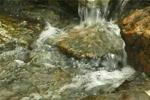}
			\includegraphics[width=.07\linewidth]{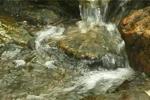}
			\hspace{1mm}
			\includegraphics[width=.07\linewidth]{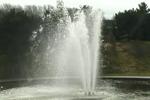}
			\includegraphics[width=.07\linewidth]{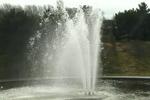}
			\includegraphics[width=.07\linewidth]{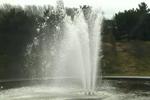}
			\includegraphics[width=.07\linewidth]{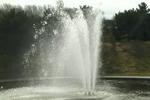}
			\includegraphics[width=.07\linewidth]{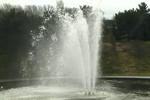}
			\includegraphics[width=.07\linewidth]{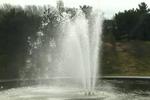} \\
			\vspace{1mm}
			
			\includegraphics[width=.07\linewidth]{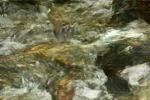}
			\includegraphics[width=.07\linewidth]{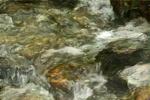}
			\includegraphics[width=.07\linewidth]{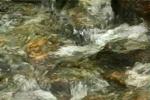}
			\includegraphics[width=.07\linewidth]{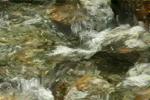}
			\includegraphics[width=.07\linewidth]{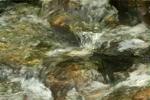}
			\includegraphics[width=.07\linewidth]{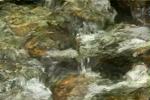}
			\hspace{1mm}
			\includegraphics[width=.07\linewidth]{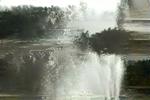}
			\includegraphics[width=.07\linewidth]{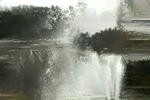}
			\includegraphics[width=.07\linewidth]{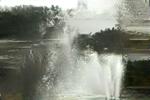}
			\includegraphics[width=.07\linewidth]{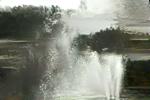}
			\includegraphics[width=.07\linewidth]{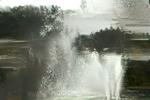}
			\includegraphics[width=.07\linewidth]{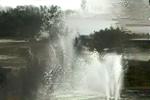} \\
			\vspace{1mm}
			
			\includegraphics[width=.07\linewidth]{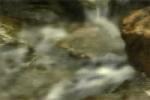}
			\includegraphics[width=.07\linewidth]{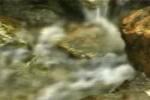}
			\includegraphics[width=.07\linewidth]{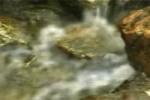}
			\includegraphics[width=.07\linewidth]{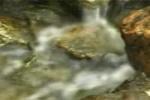}
			\includegraphics[width=.07\linewidth]{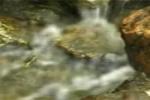}
			\includegraphics[width=.07\linewidth]{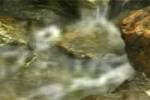}
			\hspace{1mm}
			\includegraphics[width=.07\linewidth]{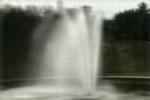}
			\includegraphics[width=.07\linewidth]{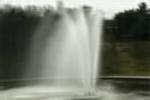}
			\includegraphics[width=.07\linewidth]{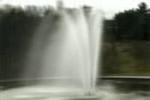}
			\includegraphics[width=.07\linewidth]{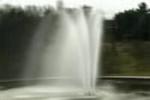}
			\includegraphics[width=.07\linewidth]{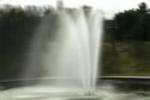}
			\includegraphics[width=.07\linewidth]{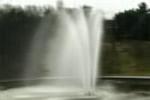} \\
			\vspace{1mm}

			\includegraphics[width=.07\linewidth]{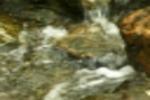}
			\includegraphics[width=.07\linewidth]{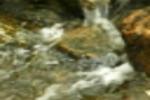}
			\includegraphics[width=.07\linewidth]{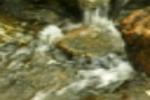}
			\includegraphics[width=.07\linewidth]{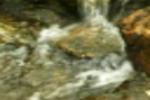}
			\includegraphics[width=.07\linewidth]{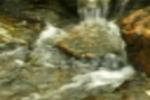}
			\includegraphics[width=.07\linewidth]{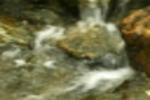}
			\hspace{1mm}
			\includegraphics[width=.07\linewidth]{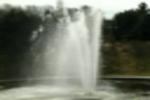}
			\includegraphics[width=.07\linewidth]{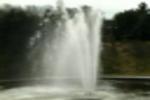}
			\includegraphics[width=.07\linewidth]{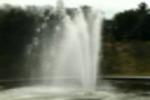}
			\includegraphics[width=.07\linewidth]{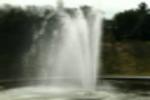}
			\includegraphics[width=.07\linewidth]{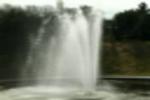}
			\includegraphics[width=.07\linewidth]{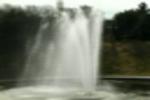} \\
			\vspace{1mm}
			
			\includegraphics[width=.07\linewidth]{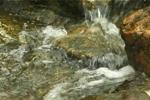}
			\includegraphics[width=.07\linewidth]{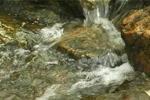}
			\includegraphics[width=.07\linewidth]{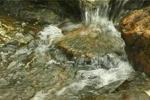}
			\includegraphics[width=.07\linewidth]{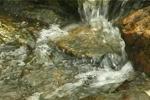}
			\includegraphics[width=.07\linewidth]{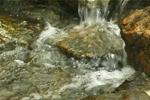}
			\includegraphics[width=.07\linewidth]{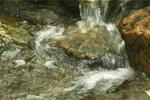}
			\hspace{1mm}
			\includegraphics[width=.07\linewidth]{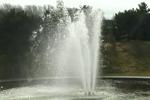}
			\includegraphics[width=.07\linewidth]{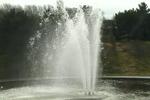}
			\includegraphics[width=.07\linewidth]{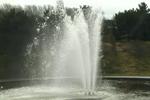}
			\includegraphics[width=.07\linewidth]{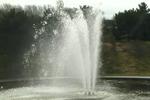}
			\includegraphics[width=.07\linewidth]{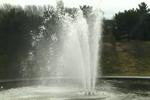}
			\includegraphics[width=.07\linewidth]{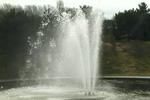} \\
			\hspace{1mm} (e) Spring water \hspace{20mm} (f) Water spray
			
			\caption{Visual quality comparison between three neural-network-based methods and our method for six different DT videos. For each category, the first row displays 6 frames of the observed sequence, and the other rows display the corresponding frames of the sequences generated by different methods (from top to bottom: TwoStream \cite{tesfaldet2018two}, STGCN \cite{Xie2021LearningES}, DG \cite{Xie2019LearningDG} and our method). }
			\label{fig:different-method-images}
		\end{center}
	\end{figure}
	
	Next, our model is compared with three neural-network-based methods, including TwoStream \cite{tesfaldet2018two} and STGCN \cite{Xie2021LearningES} and DG \cite{Xie2019LearningDG} on six DT videos (e.g., elector , flashlights and beach, etc). For a fair comparison, we display six same index frames of generated sequences. As shown in Figure \ref{fig:different-method-images}, the DT sequences generated by TwoStream are divergent because this method cannot effectively synthesize DTs that are not spatially homogeneous (e.g., elevator, water spray). As for STGCN and DG, the DT sequences generated by these appear blurred because these two methods rely on more training data. Intuitively, our method generates high-fidelity DT sequences, including realistic details of DTs.

	Finally, we report the time consumption of different DT synthesis methods including neural-network-based and non-neural-network-based methods. As shown in Table \ref{Table:All-Time-consuming}, our method non-neural-network-based methods (except for Kernel-DT) achieve real-time (25 fps) generation with 46.040 fps, while neural-network-based methods fail. Moreover, the neural-network-based methods are time-consuming and computationally expensive to train. In summary, our method synthesizes high-quality DT videos with high speed and low computation, making it superior to the state-of-the-art DT methods. This can be attributed to its discriminative representation of kernel similarity embedding for exhibiting the spatial-temporal transition of DTs. It directly shows that the similarity correlation of different frames is critical prior knowledge for DT synthesis.

	\begin{table*}[t]
		\centering
		\captionsetup{justification=centering}
		\caption{\textsc{Comparison with State-of-the-art DT Synthesis Methods in Terms of Time consumption.}}\label{Table:All-Time-consuming}
		\setlength{\tabcolsep}{0.6mm}{ 
			\begin{tabular}{l|c|c|c|c|c|c|c|c|c|c}
				\hline        
				& \textbf{Ours} & FFT-LDS \cite{Abraham2005DynamicTW} & HOSVD \cite{Costantini2008HigherOS} &  KPCR \cite{you2016kernel}  & LDS \cite{doretto2003dynamic}  & SLDS \cite{Siddiqi2007ACG} & Kernel-DT \cite{Chan2007ClassifyingVW} & TwoStream \cite{tesfaldet2018two} & STGCN \cite{Xie2021LearningES} & DG \cite{Xie2019LearningDG} \\   
				\hline
				Train. time (Sec.) & 0.090 & 0.928 & 1.399 & 1.990 & 0.148 & 2.475 & 0.830 & - & 4188 & 3904.418  \\
				Test time (Sec.) & 26.060 & 5.516 &4.922 &  12.214 & 4.048  & 3.799 & 1007.260 & 8235 & 7.210 & 52.292  \\
				Generated frames & 1200 & 1200 & 1200 & 1200 & 1200 & 1200 & 1200  &12 & 70 & 120 \\
				Using GPU & $\times$ & $\times$ & $\times$ & $\times$ & $\times$ & $\times$ & $\times$ &  $\checkmark$ & $\checkmark$ & $\checkmark$ \\
				FPS  & 46.040 & 217.560 & 243.790 & 98.248 & 296.450 & 315.900 & 1.191 & 0.002 & 9.709 & 2.295\\
				\hline
		\end{tabular}}
	\end{table*}

	\section{dicussion}\label{sec5}
	In this study, we propose a novel DT synthesis method to address the high-dimensionality and small sample issues of DT synthesis. Specifically, our method leverages a kernel similarity matrix to mine and capture the similarity as prior knowledge of DT, which is incorporated into the kernel similarity embedding. Then, high-fidelity DTs are synthesized iteratively by the learned model. Notably, our method is dissimilar to the existing kernel-based DT synthesis methods \cite{Chan2007ClassifyingVW,you2016kernel}, which use the kernel function to learn a nonlinear observation function for dimensionality-reduction. The experimental results on several benchmarks show that the similarity correlation is critical prior knowledge for representing DTs and the kernel similarity embedding effectively solves the aforementioned issues. Thus, our method can achieve promising results for DT synthesis.
	
	\begin{figure}[t]
		\begin{center}
			\includegraphics[width=.076\linewidth]{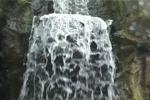}
			\includegraphics[width=.076\linewidth]{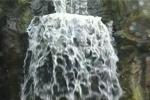}
			\includegraphics[width=.076\linewidth]{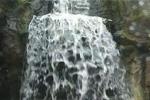}
			\includegraphics[width=.076\linewidth]{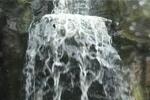}
			\includegraphics[width=.076\linewidth]{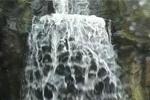}
			\includegraphics[width=.076\linewidth]{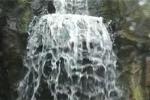}
			\includegraphics[width=.076\linewidth]{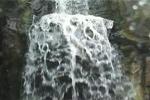}
			\includegraphics[width=.076\linewidth]{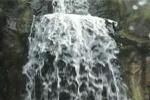}
			\includegraphics[width=.076\linewidth]{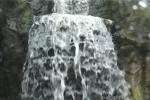}
			\includegraphics[width=.076\linewidth]{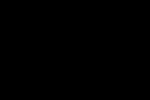}
			\includegraphics[width=.076\linewidth]{./limitation/blank-frame.jpg}
			\\
			\vspace{1mm}
			
			\includegraphics[width=.076\linewidth]{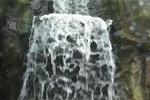}
			\includegraphics[width=.076\linewidth]{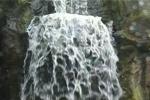}
			\includegraphics[width=.076\linewidth]{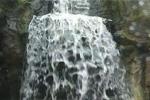}
			\includegraphics[width=.076\linewidth]{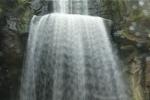}
			\includegraphics[width=.076\linewidth]{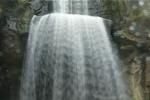}
			\includegraphics[width=.076\linewidth]{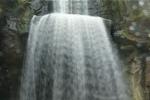}
			\includegraphics[width=.076\linewidth]{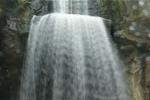}
			\includegraphics[width=.076\linewidth]{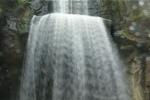}
			\includegraphics[width=.076\linewidth]{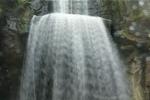}
			\includegraphics[width=.076\linewidth]{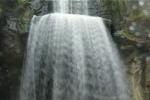}
			\includegraphics[width=.076\linewidth]{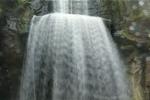}
			\\ (a) Waterfall
			\vspace{1mm}

			\includegraphics[width=.076\linewidth]{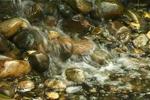}
			\includegraphics[width=.076\linewidth]{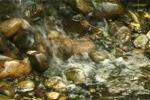}
			\includegraphics[width=.076\linewidth]{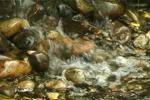}
			\includegraphics[width=.076\linewidth]{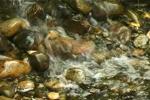}
			\includegraphics[width=.076\linewidth]{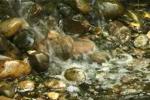}
			\includegraphics[width=.076\linewidth]{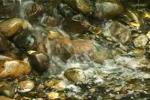}
			\includegraphics[width=.076\linewidth]{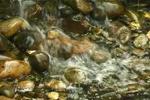}
			\includegraphics[width=.076\linewidth]{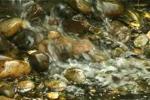}
			\includegraphics[width=.076\linewidth]{./limitation/blank-frame.jpg}
			\includegraphics[width=.076\linewidth]{./limitation/blank-frame.jpg}
			\includegraphics[width=.076\linewidth]{./limitation/blank-frame.jpg}
			\\
			\vspace{1mm}
			
			\includegraphics[width=.076\linewidth]{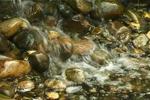}
			\includegraphics[width=.076\linewidth]{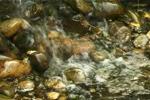}
			\includegraphics[width=.076\linewidth]{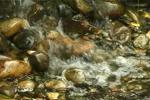}
			\includegraphics[width=.076\linewidth]{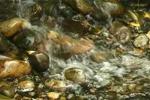}
			\includegraphics[width=.076\linewidth]{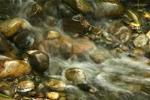}
			\includegraphics[width=.076\linewidth]{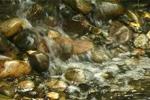}
			\includegraphics[width=.076\linewidth]{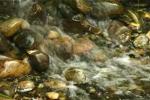}
			\includegraphics[width=.076\linewidth]{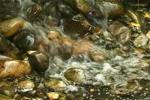}
			\includegraphics[width=.076\linewidth]{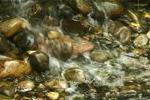}
			\includegraphics[width=.076\linewidth]{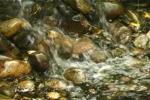}
			\includegraphics[width=.076\linewidth]{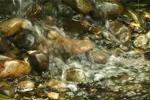}
			\\ (b) Spring water
			\vspace{1mm}

			\includegraphics[width=.076\linewidth]{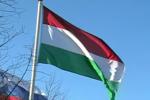}
			\includegraphics[width=.076\linewidth]{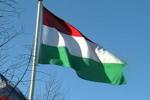}
			\includegraphics[width=.076\linewidth]{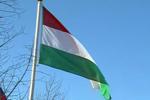}
			\includegraphics[width=.076\linewidth]{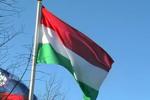}
			\includegraphics[width=.076\linewidth]{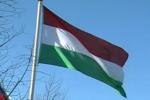}
			\includegraphics[width=.076\linewidth]{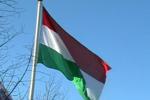}
			\includegraphics[width=.076\linewidth]{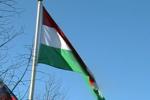}
			\includegraphics[width=.076\linewidth]{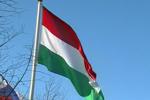}
			\includegraphics[width=.076\linewidth]{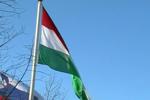}
			\includegraphics[width=.076\linewidth]{./limitation/blank-frame.jpg}
			\includegraphics[width=.076\linewidth]{./limitation/blank-frame.jpg}
			\\
			\vspace{1mm}
			
			\includegraphics[width=.076\linewidth]{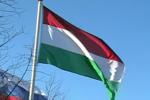}
			\includegraphics[width=.076\linewidth]{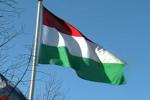}
			\includegraphics[width=.076\linewidth]{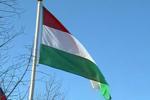}
			\includegraphics[width=.076\linewidth]{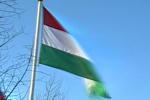}
			\includegraphics[width=.076\linewidth]{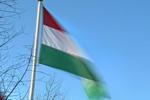}
			\includegraphics[width=.076\linewidth]{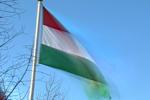}
			\includegraphics[width=.076\linewidth]{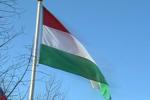}
			\includegraphics[width=.076\linewidth]{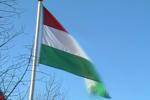}
			\includegraphics[width=.076\linewidth]{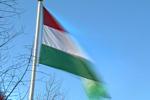}
			\includegraphics[width=.076\linewidth]{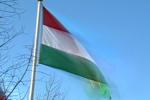}
			\includegraphics[width=.076\linewidth]{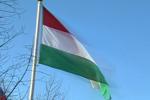}
			\\ (c) Flag
			
			\caption{Some generated frames of DT sequences displaying the limitations of our method. For each category, the first row displays the 11 frames of the observed sequence (black frame indicates that the corresponding frame of the observed sequence is lacking), and the second row shows the corresponding frames of videos synthesized by our method. From left to right, the columns are the 2-nd, 100-th, 200-th, 300-th, 400-th, 500-th, 600-th, 700-th, 800-th, 900-th, 1000-th frames of the observed and synthesized sequences.}
			\label{fig:limitation-images}
		\end{center}
	\end{figure}
	
	To evaluate our method, we intuitively and theoretically analyzed the effectiveness of the kernel similarity embedding for DT synthesis (Section \ref{sec3.3}). Then, we evaluated the influence of the kernel function, the regularization factor $\lambda$, and the kernel size $\gamma$ (Section \ref{sec4.2}). Meanwhile, we intuitively validated the sustainability and generalization of our method using vision quality and quantitative evaluation metrics (Section \ref{sec4.3} and Section \ref{sec4.4}). Finally, we compared our method to nine state-of-the-art methods (including neural-network-based and non-neural-network-based methods) (Section \ref{sec4.5}).
	
	Although our method achieves significant results, it is limited in that its sustainability is impacted if the DTs lack stochastic repetitiveness in the temporal dimension (e.g., waterfall, spring water). As can be seen from Figure \ref{fig:limitation-images}, our method falls into visual decay after 200 frames (the length of the training frames), and thus fails to synthesize high-fidelity DT in the long-term. In the future, scalable kernel similarity embedding may be a potential choice to overcome this limitation. Because scalable kernel similarity embedding can learn an optimal similarity representation for various data, and thus the statistical stationarity in the spatial domain and stochastic repetitiveness in the temporal dimension can be well learned. To achieve this goal, we argue that multiple kernel learning (MKL) \cite{Zhou2020MultipleKC,Liu2020OptimalNM,Liu2020AbsentMK,Liu2016MultipleKK,Liu2021HierarchicalMK,AlioschaPrez2020SVRGMKLAF} may be a good solution for constructing the scalable kernel similarity matrix by learning multiple kernel features. Specifically, we could employ a group of base kernels to learn multiple kernel features, which could then be combined as a novel similarity matrix with a set of combination coefficients. Accordingly, the optimal combination coefficients and the model parameters could be jointly learned.

	\section{Conclusion}\label{sec6}
	In this paper, we have proposed a novel DT synthesis method that integrates kernel learning and the extreme learning machine into a powerfully unified synthesis method to learn kernel similarity embeddings for representing DTs. Notably, kernel similarity embedding not only effectively addresses the high-dimensionality and small sample issues by using similarity as prior knowledge, but also has the advantage of modeling nonlinear feature representation relationships for DT. The competitive results on DT videos collected from two benchmark datasets and the internet demonstrate the superiority and great potential of our method for DT synthesis. It also shows obvious advantages over all the compared state-of-the-art approaches.

	In the future, we will design an MKL model to learn a scalable kernel similarity embedding for DT synthesis, which will enable the similarity representations of DT to effectively be controlled. Furthermore, we will also incorporate multi-view methods into DT synthesis, as some DT sequences were acquired with moving cameras and different views (e.g., some DT videos in Dyntex).

	\ifCLASSOPTIONcaptionsoff
	\newpage
	\fi

	\bibliographystyle{IEEEtran}
	\bibliography{mybibfile}

	\begin{IEEEbiography}[{\includegraphics[width=1in,height=1.25in,clip,keepaspectratio]{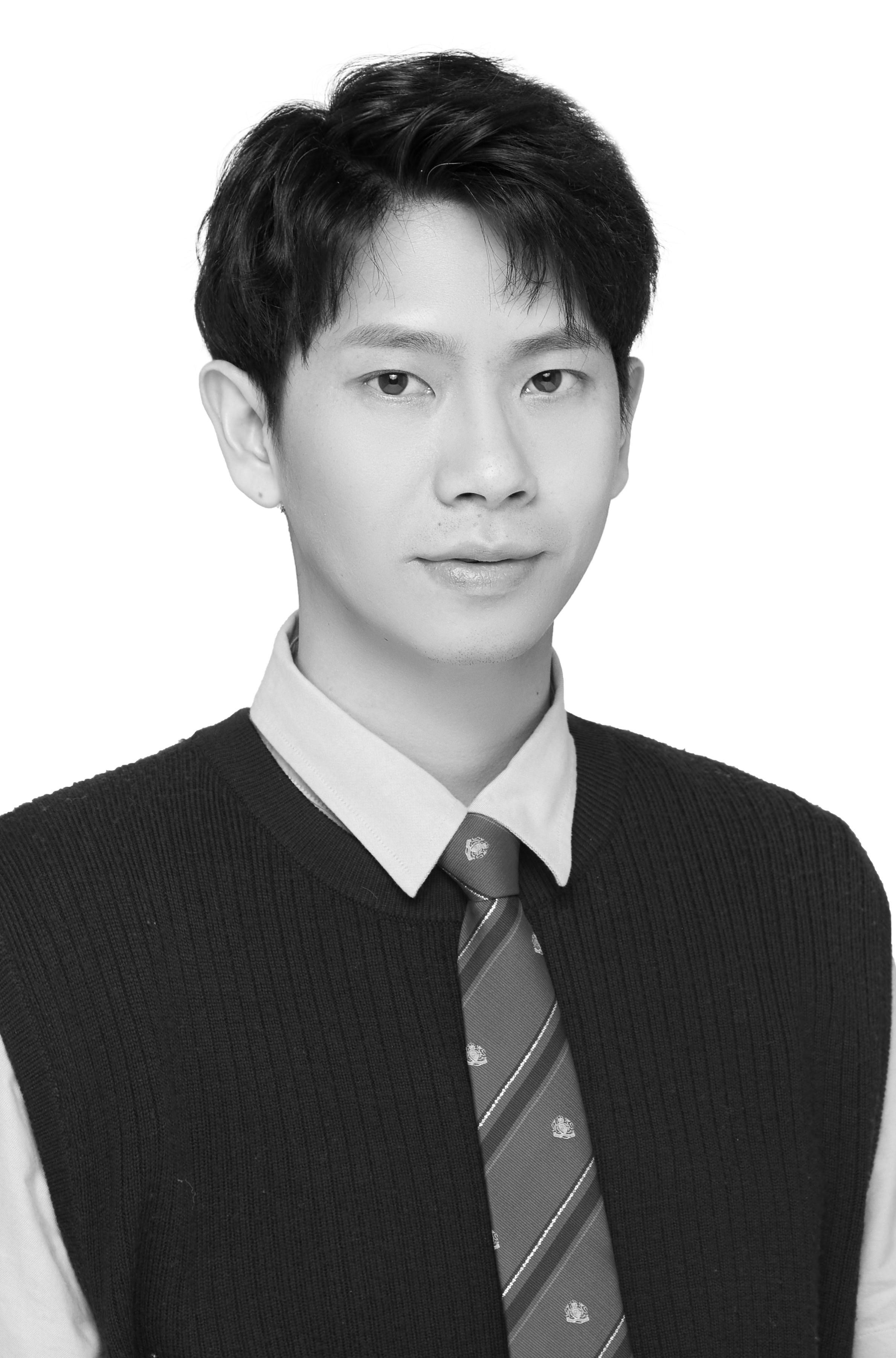}}]{Shiming Chen}
		is currently a full-time Ph.D student in the School of Electronic Information and Communications, Huazhong University of Sciences and Technology (HUST), China. His research results have expounded in 10+ publications at prestigious journals and prominent conferences, such as  IEEE T-EVC, T-NNLS, NeruIPS, CVPR, ICCV, AAAI, IJCAI, etc. He serves as the reviewer for prestigious journals such as IEEE T-PAMI, T-IP, T-NNLS, T-CYB, T-EVC, T-II, T-SMCA, Information Fusion, etc. His current research interests span computer vision and machine learning with a series of topics, such as generative modeling and learning, and zero-shot learning. 
	\end{IEEEbiography}

	\begin{IEEEbiography}[{\includegraphics[width=1in,height=1.25in,clip,keepaspectratio]{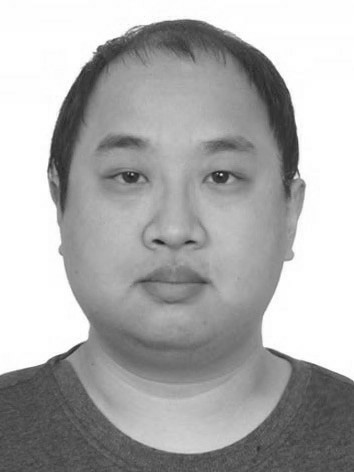}}]{Peng Zhang}
		received the Ph.D. degree in the School of Electronic Information and Communications, Huazhong University of Science and Technology, China. His research interests include computer vision, pattern recognition, and machine learning.
	\end{IEEEbiography}

	\begin{IEEEbiography}[{\includegraphics[width=1in,height=1.25in,clip,keepaspectratio]{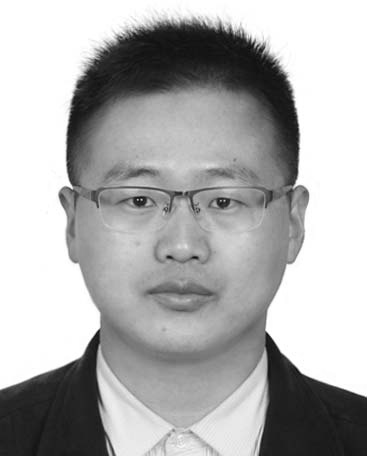}}]{Guosen Xie}
		received the Ph.D. degree from the National Laboratory of Pattern Recognition, Institute of Automation, Chinese Academy of Sciences,
		Beijing, China, in 2016. He was with the Inception Institute of Artificial Intelligence (IIAI), Abu Dhabi, UAE.  His research results have expounded in 20+ publications at prestigious journals and prominent conferences, such as  IEEE T-IP, T-NNLS, T-CSVT, CVPR, and ECCV. His research interests include computer vision and machine learning.
	\end{IEEEbiography}

	\begin{IEEEbiography}[{\includegraphics[width=1in,height=1.25in,clip,keepaspectratio]{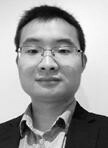}}]{Qinmu Peng}
		is currently an Assistant Professor in the School of Electronics Information and Communications, Huazhong University of Science and Technology (HUST), China. He received the Ph.D. degree in computer science from Hong Kong Baptist University, Hong Kong, in 2015. His research results have expounded in 20+ publications at prestigious journals and prominent conferences, such as PNAS, IEEE T-NNLS, T-SMCA, T-HMS, T-MM, and IJCAI. His current research interests include multimedia analysis, computer vision, and medical image analysis.
	\end{IEEEbiography}
	
	
	\vspace{2cm}
	\begin{IEEEbiography}[{\includegraphics[width=1in,height=1.25in,clip,keepaspectratio]{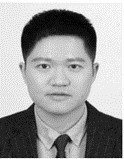}}]{Zehong Cao}
		is  associate professor of the University of South Australia and an Adjunct Fellow with the School of Computer Science, UTS. He received the Ph.D. degree in information technology from the University of Technology Sydney, Australia, in 2017.He has authored more than 50 papers published in top-tier International Conferences such as AAMAS and AAAI, and IEEE and ACM Transactions Series, with 5 ESI highly cited papers. Dr. Cao is the Leading Guest Editor of \textit{IEEE Transactions on Fuzzy Systems}, and \textit{IEEE Transactions on Industrial Informatics} (2020), and the Associate Editor for \textit{Neurocomputing} (2019-) and \textit{Scientific Data} (2019-). His current research interests include computer vision, machine learning, computational intelligence, and bio-signal processing.
	\end{IEEEbiography}

	\vspace{-0.6cm}
	\begin{IEEEbiography}[{\includegraphics[width=1in,height=1.25in,clip,keepaspectratio]{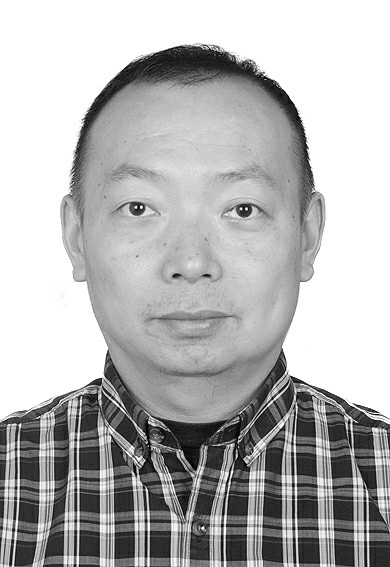}}]{Wei Yuan}
		is currently a professor with the School of Electronic Information and Communications, Huazhong University of Science and Technology, China. He received the B.S. degree in electronic engineering from Wuhan University, China, in 1999, and the Ph.D. degree in electronic engineering from the University of Science and Technology of China, Hefei, in 2006. His research results have expounded in 20+ publications at prestigious journals and prominent conferences, such as  IEEE T-SP, T-MC, T-WC, T-CYB, T-SG, T-SMCA, and WWW. His current research interests include machine learning, computer vision and information security.
	\end{IEEEbiography}
	
	\begin{IEEEbiography}[{\includegraphics[width=1in,height=1.25in,clip,keepaspectratio]{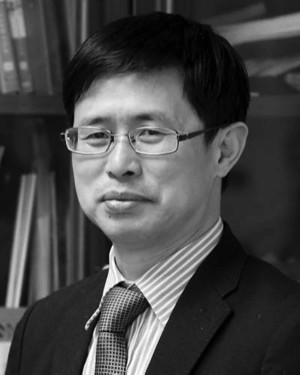}}]{Xinge You}
		(Senior Member, IEEE) is currently a Professor with the School of Electronic Information and Communications, Huazhong University of Science and Technology, Wuhan. He received the B.S. and M.S. degrees in mathematics from Hubei University, Wuhan, China, in 1990 and 2000, respectively, and the Ph.D. degree from the Department of Computer Science, Hong Kong Baptist University, Hong Kong, in 2004. His research results have expounded in 60+ publications at prestigious journals and prominent conferences, such as  IEEE T-PAMI, T-IP, T-NNLS, T-CYB, T-CSVT, CVPR, ECCV, IJCAI. He served/serves as an Associate Editor of the \textit{IEEE Transactions on Cybernetics}, \textit{IEEE Transactions on Systems, Man, Cybernetics:Systems}. His current research interests include image processing, wavelet analysis and its applications, pattern recognition, machine earning, and computer vision.
	\end{IEEEbiography}

\end{document}